\documentclass[acmsmall,screen]{acmart}

\setcopyright{none}     
\renewcommand\footnotetextcopyrightpermission[1]{}  
\makeatletter
\renewcommand{\@authorsaddresses}{} 
\makeatother

\usepackage{hyperref} 

\AtBeginDocument{
  \providecommand\BibTeX{{
    \normalfont B\kern-0.5em{\scshape i\kern-0.25em b}\kern-0.8em\TeX}}}

\usepackage{graphicx}
\usepackage{wrapfig}
\usepackage{float} 
\usepackage{csquotes}
\usepackage{xcolor}
\usepackage{bm}
\usepackage{wrapfig}
\usepackage{amsmath}  
\usepackage{mathtools}
\usepackage[most]{tcolorbox}
\usepackage{tikz}
\usetikzlibrary{mindmap}
\usepackage{booktabs}  
\usepackage{adjustbox}
\usepackage{tabularx}
\usepackage{multirow}  
\usepackage{enumitem}  
\usepackage{algorithmic}  
\usepackage{algorithm}  
\usepackage{forest} 
\usepackage{todonotes}
\usepackage{url}
\usepackage{float}  
\usepackage{comment}  
\usepackage{subfigure}
\usepackage{cleveref}
\usepackage{colortbl}
\usepackage{hyperref}

\definecolor{lightcoral}{rgb}{0.94, 0.5, 0.5}
\definecolor{lightgreen}{rgb}{0.56, 0.93, 0.56}
\definecolor{harvestgold}{rgb}{0.85, 0.57, 0.0}
\definecolor{brightlavender}{rgb}{0.75, 0.58, 0.89}
\definecolor{capri}{rgb}{0.0, 0.75, 1.0}
\definecolor{carminepink}{rgb}{0.92, 0.3, 0.26}
\definecolor{celadon}{rgb}{0.67, 0.88, 0.69}
\definecolor{darkpastelgreen}{rgb}{0.01, 0.75, 0.24}
\definecolor{DeepSkyBlue4}{RGB}{0,104,139}

\crefformat{section}{\S#2#1#3} 
\crefformat{subsection}{\S#2#1#3}
\crefformat{subsubsection}{\S#2#1#3}

\makeatletter

\renewcommand{\l@section}[2]{
  \par\addpenalty{-\@highpenalty}
  \vskip 3pt 
  \@dottedtocline{1}{0em}{2.3em}{\textbf{#1}}{#2}
}

\renewcommand{\l@subsection}[2]{
  \par\addpenalty{-\@highpenalty}
  \vskip 1pt 
  \@dottedtocline{2}{1.5em}{2.8em}{#1}{#2}
}

\renewcommand{\l@subsubsection}[2]{
  \par\addpenalty{-\@highpenalty}
  \vskip 1pt 
  \@dottedtocline{3}{3em}{3.3em}{#1}{#2}
}

\makeatother

\usepackage{ifthen}
\usepackage{pifont}

\definecolor{brandblue}{rgb}{0.54, 0.7, 1}

\newcommand{\mathbox}[2][]{
  \ifthenelse{\equal{#1}{post-pro}}{
    \tcboxmath[colback=yellow!30!white, colframe=white, boxrule=0pt, rounded corners, fontupper=\bfseries\large, left=0pt, right=0pt, top=2.2pt, bottom=2.2pt, boxsep=0pt]{#2}  
  }{
    \ifthenelse{\equal{#1}{model}}{
      \tcboxmath[colback=green!10!white, colframe=white, boxrule=0pt, rounded corners, fontupper=\bfseries\large, left=1pt, right=1pt, top=2.2pt, bottom=2.2pt, boxsep=0pt]{#2}  
    }{
      \ifthenelse{\equal{#1}{eval}}{
        \tcboxmath[colback=brandblue!30!white, colframe=white, boxrule=0pt, rounded corners, fontupper=\bfseries\large, left=3pt, right=3pt,top=2.2pt, bottom=2.2pt,  boxsep=1pt]{#2}  
      }{
        \ifthenelse{\equal{#1}{icl}}{
          \tcboxmath[colback=orange!20!white, colframe=white, boxrule=0pt, rounded corners, fontupper=\bfseries\large, left=1pt, right=1pt,top=2pt, bottom=2pt, boxsep=0pt]{#2}  
        }{
          \textcolor{black}{\large #2} 
        }
      }
    }
  }
}

\usepackage{titlesec}

\titleclass{\subsubsubsection}{straight}[\subsubsection]
\newcounter{subsubsubsection}[subsubsection]
\renewcommand\thesubsubsubsection{\thesubsubsection.\arabic{subsubsubsection}}

\titleformat{\subsubsubsection}[runin] 
  {\normalfont\normalsize\itshape}     
  {\thesubsubsubsection}               
  {1em}                                
  {}                                   
  []                                  

\titlespacing*{\subsubsubsection}
  {0pt}{3.25ex plus 1ex minus .2ex}{1em} 

\makeatletter
\def\toclevel@subsubsubsection{4}
\def\l@subsubsubsection{\@dottedtocline{4}{7em}{4em}}
\makeatother

\begin{document}

\title{A Survey on Parallel Text Generation: From Parallel Decoding to Diffusion Language Models}

\author{Lingzhe Zhang\textsuperscript{1,*}, Liancheng Fang\textsuperscript{2,*}, Chiming Duan\textsuperscript{1,*}, Minghua He\textsuperscript{1,*}, Leyi Pan\textsuperscript{3,*}, Pei Xiao\textsuperscript{1}, Shiyu Huang\textsuperscript{4}, Yunpeng Zhai\textsuperscript{5}, Xuming Hu\textsuperscript{6},  Philip S. Yu\textsuperscript{2}, Aiwei Liu\textsuperscript{3,*}}

\begingroup
\renewcommand{\thefootnote}{\fnsymbol{footnote}}
\footnotetext[1]{These authors contributed equally to this research.}
\footnotetext{Authors’ emails: Lingzhe Zhang, zhang.lingzhe@stu.pku.edu.cn; Liancheng Fang, lfang87@uic.edu; Chiming Duan, duanchiming@stu.pku.edu.cn; Minghua He, hemh2120@stu.pku.edu.cn; Leyi Pan, panly24@mails.tsinghua.edu.cn; Aiwei Liu, liuaiwei20@gmail.com}
\endgroup

\affiliation{\\
  \textsuperscript{1}Peking University\country{China}\\
  \textsuperscript{2}University of Illinois Chicago \country{United States} \\
  \textsuperscript{3}Tsinghua University\country{China}\\
  \textsuperscript{4}XPENG \country{China} \\
  \textsuperscript{5}Alibaba Group \country{China} \\
  \textsuperscript{6}The Hong Kong University of Science and Technology (Guangzhou) \country{China}\\
}

\renewcommand{\shortauthors}{Lingzhe Zhang, et al.}

\begin{abstract}\label{abstract}

\section*{Abstract}

As text generation has become a core capability of modern Large Language Models (LLMs), it underpins a wide range of downstream applications. However, most existing LLMs rely on autoregressive (AR) generation, producing one token at a time based on previously generated context—resulting in limited generation speed due to the inherently sequential nature of the process. To address this challenge, an increasing number of researchers have begun exploring parallel text generation—a broad class of techniques aimed at breaking the token-by-token generation bottleneck and improving inference efficiency. Despite growing interest, there remains a lack of comprehensive analysis on what specific techniques constitute parallel text generation and how they improve inference performance. To bridge this gap, we present a systematic survey of parallel text generation methods. We categorize existing approaches into AR-based and Non-AR-based paradigms, and provide a detailed examination of the core techniques within each category. Following this taxonomy, we assess their theoretical trade-offs in terms of speed, quality, and efficiency, and examine their potential for combination and comparison with alternative acceleration strategies. Finally, based on our findings, we highlight recent advancements, identify open challenges, and outline promising directions for future research in parallel text generation. We have also created a GitHub repository for indexing relevant papers and open resources available at \href{https://github.com/zhanglingzhe0820/Awesome-Parallel-Text-Generation}{https://github.com/zhanglingzhe0820/Awesome-Parallel-Text-Generation}.

\end{abstract}

\maketitle

\section{Introduction}\label{sec:introduction}

Text generation has emerged as a core capability of modern Large Language Models (LLMs), such as Qwen-3.0~\cite{yang2025qwen3}, GPT-4.5~\cite{jaech2024openai}, DeepSeek-R1~\cite{guo2025deepseek}. It serves as the foundation for a wide range of downstream applications, including open-ended dialogue~\cite{yi2024survey}, code generation~\cite{wang2023review}, summarization~\cite{zhang2024comprehensive}, storytelling~\cite{zhao2023more}, software maintaining~\cite{zhang2025survey, zhang2024survey, zhang2024multivariate, zhang2025log, zhang2024reducing, zhang2024towards}, and creative writing~\cite{gomez2023confederacy}. As LLMs continue to advance in scale~\cite{kaplan2020scaling,bi2024deepseek}, training data coverage~\cite{OpenAI2023GPT4TR,team2023gemini,dubey2024llama}, instruction-following ability~\cite{ouyang2022training,touvron2023llama2,zhang2025scalalog,zhang2025xraglog,liu2024direct} and reasoning ability~\cite{jaech2024openai,seed2025seed1,hong2025glm,zhang2025thinkfl,zhang2025agentfm,liu2025tisdpo}, their capacity to produce coherent, contextually appropriate, and semantically rich text has become increasingly central to both academic research and real-world deployment.

\subsection{Why Does Parallel Text Generation Matter?}

Most existing LLMs rely on \textit{autoregressive (AR) generation}, producing one token at a time based on the previously generated context~\cite{su2021nonautoregressivetextgenerationpretrained, xiao2023survey}. During inference, the model predicts the next token based on all preceding tokens, typically following a left-to-right decoding order. This step-by-step process is repeated until an end-of-sequence token is generated or a predefined maximum length is reached.

While autoregressive generation enables high-quality and coherent text output by effectively modeling strong sequential dependencies, it also introduces a significant challenge: \textbf{limited generation speed}. Since tokens are produced one at a time in a strictly sequential manner, the inference process cannot be parallelized across output positions. As a result, the total decoding time increases linearly with the length of the generated sequence. This not only limits responsiveness in latency-sensitive applications such as interactive systems and real-time dialogue, but also leads to suboptimal hardware utilization. During idle periods between token generations, computational resources (e.g., GPUs or TPUs) are often underutilized, resulting in inefficient inference execution~\cite{recasens2025mind, yao2023flover, li2024large, kang2022separation}.

To address these challenges, researchers have begun exploring \textit{parallel text generation}—a broad class of techniques designed to overcome the token-by-token generation bottleneck~\cite{chen2023accelerating, nie2025large, gulrajani2023likelihood, sun2023spectr, chen2023cascade, zhang2023draft, yang2023predictive, liu2025mtp, mehra2025multi, cai2024medusa, stern2018blockwise, chen2023accelerating, zheng2024masked, lou2024discrete, wu2025fast, israel2025accelerating, hu2025accelerating, kou2024cllms, stern2019insertion, gu2019levenshtein, geng2021learning, guo2023renewnat}. As illustrated in Figure~\ref{fig: trends}, these methods aim to improve decoding efficiency by enabling the generation of multiple tokens per inference step, or by restructuring the generation paradigm altogether to support higher-level parallelism. Such approaches can significantly boost hardware utilization, reduce end-to-end latency, and increase overall throughput, making long-form and real-time text generation more practical and deployable in real-world scenarios.

\subsection{Why a Survey for Parallel Text Generation?}

\begin{figure}[htbp]
	\centering
	\subfigure[The transition from traditional language models generating one token at a time to parallel text generation methods that produce multiple tokens simultaneously]{
		\begin{minipage}{0.52\linewidth}
			\centering   
			\includegraphics[width=\textwidth]{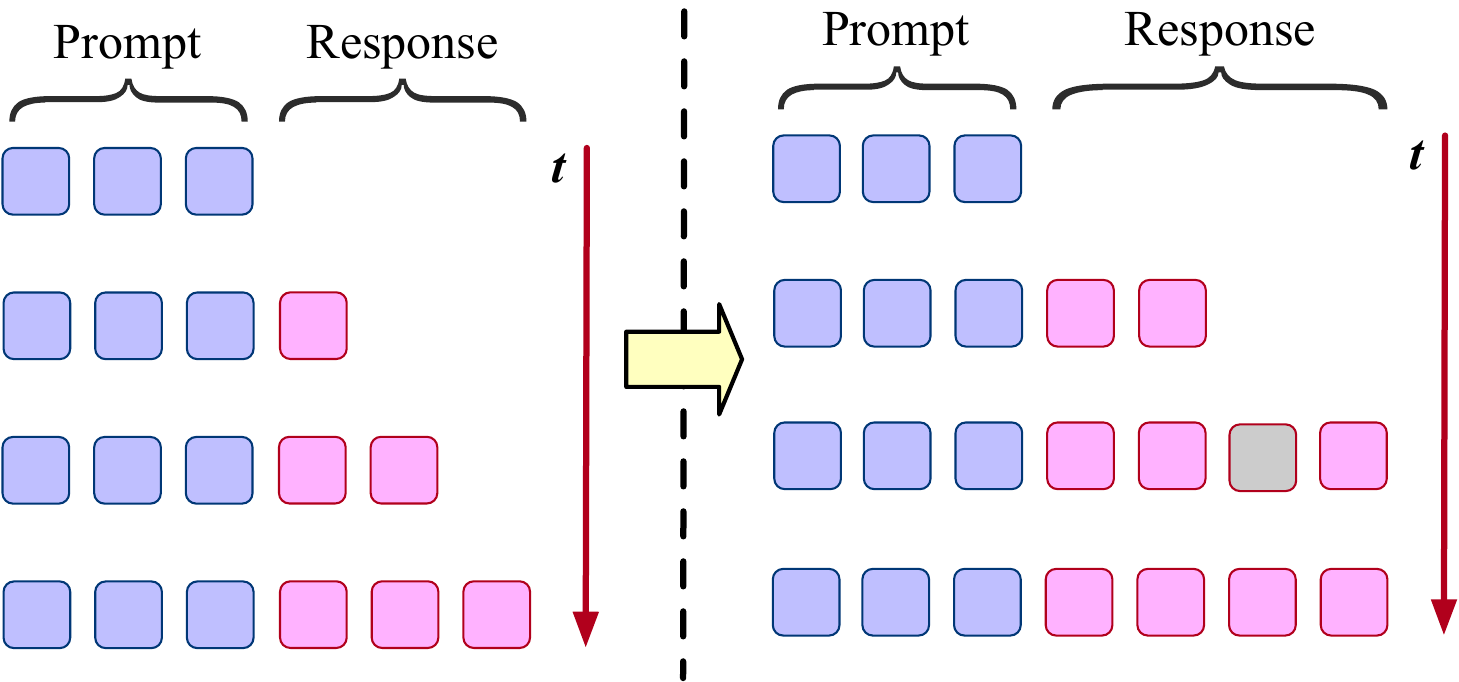}
			\label{fig: trends}
		\end{minipage}
	}
	\hspace{0.01\linewidth}
	\subfigure[Number of publications in the field of Parallel Text Generation and LLMs (the data for "\# Publications in LLMs" is extended based on~\cite{zhao2023survey})]{
		\begin{minipage}{0.43\linewidth}
			\centering
			\includegraphics[width=\textwidth]{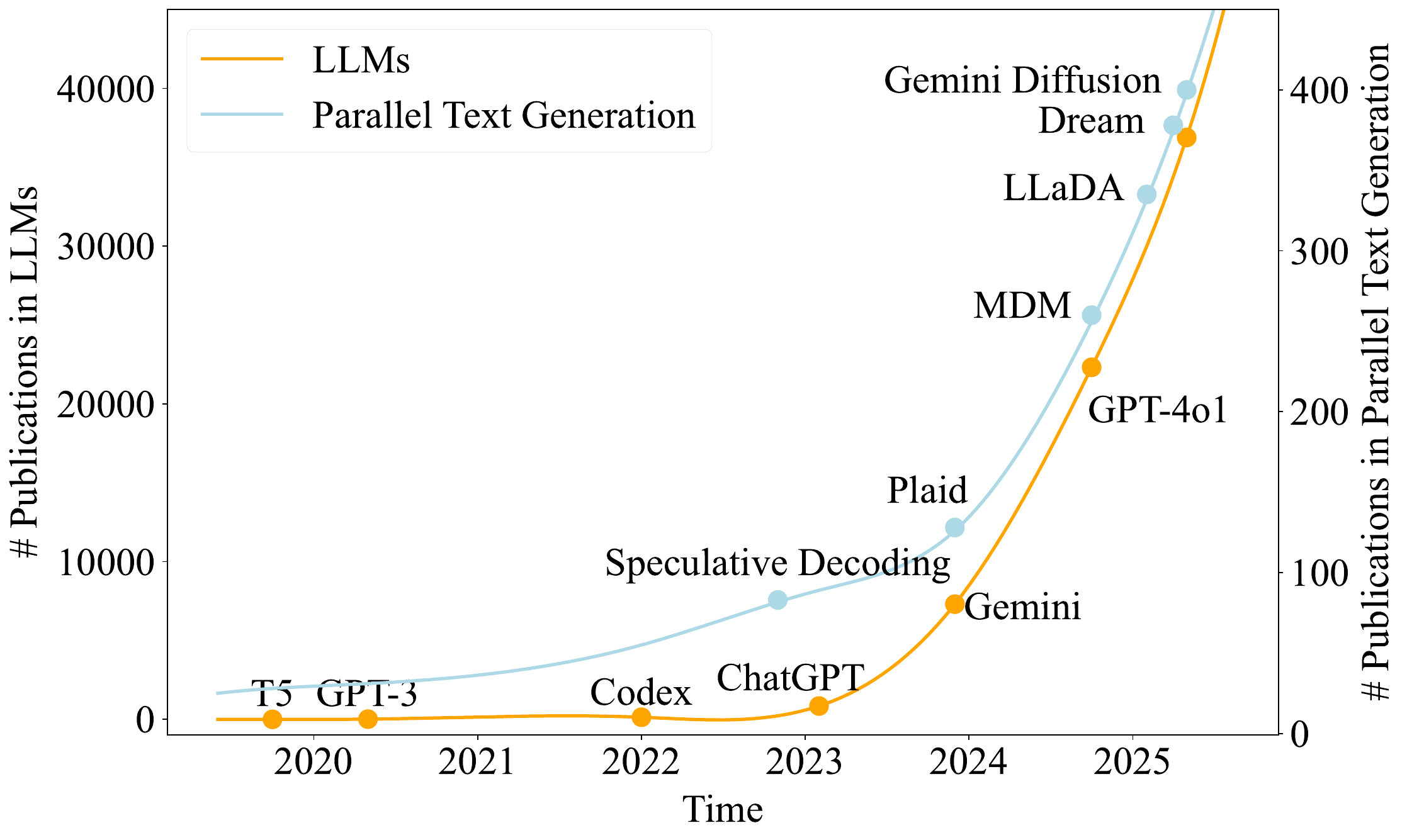}
			\label{fig: paper-num-arxiv}
		\end{minipage}
	}
	\caption{Analysis of Parallel Text Generation Compared with Traditional Large Language Models}
	\label{fig: paper-num}
\end{figure}

Due to the advantages of parallel text generation, an increasing number of research efforts have been devoted to developing diverse approaches for accelerating existing large language models. As shown in Figure~\ref{fig: paper-num-arxiv}, some early attempts at parallel text generation existed even before 2023, in the initial stage of LLM development. However, progress during that period was relatively slow.

The field began to gain momentum following two milestones: the introduction of \textit{speculative decoding} in 2023~\cite{chen2023accelerating}, and the launch of the first widely adopted LLM: ChatGPT~\cite{OpenAI2023GPT4TR}. These early advances were primarily built upon the traditional Transformer architecture, with a focus on improving decoding efficiency through techniques collectively referred to as \textit{parallel decoding}. Representative approaches in this category include speculative decoding~\cite{chen2023accelerating}, blockwise parallel decoding~\cite{stern2018blockwise}, pipelined decoding~\cite{yang2023predictive}, and mask-predict~\cite{ghazvininejad2019mask}.

Over time, this line of work has extended beyond decoding-time acceleration and evolved into fundamentally different generation paradigms. In particular, the emergence of \textit{diffusion-based generation}—which is inherently well-suited for parallel computation due to its iterative, non-autoregressive nature—has significantly broadened the research landscape~\cite{nie2025large}. This trend accelerated rapidly with the introduction of \textit{Plaid} in December 2023~\cite{gulrajani2023likelihood} and culminated in the release of \textit{Google's Gemini Diffusion}. This marked a notable shift in both research focus and industrial deployment, signaling an exciting new phase for parallel text generation.

In fact, in recent years, a number of literature reviews have summarized research related to parallel text generation. However, as shown in Table~\ref{tab: survey-comparison}, these surveys have not systematically covered the full spectrum of approaches that accelerate language models through parallel text generation.

\begin{table}[htbp]
	\centering
	\caption{Comparison of related surveys}
	\label{tab: survey-comparison}
	\begin{tabular}{ccp{9.5cm}<{\raggedright\arraybackslash}}
		\toprule
		Reference & Year & Scope of Parallel Text Generation \\
		\midrule
		Heming et al.~\cite{xia2024unlocking} & 2024 & AR-Based (Speculative Decoding) \\
		\midrule
		Chen et al.~\cite{zhang2024beyond} & 2024 & AR-Based (Speculative Decoding) \\
		\midrule
		Yifan et al.~\cite{li2023diffusion} & 2023 & Non-AR-Based (Diffusion-Based Text Generation) \\
		\midrule
		Qiuhua et al.~\cite{yi2024diffusion} & 2024 & Non-AR-Based (Diffusion-Based Text Generation) \\
		\midrule
		Zhongwei et al.~\cite{wanefficient} & 2023 & AR-Based \\
		\midrule
		\textbf{Our work} & - & AR-Based; Non-AR-Based \\
		\bottomrule
	\end{tabular}
\end{table}

In summary, comprehensive studies that systematically cover the full spectrum of approaches for accelerating language models through parallel text generation are still lacking. In this work, we present the first comprehensive survey that unifies and categorizes these approaches, providing a theoretical comparison of the different parallel generation paradigms in terms of their efficiency and design principles. This survey aims to offer researchers an in-depth understanding of parallel text generation methods and to highlight promising directions for future research.

\subsection{Structure of the paper}

The remainder of this survey is structured as follows:  
Section~2 introduces the necessary background and presents a taxonomy of parallel text generation methods.  
Sections~3 through~5 categorize and discuss representative approaches from three different perspectives.  
Section~6 provides a theoretical comparison and analysis of the various parallel generation paradigms described earlier.  
Section~7 discusses ongoing challenges and potential future research directions in parallel text generation.  
Finally, Section~8 concludes the survey.

\addtocontents{toc}{\protect\setcounter{tocdepth}{-1}}

\clearpage  
\phantomsection  
\addcontentsline{toc}{section}{Table of Contents} 
\tableofcontents
\clearpage

\addtocontents{toc}{\protect\setcounter{tocdepth}{2}}

\section{Overview}\label{sec:overview}

In this section, we first introduce the necessary background, including the foundational concepts of autoregressive and non-autoregressive generation. We then present a taxonomy of the core methods discussed in this paper, organized within this framework.

\subsection{Background}

\subsubsection{Autoregressive Generation}

Autoregressive (AR) generation has been the prevailing paradigm in natural language generation. In its classic and strictest form, it models a text sequence $Y = \{y_1, y_2, ..., y_T\}$ by generating one token at a time, from left to right. Each token is conditioned on all previously generated tokens, as formalized by the chain rule of probability:
\begin{equation}
    P_{\text{AR}}(Y) = \prod_{t=1}^{T} P(y_t \mid y_{<t}, X)
    \label{eq: ar_strict}
\end{equation}
where $X$ denotes the input (if any), and $y_{<t}$ is the semantic prefix generated so far. This strict sequential dependency allows AR models like GPT~\cite{radford2019language} and BART~\cite{lewis2020bart} to produce highly coherent, high-quality text.

However, the landscape of parallel text generation has evolved. Many modern techniques aim to accelerate this process while retaining its core causal nature. To build a unified taxonomy, we adopt a broader definition of autoregressive generation for this survey. We define AR generation as \textbf{any process that maintains a strict unidirectional information flow, where the generation of a token can only depend on previously generated tokens, and not on any future tokens.}

This broader view encompasses methods that generate text in blocks or chunks (e.g., Multiple Token Prediction) rather than token-by-token. While they introduce parallelism within each step, the overall process remains causally ordered—the generation of a later block still depends on the completion of earlier blocks.

\subsubsection{Non-Autoregressive Generation}

In contrast, Non-AR generation paradigms aim for maximum parallelism by \textbf{breaking or entirely removing this strict, left-to-right causal dependency.} In a Non-AR framework, the generation of a token $y_t$ is not constrained to depend only on the prefix $y_{<t}$. Instead, tokens can be generated conditioned on a global context, latent variables, or representations of the entire sequence from a previous refinement step. This allows for the simultaneous generation of all tokens in a single pass (in one-shot models) or parallel updates across the sequence (in iterative models).

\subsubsection{Formalizing the Distinction}

To formalize this distinction, we define the two paradigms based on the nature of the conditioning context $C_t$ used to generate a token $y_t$:

\begin{itemize}
    \item \textbf{AR-based generation:} The generation process is causally constrained. The context $C_t$ for generating token $y_t$ is derived exclusively from information in the prefix $y_{<t}$. This maintains a strict left-to-right dependency, even if implemented in a block-wise fashion.
    
    \item \textbf{Non-AR-based generation:} The generation process is not causally constrained. The context $C_t$ can depend on global information derived from the entire source $X$ or a complete version of the target sequence from a prior state, allowing $y_t$'s generation to be influenced by information from positions $y_{>t}$.
\end{itemize}


This fundamental difference in handling dependencies is the primary reason for the trade-off between generation speed and quality observed across these two paradigms.

\subsubsection{Parallel Text Generation}

\emph{Parallel text generation refers to the process in which two or more tokens are produced within a single inference step.} In other words, we characterize a generation process as parallel if the ratio between the number of inference rounds and the number of output tokens is less than 1, as shown in Equation~\ref{eq:parallel-generation}:

\begin{equation}
    \frac{\text{\# inference steps}}{\text{\# output tokens}} < 1
    \label{eq:parallel-generation}
\end{equation}

This criterion captures the core intuition behind parallel generation: if at least one decoding step emits multiple tokens, the process deviates from strict token-by-token autoregressive decoding and qualifies as parallel.

\vspace{0.5em}
\textbf{AR-compatible approaches.}  
For AR generation, the sequential dependency structure imposes a fundamental limitation: tokens must be generated one at a time, each conditioned on all previous outputs. This makes the decoding process inherently slow and difficult to parallelize. The bottleneck becomes increasingly pronounced when deploying large-scale language models in latency-sensitive or high-throughput applications. To alleviate this issue, several acceleration techniques have been proposed that maintain the AR factorization while partially relaxing the sequential decoding process. These include speculative decoding, blockwise or multiple-token prediction, skeleton-based generation, etc.

\vspace{0.5em}
\textbf{Non-autoregressive approaches.}  
In contrast, Non-AR generation methods remove or significantly relax the dependency between tokens, enabling inherently parallel decoding. Since the generation of one token does not depend on the previously generated ones, all tokens (or large subsets thereof) can be generated simultaneously. This class of methods is naturally aligned with parallelism and includes models based on iterative denoising, masking, insertion, or diffusion. For instance, diffusion-based text generation models generate complete sequences through a series of globally parallel refinement steps, making them strong representatives of fully parallelizable decoding paradigms.

\subsection{Taxonomy}

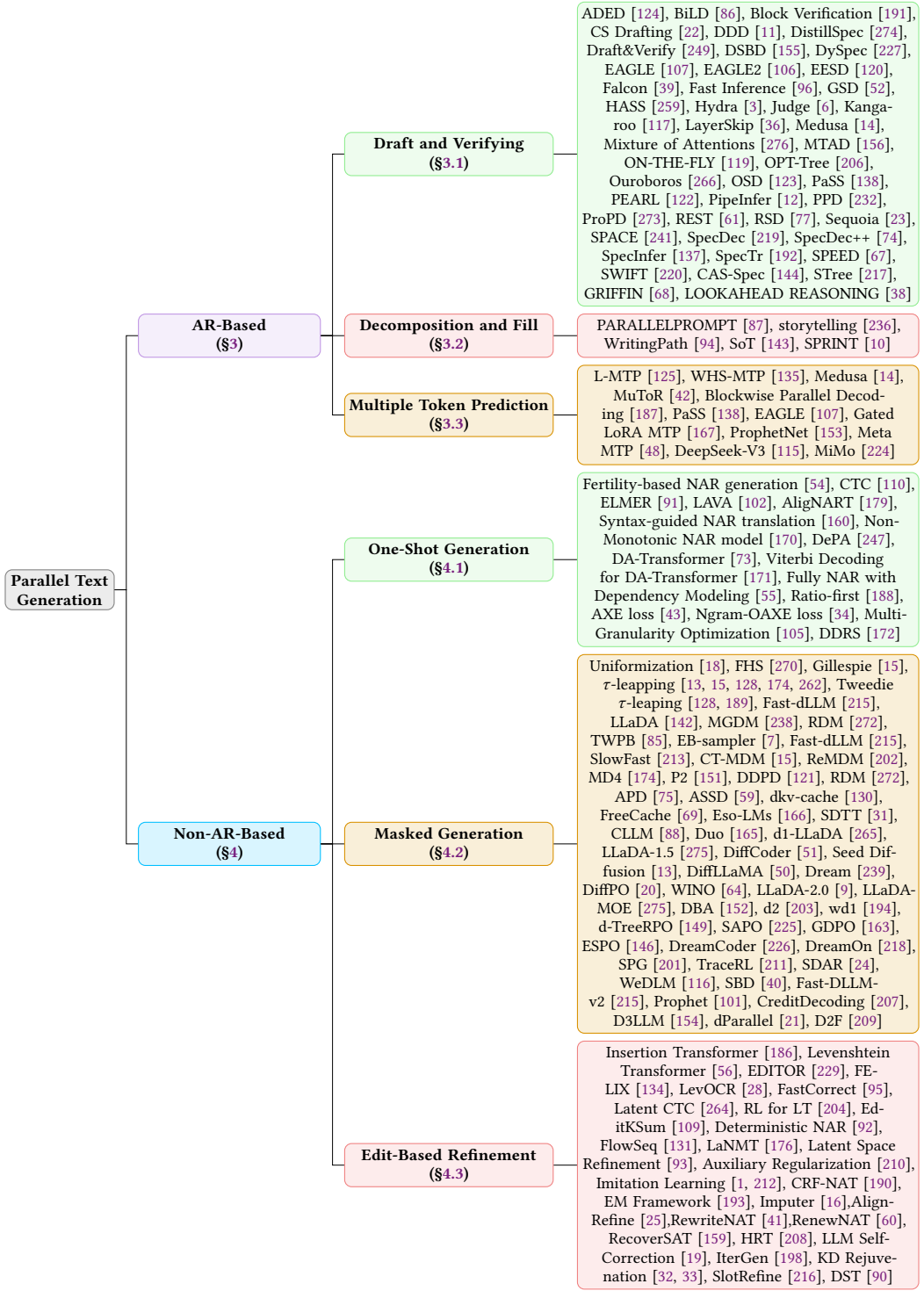
\begin{figure}[htbp]
	\centering
	\tikzset{
		my node/.style={
			draw,
			align=center,
			thin,
			text width=1.2cm, 
			rounded corners=3,
		},
		my leaf/.style={
			draw,
			align=center,
			thin,
			text width=8.5cm, 
			rounded corners=3,
		}
	}
	\forestset{
		every leaf node/.style={
			if n children=0{#1}{}
		},
		every tree node/.style={
			if n children=0{minimum width=1em}{#1}
		},
	}
	\begin{forest}
		nonleaf/.style={font=\bfseries\scriptsize},
		for tree={%
			every leaf node={my leaf, font=\scriptsize},
			every tree node={my node, font=\scriptsize, l sep-=4.5pt, l-=1.pt},
			anchor=west,
			inner sep=2pt,
			l sep=10pt, 
			s sep=3pt, 
			fit=tight,
			grow'=east,
			edge={ultra thin},
			parent anchor=east,
			child anchor=west,
			if n children=0{}{nonleaf}, 
			edge path={
				\noexpand\path [draw, \forestoption{edge}] (!u.parent anchor) -- +(5pt,0) |- (.child anchor)\forestoption{edge label};
			},
			if={isodd(n_children())}{
				for children={
					if={equal(n,(n_children("!u")+1)/2)}{calign with current}{}
				}
			}{}
		}
		[Parallel Text \\ Generation, draw=gray, fill=gray!15, text width=1.5cm, text=black
		[AR-Based\\(\cref{sec:AR}), color=brightlavender, fill=brightlavender!15, text width=2.6cm, text=black
		[Draft and Verifying\\(\cref{sec:dav}), color=lightgreen, fill=lightgreen!15, text width=3.0cm, text=black
                [{
    ADED~\cite{liu2024aded}, 
    BiLD~\cite{kim2023bild}, 
    Block Verification~\cite{sun2025block}, 
    CS Drafting~\cite{chen2023cascade}, 
    DDD~\cite{brown2024ddd}, 
    DistillSpec~\cite{zhou2023distillspec}, 
    Draft\&Verify~\cite{zhang2023draft}, 
    DSBD~\cite{qin2024dsbd}, 
    DySpec~\cite{xiong2024dyspec}, 
    EAGLE~\cite{li2024eagle}, 
    EAGLE2~\cite{li2024eagle2}, 
    EESD~\cite{liu2024eesd}, 
    Falcon~\cite{gao2024falcon}, 
    Fast Inference~\cite{leviathan2023fast}, 
    GSD~\cite{gong2024gsd}, 
    HASS~\cite{zhang2024hass}, 
    Hydra~\cite{ankner2024hydra}, 
    Judge~\cite{bachmann2025judge}, 
    Kangaroo~\cite{liu2024kangaroo}, 
    LayerSkip~\cite{elhoushi2024layerskip}, 
    Medusa~\cite{cai2024medusa}, 
    Mixture of Attentions~\cite{zimmer2024mixture}, 
    MTAD~\cite{qin2024mtad}, 
    ON-THE-FLY~\cite{liu2025dropin}, 
    OPT-Tree~\cite{wang2024opttree}, 
    Ouroboros~\cite{zhao2024ouroboros}, 
    OSD~\cite{liu2023online}, 
    PaSS~\cite{monea2023pass}, 
    PEARL~\cite{liu2024pearl}, 
    PipeInfer~\cite{butler2024pipeinfer}, 
    PPD~\cite{yang2023predictive}, 
    ProPD~\cite{zhong2024propd}, 
    REST~\cite{he2023rest}, 
    RSD~\cite{jeon2024rsd}, 
    Sequoia~\cite{chen2024sequoia}, 
    SPACE~\cite{yi2024space}, 
    SpecDec~\cite{xia2023speculative}, 
    SpecDec++~\cite{huang2024specdec++}, 
    SpecInfer~\cite{miao2023specinfer}, 
    SpecTr~\cite{sun2023spectr}, 
    SPEED~\cite{hooper2023speed}, 
    SWIFT~\cite{xia2024swift},
    CAS-Spec~\cite{ningcas},
    STree~\cite{wu2025stree},
    GRIFFIN~\cite{hu2025griffin},
    LOOKAHEAD REASONING~\cite{fu2025scaling}
}, color=lightgreen, fill=lightgreen!15, text width=5.0cm, text=black]
            ]
		[Decomposition and Fill\\(\cref{sec:sfg}), color=lightcoral, fill=lightcoral!15, text width=3.0cm, text=black
                [{PARALLELPROMPT~\cite{kolawole2025parallelprompt}, storytelling~\cite{yao2019plan}, WritingPath~\cite{lee2025navigating}, SoT~\cite{ning2023skeleton}, SPRINT~\cite{biju2025sprint}}, color=lightcoral, fill=lightcoral!15, text width=5.0cm, text=black]
            ]
		[Multiple Token Prediction\\(\cref{sec:mtp}), color=harvestgold, fill=harvestgold!15, text width=3.0cm, text=black
                [{L-MTP~\cite{liu2025mtp}, WHS-MTP~\cite{mehra2025multi},
                Medusa~\cite{cai2024medusa}, MuToR ~\cite{gerontopoulos2025multi},
                Blockwise Parallel Decoding~\cite{stern2018blockwise}, PaSS~\cite{monea2023pass},
                EAGLE~\cite{li2024eagle}, Gated LoRA MTP~\cite{samragh2025your}, ProphetNet~\cite{qi2020prophetnet}, Meta MTP~\cite{gloeckle2024better},
                DeepSeek-V3~\cite{liu2024deepseek}, MiMo~\cite{xiaomi2025mimo}}, color=harvestgold, fill=harvestgold!15, text width=5.0cm, text=black]
            ]
		]
		[Non-AR-Based\\(\cref{sec:Non-AR}), color=capri, fill=capri!15, text width=2.6cm, text=black
		[One-Shot Generation\\(\cref{sec:osg}), color=lightgreen, fill=lightgreen!15, text width=3.0cm, text=black
                [{Fertility-based NAR generation~\cite{gu2017non},
CTC~\cite{libovicky2018end},
ELMER~\cite{lee2018deterministicnonautoregressiveneuralsequence},
LAVA~\cite{li2020lava}, 
AligNART~\cite{song2021alignart}, 
Syntax-guided NAR translation~\cite{ran2021guiding}, 
Non-Monotonic NAR model~\cite{shao2022non}, 
DePA ~\cite{zhan2023depaimprovingnonautoregressivemachine}, 
DA-Transformer~\cite{huang2022directed}, 
Viterbi Decoding for DA-Transformer~\cite{shao2022viterbi}, 
Fully NAR with Dependency Modeling~\cite{gu2020fullynonautoregressiveneuralmachine}, 
Ratio-first~\cite{su2021nonautoregressivetextgenerationpretrained},
AXE loss~\cite{ghazvininejad2020aligned},
Ngram-OAXE loss~\cite{du2022ngram},
Multi-Granularity Optimization~\cite{li2022multi},
DDRS~\cite{shao2022one} 
}, color=lightgreen, fill=lightgreen!15, text width=5.0cm, text=black]
            ]
		[Masked Generation\\(\cref{sec:mg}), color=harvestgold, fill=harvestgold!15, text width=3.0cm, text=black
                [{
                Uniformization~\cite{chen2023accelerating}, 
                FHS~\cite{zheng2024masked}, Gillespie~\cite{campbell2022continuous},
                $\tau$-leapping~\cite{lou2024discrete,campbell2022continuous,shi2024simplified,seed2025diffusion,zhang2025target},
                Tweedie $\tau$-leaping~\cite{lou2024discrete,sun2022score},
                Fast-dLLM~\cite{wu2025fast},
                LLaDA~\cite{nie2025large},
                MGDM~\cite{ye2024beyond}, RDM~\cite{zheng2023reparameterized},
                TWPB~\cite{kim2025train},
                EB-sampler~\cite{ben2025accelerated}, 
                Fast-dLLM~\cite{wu2025fast}, SlowFast~\cite{wei2025accelerating},
                CT-MDM~\cite{campbell2022continuous}, ReMDM~\cite{wang2025remasking},
                MD4~\cite{shi2024simplified},
                P2~\cite{peng2025path}, 
                DDPD~\cite{liu2024think}, RDM~\cite{zheng2023reparameterized},
                APD~\cite{israel2025accelerating},
                ASSD~\cite{guo2025reviving},
                dkv-cache~\cite{ma2025dkv},
                FreeCache~\cite{hu2025accelerating},
                Eso-LMs~\cite{sahoo2025esoteric},
                SDTT~\cite{deschenaux2024beyond}, 
                CLLM~\cite{kou2024cllms}, Duo~\cite{sahoo2025diffusion},
                d1-LLaDA~\cite{zhao2025d1},
                LLaDA-1.5~\cite{zhu2025llada},
                DiffCoder~\cite{gong2025diffucoder},
                Seed Diffusion~\cite{seed2025diffusion},
                DiffLLaMA~\cite{gong2024scaling},
                Dream~\cite{dream2025},
                DiffPO~\cite{chen2025diffpo},
                WINO~\cite{hong2025wide},
                LLaDA-2.0~\cite{bie2025llada2},
                LLaDA-MOE~\cite{zhu2025llada},
                DBA~\cite{prabhudesai2025diffusion},
                d2~\cite{wang2025d2},
                wd1~\cite{tang2025wd1},
                d-TreeRPO~\cite{pan2025d},
                SAPO~\cite{sapo},
                GDPO~\cite{gdpo},
                ESPO~\cite{espo},
                DreamCoder~\cite{xie2025dream},
                DreamOn~\cite{wu2026dreamon},
                SPG~\cite{wang2025spg},
                TraceRL~\cite{wang2025revolutionizing},
                SDAR~\cite{cheng2025sdar},
                WeDLM~\cite{liu2025wedlm},
                SBD~\cite{gat2025set},
                Fast-DLLM-v2~\cite{wu2025fast},
                Prophet~\cite{prophet},
            CreditDecoding~\cite{wang2025creditdecoding},
                D3LLM~\cite{qian2026d3llm},
                dParallel~\cite{chen2025dparallel},
                D2F~\cite{d2f}
                }, color=harvestgold, fill=harvestgold!15, text width=5.0cm, text=black]
            ],
		[Edit-Based Refinement\\(\cref{sec:ebr}), color=lightcoral, fill=lightcoral!15, text width=3.0cm, text=black
                [{Insertion Transformer~\cite{stern2019insertion},
Levenshtein Transformer~\cite{gu2019levenshtein}, EDITOR~\cite{xu2021editor}, FELIX~\cite{mallinson2020felix}, 
LevOCR~\cite{da2022levenshteinocr}, FastCorrect~\cite{leng2022fastcorrectfasterrorcorrection}, 
Latent CTC~\cite{zhang2023non}, RL for LT~\cite{wang-etal-2024-reinforcement}, EditKSum~\cite{liang2024summarizing},
Deterministic NAR~\cite{lee-etal-2018-deterministic}, 
FlowSeq~\cite{ma2019flowseq}, LaNMT~\cite{shu2020latent},
Latent Space Refinement~\cite{lee2020iterative},
Auxiliary Regularization~\cite{wang2019non},  
Imitation Learning~\cite{wei2019imitation, agrawal2022imitation},
CRF-NAT~\cite{sun2019fast}, EM Framework~\cite{sun2020approach},
Imputer~\cite{chan2020imputer},Align-Refine~\cite{chi2020alignrefinenonautoregressivespeechrecognition},RewriteNAT~\cite{geng2021learning},RenewNAT~\cite{guo2023renewnat}, RecoverSAT~\cite{ran2020learning}, HRT~\cite{wang2022hybrid},
LLM Self-Correction~\cite{chen2024iterativetranslationrefinementlarge}, IterGen~\cite{ugare2025itergeniterativesemanticawarestructured},
KD Rejuvenation~\cite{ding2021rejuvenating, ding2021understandingimprovinglexicalchoice}, 
SlotRefine~\cite{wu2020slotrefine}, DST~\cite{le2020non}}, color=lightcoral, fill=lightcoral!15, text width=5.0cm, text=black]
            ]
		]
		]
	\end{forest}
	\caption{Taxonomy of parallel text generation methods}
	\label{fig: overview-taxonomy}
\end{figure}

Building on the background analysis and definitions above, we categorize existing work on parallel text generation according to the taxonomy illustrated in Figure~\ref{fig: overview-taxonomy}.

We first divide parallel text generation methods into two broad categories: \textbf{AR-based} and \textbf{Non-AR-based}, based on their training objectives. Specifically, a method is considered AR-based if, during training, it preserves a left-to-right semantic dependency—i.e., each token is generated conditioned on previous ones, even if multiple tokens may be produced per inference step at decoding time. In contrast, Non-AR-based methods break this semantic dependency during training by treating token generation as conditionally independent or structuring it differently (e.g., through denoising or diffusion), thereby enabling inherently parallel decoding.

Within the \textbf{AR-based} category, we identify three representative subtypes:

\begin{itemize}
  \item \textbf{Draft-and-Verifying}: These methods, such as speculative decoding, first generate a full or partial draft using a lightweight model, and then verify or correct it using a stronger model. This allows speculative parallel decoding while maintaining autoregressive consistency.

  \item \textbf{Decomposition-and-Fill}: These methods first decompose the generation task into semantically or structurally coherent components—such as outlines, key phrases, or prompt segments—and then generate the full text by filling in each component, potentially in parallel. This two-stage process improves generation efficiency by enabling partial parallelism, while still preserving autoregressive coherence within each individual segment.
  
  \item \textbf{Multiple Token Prediction}: Instead of emitting one token per step, these approaches predict multiple future tokens in parallel within an AR framework. To enhance the reliability of the generated outputs, they are often integrated with draft-and-verifying mechanisms.
\end{itemize}

On the other hand, the \textbf{Non-AR-based} category includes three major paradigms:

\begin{itemize}
  \item \textbf{One-Shot Generation}: All tokens are generated in a single forward pass, typically using models like the Non-Autoregressive Transformer (NAT). While this approach offers maximum parallelism, it often suffers from lower generation quality due to the absence of sequential dependencies.

  \item \textbf{Masked Generation}: These methods iteratively mask and predict subsets of tokens in parallel. Representative techniques include masked language modeling, denoising autoencoders, and more recently, diffusion-based text generation models that have gained significant attention for their flexibility and generation quality.

  \item \textbf{Edit-Based Refinement}: These approaches treat generation as an iterative editing process over an initial draft, allowing insertion, deletion, or substitution operations to refine the output. Examples include the Insertion Transformer and Levenshtein Transformer, which balance flexibility and quality while enabling partial parallelism.
\end{itemize}
\section{AR-Based} \label{sec:AR}

\subsection{Draft-and-Verifying} \label{sec:dav}

To address the high-latency issue inherent in the AR decoding of LLMs, the research community has explored various parallel decoding strategies~\cite{khoshnoodi2024comprehensive}. The core idea of these methods is to reduce the total number of memory-bandwidth-constrained sequential decoding steps by performing more parallel computations in a single decoding step~\cite{xia2024unlocking}. These efforts can be roughly categorized into a unified \textbf{Draft-and-Verify} paradigm. Under this paradigm, an efficient mechanism first ``drafts'' one or more candidate sequences of future tokens, which are then ``verified'' in parallel by the \textit{target model}—achieving acceleration with little or no sacrifice in generation quality~\cite{zhang2023draft}.

\begin{figure}[htbp]
\centerline{
\includegraphics[width=0.7\linewidth]{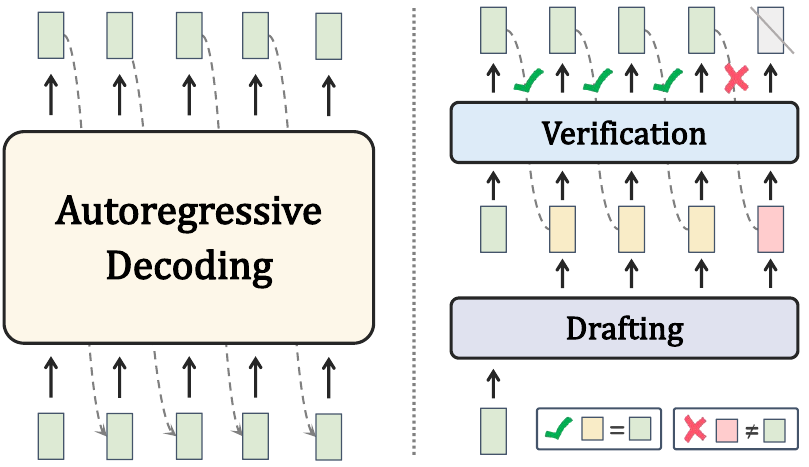}
}
\caption{Comparison of draft-and-verify with autoregressive decoding. Autoregressive decoding generates tokens one by one in an autoregressive manner, resulting in an unsatisfactory decoding speed. In contrast, speculative decoding employs a more efficient model as a drafter to rapidly generate tokens, which are then verified by the target model. High-quality tokens are accepted while low-quality ones are discarded, thus achieving a form of parallelized generation.}
\label{draft_and_verify_teaser}
\end{figure}

Speculative Decoding operates as a \textbf{Draft-and-Verify} paradigm. At each decoding step, it first efficiently drafts multiple future tokens and then verifies them in parallel using the \textit{target model} $\mathcal{M}_q$ to accelerate inference. Below, we formalize the two core substeps:

\paragraph{\textbf{Drafting}}
Given the current input sequence $x_{1}, \ldots, x_{t}$ and the \textit{target model} $\mathcal{M}_q$, an efficient \textit{draft model} $\mathcal{M}_p$ (e.g., a smaller language model (LM)) generates $K$ speculated tokens. Formally:
\begin{align}
p_{1}, \ldots, p_{K} &= \textsc{Draft}\left( x_{\leq t}, \mathcal{M}_{p} \right), \label{eq:draft-prob} \\
\widetilde{x}_{i} &\sim p_{i}, \quad i = 1, \ldots, K, \label{eq:draft-sample}
\end{align}
where $\textsc{Draft}(\cdot)$ denotes the drafting strategy, $p_i$ is the conditional probability distribution from $\mathcal{M}_p$, and $\widetilde{x}_i$ is the $i$-th drafted token sampled from $p_i$.

\paragraph{\textbf{Verification}}
The drafted tokens $\widetilde{x}_{1}, \ldots, \widetilde{x}_{K}$ are verified in parallel by $\mathcal{M}_q$. The \textit{target model} computes $K+1$ distributions:
\begin{equation}
q_{i} = \mathcal{M}_{q} \left( x \mid x_{\leq t}, \widetilde{x}_{<i} \right), \quad i=1, \ldots, K+1. \label{eq:verify-compute}
\end{equation}
Each token $\widetilde{x}_i$ is validated via a criterion $\textsc{Verify}(\widetilde{x}_{i}, p_{i}, q_{i})$.
\begin{itemize}
    \item \textbf{Acceptance}: If the criterion is satisfied, $\widetilde{x}_i$ is output as $x_{t+i}$.
    \item \textbf{Rejection}: At the first position $c$ where verification fails, the token $x_{t+c}$ is resampled via a correction strategy $\textsc{Correct}(p_c, q_c)$ (e.g., $x_{t+c} \leftarrow \arg\max q_c$), and all subsequent drafted tokens are discarded.
    \item \textbf{Continuation}: If all $K$ tokens pass verification, an additional token is sampled from the final distribution: $x_{t+K+1} \sim q_{K+1}$.
\end{itemize}

\paragraph{\textbf{Objective and Acceleration}}
The ultimate goal of this paradigm is to increase the number of tokens accepted per unit of time (i.e., throughput), thereby reducing overall generation latency. Let $L(\mathcal{M})$ denote the latency of a single forward pass for a model $\mathcal{M}$. In a single Speculative Decoding step, the total latency is the sum of drafting and verification, $L(\mathcal{M}_p) + L(\mathcal{M}_q)$, while the number of accepted tokens is $A$. The objective is to maximize the expected throughput:

\begin{equation}
\max \mathbb{E}\left[ \frac{A}{L(\mathcal{M}_p) + L(\mathcal{M}_q)} \right] \label{eq:throughput}
\end{equation}

Achieving this maximization hinges on balancing two competing factors:
\begin{itemize}
    \item \textbf{Speculation Accuracy:} This focuses on maximizing the numerator, $\mathbb{E}[A]$. It requires the \textit{draft model} $\mathcal{M}_p$ to generate token sequences with a high probability of being validated by the \textit{target model} $\mathcal{M}_q$.
    \item \textbf{Drafting Efficiency:} This aims to minimize the denominator, particularly the drafting latency $L(\mathcal{M}_p)$. The drafting process must be substantially more lightweight than a full forward pass of $\mathcal{M}_q$.
\end{itemize}
These two objectives often involve a fundamental trade-off: a more powerful (and potentially more accurate) \textit{draft model} may incur higher latency, reducing overall speedup. Consequently, a core challenge in designing Speculative Decoding systems is to strike an optimal balance between the \textit{drafter}’s predictive power and its computational cost.

Overall, \textbf{Parallelism} is achieved when the throughput from Equation~\ref{eq:throughput} exceeds that of standard autoregressive decoding:
$$
\mathbb{E}\left[ \frac{A}{L(\mathcal{M}_p) + L(\mathcal{M}_q)} \right] > \frac{1}{L(\mathcal{M}_q)}
$$
Given that the \textit{draft model} is lightweight ($L(\mathcal{M}_p) \ll L(\mathcal{M}_q)$), this condition simplifies to requiring the expected number of accepted tokens, $\mathbb{E}[A]$, to be greater than one.

\paragraph{\textbf{Literature Selection}}

To systematically trace the evolution of Speculative Decoding, this survey reviews the pivotal works that have shaped its development. Our literature selection focuses on high-impact research, with the vast majority of the nearly 50 papers analyzed originating from top-tier machine learning conferences, including NeurIPS, ICML, and ICLR. Our inclusion criteria prioritized three categories of work: (1) foundational papers that introduced the core concepts of speculative decoding; (2) studies that achieved significant breakthroughs in decoding efficiency or accuracy; and (3) representative works that explored novel variants or new application domains of the technique. Through this curated selection, we aim to provide a comprehensive and structured overview of the field's trajectory and current frontiers.

\begin{figure}[htbp]
	\centering
	\tikzset{
		my node/.style={
			draw,
			align=center,
			thin,
			text width=1.2cm, 
			rounded corners=3,
		},
		my leaf/.style={
			draw,
			align=center,
			thin,
			text width=8.5cm, 
			rounded corners=3,
		}
	}
	\forestset{
		every leaf node/.style={
			if n children=0{#1}{}
		},
		every tree node/.style={
			if n children=0{minimum width=1em}{#1}
		},
	}
	\begin{forest}
		nonleaf/.style={font=\bfseries\scriptsize},
		for tree={%
			every leaf node={my leaf, font=\scriptsize},
			every tree node={my node, font=\scriptsize, l sep-=4.5pt, l-=1.pt},
			anchor=west,
			inner sep=2pt,
			l sep=10pt, 
			s sep=3pt, 
			fit=tight,
			grow'=east,
			edge={ultra thin},
			parent anchor=east,
			child anchor=west,
			if n children=0{}{nonleaf}, 
			edge path={
				\noexpand\path [draw, \forestoption{edge}] (!u.parent anchor) -- +(5pt,0) |- (.child anchor)\forestoption{edge label};
			},
			if={isodd(n_children())}{
				for children={
					if={equal(n,(n_children("!u")+1)/2)}{calign with current}{}
				}
			}{}
		}
		[Draft and Verify, draw=gray, fill=gray!15, text width=1.5cm, text=black
		[Speeding up Draft and Verifying\\(\cref{sec:edav}), color=brightlavender, fill=brightlavender!15, text width=2.6cm, text=black
		[Efficient Drafting\\(\cref{sec:ed}), color=lightgreen, fill=lightgreen!15, text width=3.0cm, text=black
                [
                {SpecTr~\cite{sun2023spectr}, CS Drafting~\cite{chen2023cascade},\\
                Draft\&Verify~\cite{zhang2023draft}, SWIFT~\cite{xia2024swift},\\
                 LayerSkip~\cite{elhoushi2024layerskip}, Kangaroo~\cite{liu2024kangaroo},\\
                 SPEED~\cite{hooper2023speed}, Medusa~\cite{cai2024medusa},\\
                 Hydra~\cite{ankner2024hydra}, EAGLE~\cite{li2024eagle}, \\
                Falcon~\cite{gao2024falcon}, CAS-Spec~\cite{ningcas}
                },
                color=lightgreen, fill=lightgreen!15, text width=4.0cm, text=black
                ]
            ]
		[Efficient Verification\\(\cref{sec:ev}), color=harvestgold, fill=harvestgold!15, text width=3.0cm, text=black
                [
                {SpecInfer~\cite{miao2023specinfer}, Sequoia~\cite{chen2024sequoia},\\
                 OPT-Tree~\cite{wang2024opttree}, DySpec~\cite{xiong2024dyspec},\\
                 EAGLE2~\cite{li2024eagle2}, DDD~\cite{brown2024ddd},\\
                 GSD~\cite{gong2024gsd}, ADED~\cite{liu2024aded}, \\
                 STree~\cite{wu2025stree}
                 }
                ,color=harvestgold, fill=harvestgold!15, text width=4.0cm, text=black]
            ]
		[Efficient Pipeline\\(\cref{sec:ep}), color=lightcoral, fill=lightcoral!15, text width=3.0cm, text=black
                [
                {PPD~\cite{yang2023predictive}, PipeInfer~\cite{butler2024pipeinfer},\\
                 PaSS~\cite{monea2023pass}, CS Drafting~\cite{chen2023cascade},\\
                 PEARL~\cite{liu2024pearl}, Ouroboros~\cite{zhao2024ouroboros},\\
                 SPACE~\cite{yi2024space}}
                , color=lightcoral, fill=lightcoral!15, text width=4.0cm, text=black]
            ]
		]
		[Improving Acceptance Rate\\(\cref{sec:adav}), color=capri, fill=capri!15, text width=2.6cm, text=black
		[Accurate Drafting\\(\cref{sec:ad}), color=lightgreen, fill=lightgreen!15, text width=3.0cm, text=black
                [
                {SpecDec~\cite{xia2023speculative}, OSD~\cite{liu2023online},\\
                 DistillSpec~\cite{zhou2023distillspec}, Medusa~\cite{cai2024medusa},\\
                 EAGLE~\cite{li2024eagle}, Falcon~\cite{gao2024falcon}, HASS~\cite{zhang2024hass},\\
                 Mixture of Attentions~\cite{zimmer2024mixture}, SpecDec++~\cite{huang2024specdec++},\\
                 BiLD~\cite{kim2023bild}, ON-THE-FLY~\cite{liu2025dropin},\\
                 EESD~\cite{liu2024eesd}, Judge~\cite{bachmann2025judge}, GRIFFIN~\cite{hu2025griffin}}
                , color=lightgreen, fill=lightgreen!15, text width=4.0cm, text=black]
            ]
		[Accurate Verification\\(\cref{sec:av}), color=harvestgold, fill=harvestgold!15, text width=3.0cm, text=black
                [
                {SpecDec~\cite{xia2023speculative}, Draft\&Verify~\cite{zhang2023draft},\\
                 Fast Inference~\cite{leviathan2023fast}, Block Verification~\cite{sun2025block},\\
                 MTAD~\cite{qin2024mtad}, Medusa~\cite{cai2024medusa},\\
                 EAGLE~\cite{li2024eagle}, DSBD~\cite{qin2024dsbd}, REST~\cite{he2023rest},\\
                 ProPD~\cite{zhong2024propd}, RSD~\cite{jeon2024rsd}, \\
                 LOOKAHEAD REASONING~\cite{fu2025scaling}
                 }
                , color=harvestgold, fill=harvestgold!15, text width=4.0cm, text=black]
            ]
		]
		]
	\end{forest}
	\caption{Taxonomy of draft and verifying methods}
	\label{fig: taxonomy}
\end{figure}
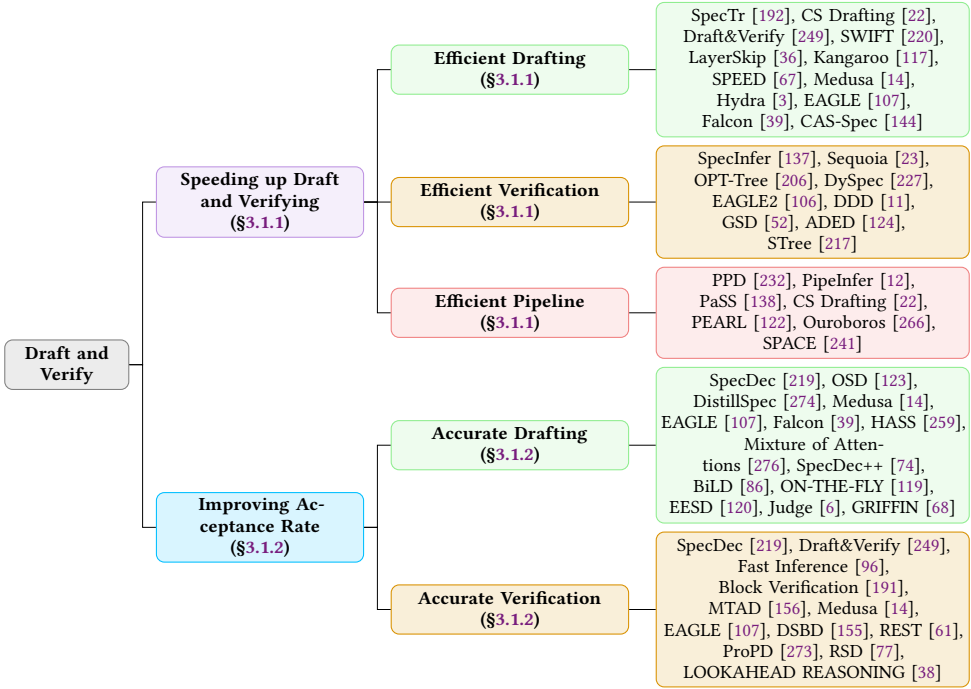

\subsubsection{Speeding up Draft and Verifying} \label{sec:edav}
As established in the objective function (Equation~\ref{eq:throughput}), acceleration is determined by the ratio of accepted tokens to total latency. This section, ``Speeding up Draft and Verifying,'' focuses on optimizing the denominator of this ratio: the cost. The core principle is to minimize the latency incurred during both the drafting and verification stages, thereby creating a more efficient backbone for the process. We will explore three primary strategies to achieve this: (1) constructing highly lightweight \textit{drafters} to reduce drafting latency, (2) designing advanced verification structures like token trees to maximize parallelism, and (3) implementing pipelined execution flows to overlap computation and minimize idle time. By reducing the fundamental cost of each decoding step, these methods lay the groundwork for substantial speedups.

\paragraph{\textbf{Efficient Drafting}} \label{sec:ed}
As the first strategy for minimizing latency, \texttt{Efficient Drafting} targets the cost of the drafter itself, $L(\mathcal{M}p)$. The goal is to make the drafting process substantially faster than a forward pass of the \textit{target model}, thereby reducing one of the two key components of the latency denominator. Research in this area focuses on designing efficient \textit{drafters}, either by employing a separate, smaller model or by utilizing the \textit{target model} in a more efficient manner.

The most direct method for creating an efficient drafter is to employ a separate, smaller language model, typically from the same family as the \textit{target model}. For instance, a smaller \textit{Llama} can be used to accelerate a larger one~\cite{chen2023accelerating, spector2023staged}. The primary advantage of this approach is its simplicity; it requires no additional training or architectural modifications, allowing for rapid adoption across various pre-trained models. Works such as \textit{SpecTr}~\cite{sun2023spectr} and \textit{CS Drafting}~\cite{chen2023cascade} have validated this paradigm, demonstrating its effectiveness for accelerating large language model inference.

\textit{Self-drafting} techniques offer an alternative to two-model systems by leveraging the \textit{target model} itself. The core strategy involves a "shallow" forward pass—executing only a subset of layers—to rapidly generate draft tokens. These candidates are then verified by a full forward pass of the original, unmodified model, significantly reducing latency while reusing parameters. Implementations of this concept vary. For instance, \textit{Draft\&Verify}~\cite{zhang2023draft} and \textit{SWIFT}~\cite{xia2024swift} adaptively skip layers, while \textit{LayerSkip}~\cite{elhoushi2024layerskip} enables early exiting from intermediate layers based on confidence, drawing on early-exit mechanisms~\cite{teerapittayanon2016branchynet}. Similarly, \textit{Kangaroo}~\cite{liu2024kangaroo} uses a shallow sub-network with a lightweight adapter to balance efficiency and accuracy, and \textit{SPEED}~\cite{hooper2023speed} further optimizes by pipelining computations to process subsequent tokens in parallel. 
\textit{CAS-Spec}~\cite{ningcas} derives a cascade of fast self-drafters from the target model and adaptively selects draft depth during inference.

Another highly efficient self-drafting strategy augments the \textit{target model} with lightweight prediction heads, making the drafting phase nearly instantaneous. This is achieved by avoiding re-computation of the expensive transformer layers; after an initial forward pass generates one token, its hidden state is fed to these auxiliary heads to rapidly produce a sequence of draft tokens. \textit{Medusa}~\cite{cai2024medusa} pioneered this with multiple non-autoregressive heads. Other examples include \textit{Hydra}~\cite{ankner2024hydra}, which uses independent heads for parallel drafting, as well as \textit{EAGLE}~\cite{li2024eagle} and \textit{Falcon}~\cite{gao2024falcon}.

\paragraph{\textbf{Efficient Verifying}} \label{sec:ev}
Complementing the efforts to reduce drafting latency, \texttt{Efficient Verifying} focuses on the second term in the latency denominator, the verification cost $L(\mathcal{M}_q)$. While the latency of a single forward pass through the \textit{target model} is largely fixed, its efficiency can be dramatically improved by increasing the number of tokens processed in parallel within that single step. The primary goal is therefore to design verification structures, such as token trees, that maximize this parallelism, allowing the \textit{target model} to evaluate a large batch of tokens simultaneously and thus reducing the effective latency per generated token.

A key strategy to increase verification throughput is evolving the verification structure from a linear path to a parallel, multi-branch format. A significant advancement is \textbf{tree-based verification}, which merges multiple candidate sequences with common prefixes into a single \textbf{token tree}. The \textit{target model} then processes this tree in one forward pass, using a specialized \textbf{tree attention mask} to maintain causal dependencies during parallel computation. The pioneering work of \textit{SpecInfer}~\cite{miao2023specinfer} first demonstrated the feasibility and efficiency of this scheme. Extending tree-based verification beyond Transformers, \textit{STree}~\cite{wu2025stree} enables scalable token-tree verification for hybrid state-space models via a packed-tree execution that reuses state-transition computations.

Subsequent research has focused on optimizing the tree's geometry to balance potential gains against computational cost. Some approaches treat this as a static optimization problem: \textit{Sequoia}~\cite{chen2024sequoia} uses a hardware-aware optimizer to determine tree dimensions, while \textit{OPT-Tree}~\cite{wang2024opttree} maximizes the expected number of accepted tokens. Others employ dynamic, on-the-fly adjustments: \textit{DySpec}~\cite{xiong2024dyspec} expands the tree based on the \textit{draft model}'s confidence, whereas \textit{EAGLE2}~\cite{li2024eagle2} and \textit{DDD}~\cite{brown2024ddd} introduce context-aware construction and dynamic depth decoding, respectively. The framework has also inspired more complex structures, such as the graph-based methods of \textit{GSD}~\cite{gong2024gsd} and the adaptive depth mechanisms in \textit{ADED}~\cite{liu2024aded} for more intricate dependency modeling.

\paragraph{\textbf{Efficient Pipeline}} \label{sec:ep}
The third strategy for efficiency, \texttt{Efficient Pipeline}, addresses the sequential nature of the latency, $L(\mathcal{M}_p) + L(\mathcal{M}_q)$. Standard Speculative Decoding (SD) incurs this cost sequentially, creating a bottleneck where one model is idle while the other works. Pipelined approaches tackle this by overlapping the drafting and verification stages. The idea is to execute them concurrently, hiding the drafting latency $L(\mathcal{M}_p)$ behind the verification latency $L(\mathcal{M}_q)$, thereby maximizing resource utilization and minimizing the wall-clock time per decoding step.

A foundational approach, \textit{Predictive Pipelined Decoding (PPD)}~\cite{yang2023predictive}, accelerates greedy decoding by using surplus capacity to speculatively pipeline the generation of subsequent tokens. It guarantees output that is identical to standard greedy decoding, ensuring acceleration without altering results. Building on this concept, \textit{PipeInfer}~\cite{butler2024pipeinfer} integrates pipelining with speculative decoding. It enables the target model's inference to run concurrently with multiple speculative generations on auxiliary models and features an early cancellation mechanism for invalid branches. This design reduces inter-token latency, improves system utilization, and enhances resilience to low acceptance rates, achieving speedups over standard speculative inference.

Other methods create structured or lookahead-based parallel drafting mechanisms. For instance, \textit{Parallel Speculative Sampling (PaSS)}~\cite{monea2023pass} shortens critical path latency by preparing future token embeddings in advance. Similarly, \textit{Cascade Speculative Drafting (CS Drafting)}~\cite{chen2023cascade} organizes drafting into rigid cascade structures for efficient parallel execution. More advanced strategies introduce dynamic adjustments: \textit{PEARL}~\cite{liu2024pearl} employs adaptive draft lengths and multi-stage verification to maximize efficiency; \textit{Ouroboros}~\cite{zhao2024ouroboros} generates longer, coherent drafts on a "phrase by phrase" basis; and \textit{SPACE}~\cite{yi2024space} uses a "smart parallel auto-correct decoding" scheme to efficiently correct draft errors in parallel, improving overall throughput.

\subsubsection{Improving Acceptance Rate} \label{sec:adav}
While minimizing computational latency is crucial, an efficient but inaccurate system will fail to deliver meaningful acceleration. This section, ``Improving Acceptance Rate,'' shifts the focus to the numerator of the throughput equation (Equation~\ref{eq:throughput}): maximizing the expected number of accepted tokens, $\mathbb{E}[A]$. The central goal here is to enhance the quality and alignment of the drafted tokens so that they are more likely to be validated by the \textit{target model}. A high acceptance rate is the most direct way to increase the number of tokens generated per decoding step. We will examine two complementary approaches: (1) improving the predictive power of the \textit{drafter} through specialized training and dynamic adaptation, and (2) refining the verification logic to be more flexible and less conservative, thereby increasing the probability of accepting valid tokens. Ultimately, these methods aim to ensure that the speculative work performed is not wasted, directly boosting the overall generation throughput.

\paragraph{\textbf{Accurate Drafting}} \label{sec:ad}
To maximize the expected number of accepted tokens, $\mathbb{E}[A]$, the direct approach is to improve the quality of the drafts at their source. \texttt{Accurate Drafting} focuses on ensuring that the \textit{draft model}'s predictions align as closely as possible with the \textit{target model}. A more accurate \textit{drafter} leads to a higher acceptance rate, reducing wasted computation and boosting throughput. Research in this area improves this alignment through two main strategies: statically enhancing the \textit{drafter} via specialized training or dynamically adapting its behavior during inference.

The first approach, static alignment, improves accuracy by explicitly training or fine-tuning a \textit{drafter} \textit{before} inference to better mimic the \textit{target model}'s distribution. 
Initial approaches focused on specialized drafter models, pioneered by \textit{SpecDec}~\cite{xia2023speculative} with its lightweight transformer architecture. Subsequent efforts aimed to improve drafter-target alignment, primarily through knowledge distillation techniques~\cite{liu2023online, zhou2023distillspec}. 
To mitigate train--test token misalignment that degrades draft quality, \textit{GRIFFIN}~\cite{hu2025griffin} introduces token-alignable training via top-$k$ loss masking and a token-guided draft architecture, improving draft--target alignment and thus acceptance length.
Other works enhanced draft quality via more sophisticated generation strategies, such as optimizing partial forward passes within the drafter~\cite{zhang2023draft}.
Another paradigm involves adding auxiliary heads to the target model. This approach was pioneered by \textit{Medusa}~\cite{cai2024medusa} with multiple non-autoregressive heads trained atop a frozen model, and later advanced by \textit{EAGLE}~\cite{li2024eagle} with a more coherent auto-regressive head that reuses the model's features. Further research has focused on enhancing draft quality through more sophisticated head designs, such as semi-autoregressive frameworks, knowledge distillation, and diverse attention mechanisms~\cite{gao2024falcon, zhang2024hass, zimmer2024mixture}.

In contrast to static alignment, dynamic methods increase accuracy by adapting the drafting process \textit{during} inference. This allows the system to adjust its behavior based on real-time feedback, such as the \textit{drafter}'s confidence or the historical acceptance rate, thereby optimizing the trade-off between drafting more tokens and ensuring their correctness. 
For instance, some methods dynamically determine the draft length: \textit{SpecDec++}~\cite{huang2024specdec++} adds a prediction head to estimate token acceptance probability; \textit{ON-THE-FLY}~\cite{liu2025dropin} adjusts length based on historical accuracy; and \textit{EESD}~\cite{liu2024eesd} employs control mechanisms like \textit{Thompson Sampling}~\cite{slivkins2019introduction}.
Other approaches control the verification step: \textit{BiLD}~\cite{kim2023bild} invokes the target model only when the drafter's confidence falls below a threshold, while some works introduce special tokens for the draft model to request verification~\cite{liu2025knowledge}.
Furthermore, for specialized domains like mathematics, \textit{Judge}~\cite{bachmann2025judge} relaxes alignment by using a learned verification layer to assess contextual correctness rather than requiring exact token matching.

\paragraph{\textbf{Accurate Verifying}} \label{sec:av}
Beyond improving the draft source, the verification logic itself presents a key opportunity for increasing the acceptance rate. 
\texttt{Accurate Verifying} focuses on optimizing the decision criterion, $\textsc{Verify}(\cdot)$, to be less conservative without compromising the output distribution's integrity. Even if a draft is plausible, an overly strict verification rule can lead to its rejection, reducing $\mathbb{E}[A]$. Therefore, the goal is to design more probabilistic acceptance criteria that can approve a broader range of valid tokens, thereby maximizing the yield from each verification step.

A foundational technique for refining verification logic is linear verification, where a candidate sequence is validated sequentially. 
Early methods like \textit{SpecDec}~\cite{xia2023speculative} and \textit{Draft\&Verify}~\cite{zhang2023draft} used deterministic rejection sampling, accepting a token only if it matched the \textit{target model}'s greedy prediction. This strict matching, however, led to low acceptance rates. To improve this, speculative sampling, introduced in works like \textit{Fast Inference}~\cite{leviathan2023fast} and Chen et al.~\cite{chen2023accelerating}, uses probabilistic validation. It accepts or rejects a token based on its probability ratio under the \textit{target} and \textit{draft models}, significantly boosting acceptance rates while preserving the output distribution. An orthogonal improvement, block verification, evaluates tokens as a collective unit. Methods such as \textit{Block Verification}~\cite{sun2025block} and \textit{MTAD}~\cite{qin2024mtad} assess a draft sequence based on its joint probability. This holistic approach can approve a globally probable sequence despite locally non-optimal tokens, enabling more effective draft use and higher acceptance rates. \textit{LOOKAHEAD REASONING}~\cite{fu2025scaling} performs step-level semantic verification so that semantically correct draft steps are accepted, increasing speculative decoding's acceptance rate in reasoning.

A complementary strategy for boosting acceptance rates is to enhance the quality of the drafted tokens themselves, as better candidates are inherently more likely to pass verification. One approach leverages the \textit{target model}'s architecture: \textit{Medusa}~\cite{cai2024medusa} employs multiple lightweight prediction heads, while \textit{EAGLE}~\cite{li2024eagle} uses a single, more powerful autoregressive head. Other methods use external \textit{drafters}: \textit{DSBD}~\cite{qin2024dsbd} utilizes a smaller \textit{draft model} with \textit{beam search}, and \textit{REST}~\cite{he2023rest} retrieves continuations from a datastore. More advanced strategies intertwine generation with verification, such as \textit{ProPD}'s~\cite{zhong2024propd} progressive refinement and \textit{RSD}'s~\cite{jeon2024rsd} recursive verification. This area has also spurred \textit{Multi-Draft Speculative Decoding (MDSD)}, which optimizes parallel draft verification. For instance, Hu et al.~\cite{hu2025towards} proposed a hybrid sampling strategy, while Khisti et al.~\cite{khisti2024multidraft} used \textit{importance sampling} in a two-phase process to pre-select promising drafts. These ongoing innovations continue to advance the efficiency of speculative decoding.

\subsection{Decomposition-and-Fill}
\label{sec:sfg}

The core idea behind the decomposition-and-fill paradigm is to break down a complex generation task into several independent subtasks, and then execute the generation of each component in parallel. This process typically involves two main stages: 
\begin{itemize}
    \item \textbf{Stage 1: Task Decomposition}, where an LLM or a system breaks the overall task into a structured set of smaller, independent components to be processed in parallel.
    \item \textbf{Stage 2: Parallel Content Filling}, where multiple LLM calls are made concurrently to flesh out the details of each component.
\end{itemize}
This approach not only significantly reduces end-to-end latency by replacing a long sequential decoding process with shorter, parallel ones, but can also enhance the quality and coherence of the final output by enforcing a logical structure from the outset. The decomposition-and-fill paradigm can be broadly divided into two distinct approaches: \textbf{Query-Level Decomposition}, where the system identifies and extracts parallelizable subtasks inherent in the user's initial prompt, and \textbf{Answer Structure Planning}, where the model first generates a high-level outline for its response and then simultaneously executes multiple fill operations in parallel.

\paragraph{\textbf{Query-Level Decomposition}} This approach focuses on identifying and extracting parallelizable subtasks that are already present within the user's query. Many real-world prompts naturally contain multiple independent subtasks, such as requests to analyze a list of items or translate several sentences. A seminal contribution in this domain is PARALLELPROMPT~\cite{kolawole2025parallelprompt}. Through comprehensive analysis of over 37,000 real-world prompts extracted from public chat logs, specifically the LMSYS-Chat-1M~\cite{zheng2023lmsys} and WildChat-1M~\cite{zhao2024wildchat} datasets, the researchers discovered that 10.3\% of prompts contain inherent ``latent semantic parallelism." By extracting a structured schema (containing a task template, shared context, and iterable data) and executing the subtasks concurrently, their evaluation demonstrates significant latency reductions on a variety of tasks. For example, Reading Comprehension tasks show a raw speedup of 5.72×, Repeated Generation achieves 4.39×, and Keyword Extraction gains 2.54×, all with minimal degradation in output quality. This research provides a crucial testbed and justification for building structure-aware execution pipelines that can automatically detect and exploit parallelism in user inputs.

\paragraph{\textbf{Answer Structure Planning}} Whereas Query-Level Decomposition focuses on uncovering existing parallelizable structures within a user's prompt, Answer Structure Planning shifts the focus to such structures in the model's output. In this approach, the LLM first generates a plan—typically a skeleton or outline—that serves as a scaffold for the response. Each point within this scaffold can then be expanded in parallel. 

The foundational concept of separating planning from execution was explored in early work, such as the "plan-and-write" framework for storytelling~\cite{yao2019plan}. This approach first generated a complete storyline outline before filling in the narrative for each section. While the implementation of the fill stage was sequential, this architectural separation introduced the potential for parallelism. More recently, WritingPath~\cite{lee2025navigating} inherited this idea, demonstrating on modern LLMs including GPT-4~\cite{OpenAI2023GPT4TR} that upfront planning could significantly improve the quality and logical coherence of generated text, further validating the benefits of the planning stage itself.

The first work to explicitly realize this potential and enable parallel generation for modern LLMs was Skeleton-of-Thought (SoT)~\cite{ning2023skeleton}. SoT prompts an LLM to first output a concise skeleton of the answer, which consists of several short points (e.g., 3-5 words each). Then, in the point-expanding stage, it uses parallel API calls (for proprietary models) or batched decoding (for open-source models) to have the LLM expand on each skeleton point concurrently. This approach yielded significant latency reductions, achieving speedups of up to 2.39× across 12 different LLMs on benchmarks like Vicuna-80~\cite{chiang2023vicuna}. While SoT could improve answer quality on certain categories (like knowledge or common-sense questions), its reliance on a static, upfront plan made it unsuitable for tasks requiring step-by-step reasoning, such as math or coding, where later steps depend on the results of earlier ones.

To address this limitation, SPRINT~\cite{biju2025sprint} was proposed to enable high-quality parallel generation for reasoning tasks. Unlike SoT's static planning step, SPRINT introduces a dynamic, interleaved "rolling-horizon" process. It employs a planner module that iteratively assesses the current context and generates a set of independent subtasks for the current round. These tasks are then executed in parallel by a pool of executors, and the results are synchronized back to the planner to inform the next cycle of planning. This allows the model to adapt its strategy based on intermediate results, which is a crucial feature for complex problem-solving. The experimental results demonstrate that SPRINT achieves both higher performance and lower latency. In terms of quality, SPRINT achieved a higher accuracy of 92.5\% on the MATH500 benchmark~\cite{lightman2023lets}, outperforming the 91.0\% of the original model. For efficiency, which measures by the reduction in sequential tokens, SPRINT shows significant gains. It reduced the number of sequential tokens by up to 39\% on MATH500~\cite{lightman2023lets}, 45\% on the GPQA-diamond benchmark~\cite{rein2024gpqa}, and 65\% on the Countdown game~\cite{yao2023tree}.

\vspace{5pt}

In summary, the decomposition-and-fill approach represents a task-level parallel generation paradigm that has been experimentally shown to not only accelerate generation but also to maintain or even enhance content quality. However, the efficacy of this paradigm is highly contingent on the task type, proving most beneficial for assignments with low contextual dependency and limited reasoning requirements. For reasoning-heavy tasks characterized by strong sequential dependencies, generation quality can be easily compromised by parallelization. Even with successful attempts like SPRINT, it becomes evident that more complex tasks necessitate a greater reliance on multi-round, iterative planning. This, in turn, introduces new sequential overhead, thereby diminishing the overall efficiency gains promised by parallelization.

\subsection{Multiple Token Prediction} \label{sec:mtp}

Most methods discussed in previous sections operate under the assumption that LLMs are limited to next-token prediction, thus achieving parallel decoding by either employing auxiliary models or partitioning the generation task. This section introduces multi-token prediction (MTP) methods, which enable parallel decoding by allowing an LLM to predict multiple tokens in a single step. Although these models predict the subsequent N tokens at once, they are still considered autoregressive according to our definition in section 2.

The generation process using MTP typically follows a draft-and-verify paradigm, similar to the one described in Section \ref{sec:dav}. This process involves two main steps:

\begin{enumerate}
    \item \textbf{Drafting:} An MTP-enabled LLM generates a sequence of N future tokens simultaneously.
    \item \textbf{Verifying:} The same LLM then validates whether the drafted tokens are acceptable. This validation can be performed using various strategies, such as linear or tree-based verification.
\end{enumerate}

Since the draft-and-verify paradigm has been covered, this section will not focus on the text generation process itself. Instead, it will explore how to equip LLMs with MTP capabilities. These approaches can be categorized based on the development stage at which MTP is introduced: post-training optimization or pre-training integration.

\begin{figure}[htbp]
	\centering
	\tikzset{
		my node/.style={
			draw,
			align=center,
			thin,
			text width=1.2cm, 
			rounded corners=3,
		},
		my leaf/.style={
			draw,
			align=center,
			thin,
			text width=8.5cm, 
			rounded corners=3,
		}
	}
	\forestset{
		every leaf node/.style={
			if n children=0{#1}{}
		},
		every tree node/.style={
			if n children=0{minimum width=1em}{#1}
		},
	}
	\begin{forest}
		nonleaf/.style={font=\bfseries\scriptsize},
		for tree={%
			every leaf node={my leaf, font=\scriptsize},
			every tree node={my node, font=\scriptsize, l sep-=4.5pt, l-=1.pt},
			anchor=west,
			inner sep=2pt,
			l sep=10pt, 
			s sep=3pt, 
			fit=tight,
			grow'=east,
			edge={ultra thin},
			parent anchor=east,
			child anchor=west,
			if n children=0{}{nonleaf}, 
			edge path={
				\noexpand\path [draw, \forestoption{edge}] (!u.parent anchor) -- +(5pt,0) |- (.child anchor)\forestoption{edge label};
			},
			if={isodd(n_children())}{
				for children={
					if={equal(n,(n_children("!u")+1)/2)}{calign with current}{}
				}
			}{}
		}
		[Multi-Token Prediction, draw=gray, fill=gray!15, text width=2.5cm, text=black
		[Acquiring MTP via Post-training, color=brightlavender, fill=brightlavender!15, text width=3.2cm, text=black
                [
                {L-MTP~\cite{liu2025mtp}, WHS-MTP~\cite{mehra2025multi},\\
                Medusa~\cite{cai2024medusa}, MuToR ~\cite{gerontopoulos2025multi},\\
                Blockwise Parallel Decoding~\cite{stern2018blockwise}, PaSS~\cite{monea2023pass},\\
                EAGLE~\cite{li2024eagle}, Gated LoRA MTP~\cite{samragh2025your}},
                color=brightlavender, fill=brightlavender!15, text width=5.0cm, text=black
                ]
            ]
		[Building Native MTP via Pre-training, color=lightgreen, fill=lightgreen!15, text width=3.2cm, text=black
                [
                {ProphetNet~\cite{qi2020prophetnet}, Meta MTP~\cite{gloeckle2024better},\\
                DeepSeek-V3~\cite{liu2024deepseek}, MiMo~\cite{xiaomi2025mimo}},
                color=lightgreen, fill=lightgreen!15, text width=5.0cm, text=black
                ]
            ]
		]
	\end{forest}
	\caption{Taxonomy of Multi-Token Prediction Methods}
	\label{fig: mtp_taxonomy}
\end{figure}
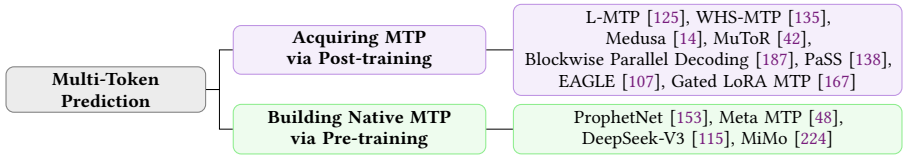

\subsubsection{Acquiring MTP Capability via Post-training Optimization}

To give existing LLMs MTP capabilities without expensive retraining, one approach is to unlock their latent potential through post-training optimization. The core idea is that the internal representations of a pre-trained LLM may already contain information about future tokens \cite{samragh2025your}. To explicitly extract this information, a common technique involves attaching one or more new prediction heads to the LLM's architecture and then fine-tuning only these heads on an MTP objective \cite{mehra2025multi, li2024eagle}. This concept builds on early work that proposed blockwise parallel decoding by training separate heads to predict tokens at different future positions \cite{stern2018blockwise}.

Modern methods refine this approach for speculative decoding. To generate better drafts for verification, PaSS trains a policy network to predict multiple future tokens, which are then sampled in parallel \cite{monea2023pass}. To address the issue of feature uncertainty that arises when multiple heads generate tokens, EAGLE improves speculative sampling by training multiple prediction heads that are conditioned on the same feature vector from the base model \cite{li2024eagle}. Other work focuses on architectural and contextual enhancements. To improve the model's ability to manage information across multiple future steps, one study suggests that MTP requires dedicated "register" tokens to store speculative future states, as standard attention mechanisms may be insufficient \cite{gerontopoulos2025multi}. To further improve prediction accuracy, L-MTP enables the model to predict tokens beyond the immediately adjacent ones by training it to leverage non-adjacent context from the input sequence \cite{liu2025mtp}.

\subsubsection{Building Native MTP Capability via Pre-training}

To build LLMs with inherent MTP capabilities from the ground up, another line of work integrates MTP directly into the pre-training phase. The motivation is that training an LLM to predict multiple tokens from the start can lead to more efficient and powerful LLMs. A foundational approach for this was ProphetNet, which was designed for sequence-to-sequence tasks and introduced a pre-training objective to predict a future n-gram rather than just the next single token \cite{qi2020prophetnet}.

This principle has been adapted for modern decoder-only LLMs. To create models that are both higher quality and faster at inference, researchers have shown that incorporating a multi-token prediction loss during pre-training is highly effective \cite{gloeckle2024better}. This strategy has been successfully implemented in the development of foundation models. To build a powerful and efficient model, the DeepSeek-V3 technical report details the use of an MTP objective as a core component of its pre-training recipe \cite{liu2024deepseek}. Similarly, to unlock more advanced reasoning abilities, the MiMo model also integrates an MTP objective during its pre-training, demonstrating that the benefits of this approach extend beyond mere acceleration to enhancing the core capabilities of LLM. \cite{xiaomi2025mimo}.
\section{Non-AR-Based} \label{sec:Non-AR}
\begin{figure}
    \centering
    \includegraphics[width=1.0\linewidth]{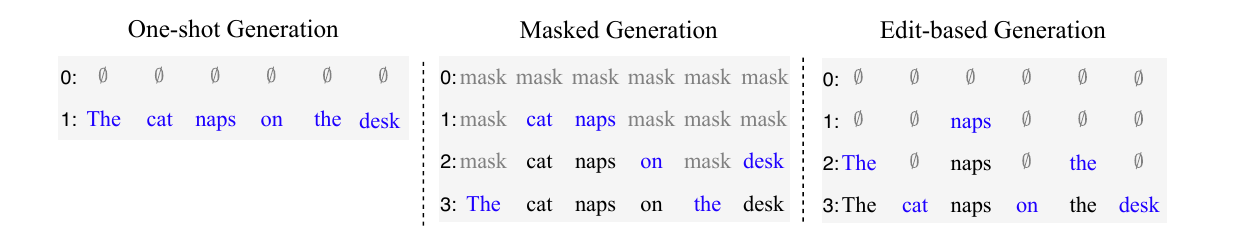}
    \caption{Comparison of decoding trajectories of One-shot Generation (left), Masked Generation (middle), and Edit-based Generation (right). One-shot generation produces all tokens in parallel in a single step. Masked generation begins from a fully masked sequence and progressively fills in tokens over multiple iterations. Edit-based generation operates by iteratively modifying an initial sequence through edits such as insertion, deletion, or replacement (in this figure, only insertion operations are shown for clarity).}
    \label{fig:non_ar_compare}
\end{figure}

Non-autoregressive (NAR) models depart from the left-to-right sequential decoding of autoregressive generation, aiming to improve inference speed by predicting multiple tokens in parallel. Existing approaches can be broadly categorized into three paradigms—\textit{one-shot}, \textit{masked}, and \textit{edit-based} generation—each offering a distinct balance between speed and output quality.
\textit{One-shot generation} (\Cref{sec:osg}) outputs the entire sequence in a single pass, maximizing speed but suffering from the conditional independence assumption, which can cause repetition, omissions, or incoherence, and is often addressed by reintroducing token dependencies or designing more tolerant training objectives.
\textit{Masked generation} (\Cref{sec:mg}) starts from a partially or fully masked sequence and progressively fills in tokens over multiple steps, allowing iterative refinement while still supporting parallel updates at each step.  
\textit{Edit-based generation} (\Cref{sec:ebr}) incrementally modifies an initial sequence through a series of learned editing operations (e.g., insertion, deletion, replacement), enabling more targeted adjustments and potentially fewer decoding steps for localized changes.
These paradigms address the limitations of sequential decoding from different perspectives, offering diverse trade-offs between latency, controllability, and output quality.  
An overview and intuitive comparison of the three paradigms is illustrated in \Cref{fig:non_ar_compare}. 

\subsection{One-shot Generation} \label{sec:osg}

The one-shot generation paradigm constitutes a significant departure from traditional sequential methods by producing all output tokens simultaneously in a single decoding pass. This parallelization eliminates the left-to-right dependency inherent in autoregressive models, thereby achieving substantial gains in inference speed. 

However, this parallelization introduces a fundamental challenge: the conditional independence assumption, in which each token is generated without awareness of its neighbors \cite{gu2017non, lee2018deterministicnonautoregressiveneuralsequence}. This assumption often leads to the "multimodality problem," resulting in errors such as token repetition, omission, and a general lack of coherence compared to autoregressive counterparts \cite{li2020lava, song2021alignart, ran2021guiding}. Consequently, a significant body of research has focused on addressing these deficiencies, which can be broadly categorized into two main approaches: reintroducing token dependencies and refining training objectives.

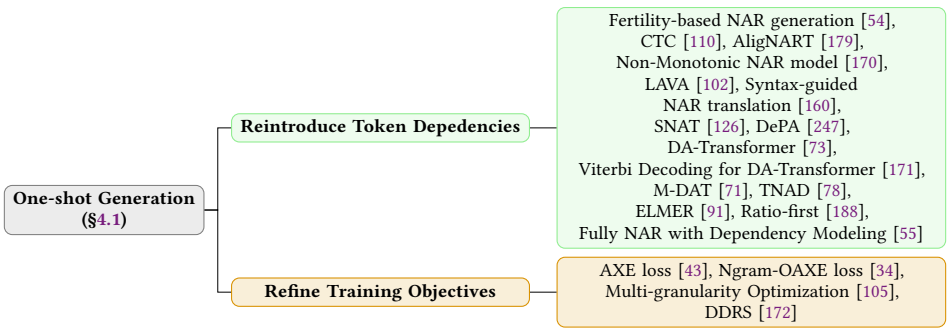
\begin{figure}[htbp]
	\centering
	\tikzset{
		my node/.style={
			draw,
			align=center,
			thin,
			text width=1.2cm, 
			rounded corners=3,
		},
		my leaf/.style={
			draw,
			align=center,
			thin,
			text width=8.5cm, 
			rounded corners=3,
		}
	}
	\forestset{
		every leaf node/.style={
			if n children=0{#1}{}
		},
		every tree node/.style={
			if n children=0{minimum width=1em}{#1}
		},
	}
	\begin{forest}
		nonleaf/.style={font=\bfseries\scriptsize},
		for tree={%
			every leaf node={my leaf, font=\scriptsize},
			every tree node={my node, font=\scriptsize, l sep-=4.5pt, l-=1.pt},
			anchor=west,
			inner sep=2pt,
			l sep=10pt, 
			s sep=3pt, 
			fit=tight,
			grow'=east,
			edge={ultra thin},
			parent anchor=east,
			child anchor=west,
			if n children=0{}{nonleaf}, 
			edge path={
				\noexpand\path [draw, \forestoption{edge}] (!u.parent anchor) -- +(5pt,0) |- (.child anchor)\forestoption{edge label};
			},
			if={isodd(n_children())}{
				for children={
					if={equal(n,(n_children("!u")+1)/2)}{calign with current}{}
				}
			}{}
		}
		[One-shot Generation\\(\cref{sec:osg}), draw=gray, fill=gray!15, text width=2.5cm, text=black
		    [Reintroduce Token Depedencies, color=lightgreen, fill=lightgreen!15, text width=3.8cm, text=black
                [
                    {
                    Fertility-based NAR generation~\cite{gu2017non},\\
                    CTC~\cite{libovicky2018end}, AligNART~\cite{song2021alignart},\\
                    Non-Monotonic NAR model~\cite{shao2022non},\\
                    LAVA~\cite{li2020lava}, 
                    Syntax-guided NAR translation~\cite{ran2021guiding}, \\
                    SNAT~\cite{liu2021enriching}, DePA~\cite{zhan2023depaimprovingnonautoregressivemachine},\\ 
                    DA-Transformer~\cite{huang2022directed},\\ 
                    Viterbi Decoding for DA-Transformer~\cite{shao2022viterbi},\\
                    M-DAT~\cite{huang2025mdat}, TNAD~\cite{ji2025tnad},\\
                    ELMER~\cite{lee2018deterministicnonautoregressiveneuralsequence}, Ratio-first~\cite{su2021nonautoregressivetextgenerationpretrained},\\
                    Fully NAR with Dependency Modeling~\cite{gu2020fullynonautoregressiveneuralmachine}
                    },
                color=lightgreen, fill=lightgreen!15, text=black, text width=5.0cm
                ]
            ]
		    [Refine Training Objectives, color=harvestgold, fill=harvestgold!15, text width=3.8cm, text=black
                [
                {
                AXE loss~\cite{ghazvininejad2020aligned},
                Ngram-OAXE loss~\cite{du2022ngram},\\
                Multi-granularity Optimization~\cite{li2022multi},\\
                DDRS~\cite{shao2022one}
                }, 
                 color=harvestgold, fill=harvestgold!15, text=black, text width=5.0cm
                 ]
            ]
		]
	\end{forest}
	\caption{Taxonomy of one-shot generation methods}
	\label{fig:taxonomy_osg}
\end{figure}

One major line of research focuses on re-introducing token dependencies that are lost in the parallel decoding process. Early work explored using latent variables, such as fertilities \cite{gu2017non} or alignments learned via Connectionist Temporal Classification (CTC), to guide generation \cite{libovicky2018end}. This was extended by incorporating more explicit and sophisticated alignment mechanisms, such as jointly learning alignments and translations \cite{song2021alignart} and allowing for non-monotonic alignments to better handle complex word reordering \cite{shao2022non}. Other strategies aim to directly enhance dependency modeling within the decoder architecture. These include a variety of representative methods. Look-around decoding accounts for neighboring token predictions in the decoding process \cite{li2020lava}. Another common approach enriches the model with external syntactic and semantic structures to enhance its performance \cite{liu2021enriching}. Specialized pre-training is also adopted to develop dependency-aware decoders for more accurate sequence generation \cite{zhan2023depaimprovingnonautoregressivemachine}. More advanced architectural changes have also been proposed, such as the Directed Acyclic Transformer (DA-Transformer), which models multiple translation paths simultaneously in a single pass \cite{huang2022directed}, later improved with Viterbi decoding to find the optimal path \cite{shao2022viterbi}. The use of powerful pre-trained language models has also been shown to inherently improve dependency modeling and overall generation quality \cite{lee2018deterministicnonautoregressiveneuralsequence, su2021nonautoregressivetextgenerationpretrained}. Moreover, a range of other optimized methods have been proposed for this task: M-DAT~\cite{huang2025mdat} attains state-of-the-art performance in non-autoregressive multilingual machine translation without resorting to knowledge distillation, by virtue of pivot back-translation. Tree-structured Non-Autoregressive Decoding (TNAD)~\cite{ji2025tnad} establishes a bridge between autoregressive (AR) and non-autoregressive (NAR) translation paradigms, realizing generation through the top-down, layer-wise expansion of constituency parse trees to enable parallel generation within individual layers while retaining syntactic dependencies across different layers.

A second stream of research addresses these shortcomings by refining the training objectives and decoding strategies. The standard cross-entropy loss is often too strict for NAR models, as it heavily penalizes minor word order shifts. To address this, alternative loss functions have been proposed, such as Aligned Cross Entropy (AXE), which tolerates monotonic reordering \cite{ghazvininejad2020aligned}, and its phrase-based extension, ngram-OAXE, which permits reordering of n-gram chunks \cite{du2022ngram}. Similarly, multi-granularity optimization provides feedback at various segment levels to train more robust models \cite{li2022multi}. The training process itself has been a focus of improvement, with techniques like diverse distillation using multiple reference translations to alleviate the constraints of learning from a single, often arbitrary, reference \cite{shao2022one}. 

In summary, while the one-shot generation paradigm offers the greatest potential for latency reduction, its core weakness lies in the conditional independence assumption. This fundamental trade-off leads to a notable degradation in translation quality, often manifesting as multimodality issues, poor word order, and token repetition. The research in this area has thus been a continuous effort to bridge this quality gap by developing sophisticated mechanisms—from advanced alignment and dependency modeling to novel loss functions and training paradigms—all while striving to preserve the essential speed advantage of parallel decoding.

\subsection{Masked Generation} \label{sec:mg}

Masked generative models generate content by iteratively filling in masked positions, starting from a fully masked input. 
In NLP, this approach was first introduced by Mask-Predict~\cite{ghazvininejad2019mask}, where a BERT-style model predicts multiple masked tokens in parallel and progressively refines the sequence over several steps. 
In image generation, MaskGIT~\cite{chang2022maskgit} adopts a similar strategy, using confidence-guided parallel token prediction. 
These models naturally support parallel decoding by allowing simultaneous updates to multiple tokens at arbitrary positions. 
Recent advances~\cite{google2025gemini,seed2025diffusion,labs2025mercury} have demonstrated that this approach can achieve substantial speedups and lower latency compared to left-to-right sequential autoregressive generation. A quantitative comparison of representative masked diffusion language models is provided in \Cref{tab:diffusion_llm_comparison}.

This section is organized as follows. We first formalize masked generative models as absorbing-state diffusion processes (\cref{sec:formalize_mdm}). We then present \textbf{quality-oriented} methods (\Cref{sec:mdm_train}), which aim to improve generation fidelity through enhanced training, including pre-training and post-training. In contrast, \textbf{speed-oriented} methods focus on accelerating the decoding speed while maintaining quality. These include decoding strategies (\Cref{sec:decoding_strategy}) that enable parallel decoding. We further discuss two additional speed-oriented techniques: step-reduction (\Cref{sec:distillation}) and system-level optimization  (\Cref{sec:kv-cache}).

\begin{figure}[htbp]
	\centering
	\tikzset{
		my node/.style={
			draw,
			align=center,
			thin,
			text width=1.2cm,
			rounded corners=3,
		},
		my leaf/.style={
			draw,
			align=center,
			thin,
			text width=8.5cm,
			rounded corners=3,
		}
	}
	\forestset{
		every leaf node/.style={ if n children=0{#1}{} },
		every tree node/.style={ if n children=0{minimum width=1em}{#1} },
	}
	\begin{forest}
		nonleaf/.style={font=\bfseries\scriptsize},
		for tree={%
			every leaf node={my leaf, font=\scriptsize},
			every tree node={my node, font=\scriptsize, l sep-=4.5pt, l-=1.pt},
			anchor=west,
			inner sep=2pt,
			l sep=10pt,
			s sep=3pt,
			fit=tight,
			grow'=east,
			edge={ultra thin},
			parent anchor=east,
			child anchor=west,
			if n children=0{}{nonleaf},
			edge path={
				\noexpand\path [draw, \forestoption{edge}] (!u.parent anchor) -- +(5pt,0) |- (.child anchor)\forestoption{edge label};
			},
			if={isodd(n_children())}{
				for children={
					if={equal(n,(n_children("!u")+1)/2)}{calign with current}{}
				}
			}{}
		}
		[Masked Generation, draw=gray, fill=gray!15, text width=1.3cm, text=black
		[Quality-oriented\\(\cref{sec:mdm_train}), color=capri, fill=capri!15, text width=2.6cm, text=black
		[Pre-training\\(\S\ref{sec:pre_train}), color=lightgreen, fill=lightgreen!15, text width=2.0cm, text=black
		[{
			LLaDA~\cite{nie2025large},
            LLaDA-2.0~\cite{bie2025llada2},
            LLaDA-MOE~\cite{zhu2025llada},
			DiffLLaMA~\cite{gong2024scaling},
			Dream~\cite{dream2025},
			DBA~\cite{prabhudesai2025diffusion},
			Seed Diffusion~\cite{seed2025diffusion}
		}, color=lightgreen, fill=lightgreen!15, text width=5.5cm, text=black]
		]
		[Post-training\\(\S\ref{sec:post_train}), color=harvestgold, fill=harvestgold!15, text width=2.0cm, text=black
		[{%
			d1-LLaDA~\cite{zhao2025d1},
            d2~\cite{wang2025d2},
            wd1~\cite{tang2025wd1},
            d-TreeRPO~\cite{pan2025d},
            SAPO~\cite{sapo},
            GDPO~\cite{gdpo},
            ESPO~\cite{espo},
			LLaDA-1.5~\cite{zhu2025llada},
            LLaDA-2.0~\cite{bie2025llada2},
			DiffCoder~\cite{gong2025diffucoder}, 
            DreamCoder~\cite{xie2025dream},
            DreamOn~\cite{wu2026dreamon},
            SPG~\cite{wang2025spg},
            TraceRL~\cite{wang2025revolutionizing},
            DiffPO~\cite{chen2025diffpo},
            SDAR~\cite{cheng2025sdar},
            WeDLM~\cite{liu2025wedlm},
            SBD~\cite{gat2025set},
            Fast-DLLM-v2~\cite{wu2025fast}
		}, color=harvestgold, fill=harvestgold!15, text width=5.5cm, text=black]
		]
		]
		[Speed-oriented\\(\Cref{sec:decoding_strategy,sec:distillation,sec:kv-cache}), color=brightlavender, fill=brightlavender!15, text width=2.6cm, text=black
		[Decoding Strategy\\(\cref{sec:decoding_strategy}), color=lightgreen, fill=lightgreen!15, text width=2.0cm, text=black
		[Non-adaptive Parallel Decoding\\(\S\ref{sec:uninformed}), color=lightgreen, fill=lightgreen!15, text width=2.0cm, text=black
		[{%
			$\tau$-leapping~\cite{shi2024simplified,campbell2022continuous,lou2024discrete,seed2025diffusion,zhang2025target},
			Tweedie $\tau$-leaping~\cite{sun2022score,lou2024discrete}
		}, color=lightgreen, fill=lightgreen!15, text width=3.7cm, text=black]
		]
		[Heuristic-based Decoding\\(\S\ref{sec:heuristic-informed}), color=harvestgold, fill=harvestgold!15, text width=2.0cm, text=black
		[{%
			Fast-dLLM~\cite{wu2025fast},
			LLaDA~\cite{nie2025large},
			MGDM~\cite{ye2024beyond},
			RDM~\cite{zheng2023reparameterized},
			TWPB~\cite{kim2025train},
			EB-sampler~\cite{ben2025accelerated},
			SlowFast~\cite{wei2025accelerating},  WINO~\cite{hong2025wide},
            Prophet~\cite{prophet},
            CreditDecoding~\cite{wang2025creditdecoding}
		}, color=harvestgold, fill=harvestgold!15, text width=3.7cm, text=black]
		]
		[Planner-based Decoding\\(\S\ref{sec:planner-informed}), color=lightcoral, fill=lightcoral!15, text width=2.0cm, text=black
		[{%
			CT-MDM~\cite{campbell2022continuous},
			ReMDM~\cite{wang2025remasking},
			MD4~\cite{shi2024simplified},
			P2~\cite{peng2025path},
			DDPD~\cite{liu2024think},
			RDM~\cite{zheng2023reparameterized}
		}, color=lightcoral, fill=lightcoral!15, text width=3.7cm, text=black]
		]
		[Externally-guided Decoding\\(\S\ref{sec:external-informed}), color=DeepSkyBlue4, fill=DeepSkyBlue4!15, text width=2.0cm, text=black
		[{%
			APD~\cite{israel2025accelerating},
			ASSD~\cite{guo2025reviving},
			FreeCache~\cite{hu2025accelerating}
		}, color=DeepSkyBlue4, fill=DeepSkyBlue4!15, text width=3.7cm, text=black]
		]
		]
		[Step Reduction\\(\cref{sec:distillation}), color=harvestgold, fill=harvestgold!15, text width=2.0cm, text=black
		[{%
			SDTT~\cite{deschenaux2024beyond},
			CLLM~\cite{kou2024cllms},
            D3LLM~\cite{qian2026d3llm},
            dParallel~\cite{chen2025dparallel},
            D2F~\cite{d2f},
			Duo~\cite{sahoo2025diffusion}
		}, color=harvestgold, fill=harvestgold!15, text width=5.5cm, text=black]
		]
		[System-level Acceleration\\(\cref{sec:kv-cache}), color=lightcoral, fill=lightcoral!15, text width=2.0cm, text=black
		[{%
			Fast-dLLM~\cite{wu2025fast},
			dkv-cache~\cite{ma2025dkv},
			FreeCache~\cite{hu2025accelerating},
            BD3-LM~\cite{arriola2025block},
			Eso-LMs~\cite{sahoo2025esoteric},
            WeDLM~\cite{liu2025wedlm},
            LOPA~\cite{xu2025lopa},
            dInfer~\cite{ma2025dinfer}
		}, color=lightcoral, fill=lightcoral!15, text width=5.5cm, text=black]
		]
		]
		]
	\end{forest}
	\caption{Taxonomy of masked generation methods}
	\label{fig: masked_taxonomy}
\end{figure}
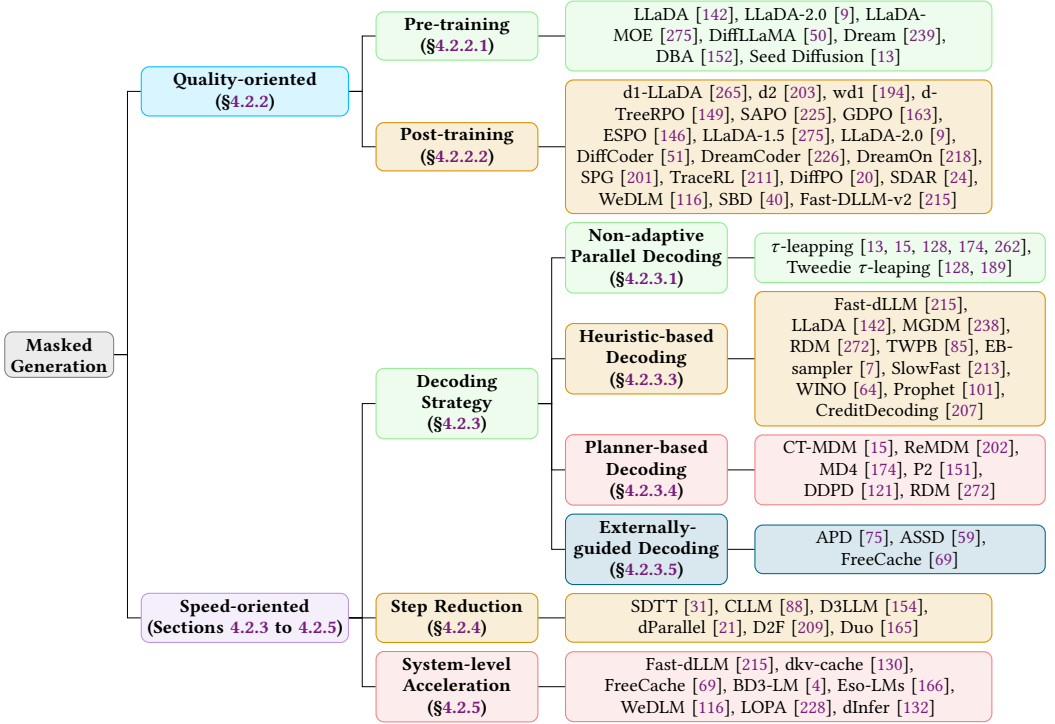

\subsubsection{Formalizing Masked Generative Models}
\label{sec:formalize_mdm}
Masked generative models can be formalized as an \textit{masked diffusion models} (MDM)~\cite{austin2021structured,ou2024your,zheng2024masked,lou2024discrete,shi2024simplified,campbell2022continuous}. 
Let \(\mathbf{x}_0 \in \mathcal{V}^N\) denote the original clean sequence of length \(N\), where \(\mathcal{V}\) is the vocabulary.
The model distribution \(p_\theta(\boldsymbol{x}_0)\) is defined via:
1) A forward process that progressively corrupts \(\boldsymbol{x}_0\) by independently replacing each token with a special mask token \([\mathrm{MASK}]\).
and 2) A reverse process that reconstructs the original sequence by reversing the corruption.

The corruption rate is controlled by a time-dependent masking schedule \(\sigma(t) \in [0,1]\), where \(\sigma(t)\) denotes the probability that a token remains unmasked at time \(t\).  
A common choice is the linear schedule \(\sigma(t) = 1 - t\), see a summary of masking schedules from literature in \citet{shi2024simplified}.
For \(t \in (0,1)\), the partially corrupted sequence \(\boldsymbol{x}_t\) is obtained by independently masking each position with probability \(1 - \sigma(t)\).

Let \(\mathcal{M}_t \subseteq \{1, \dots, N\}\) be the set of masked positions in \(\mathbf{x}_t\).  
The core to enable the reversed process is a parametric mask prediction model \(p_\theta(\boldsymbol{x}_{0}^{\mathcal{M}_t} \mid \boldsymbol{x}_t)\) that takes a corrupted input $x_t$ as input and predicts all masked tokens simultaneously.
\begin{equation} \label{eq:cond_indep}
p_\theta(\boldsymbol{x}_{0}^{\mathcal{M}_t} \mid \boldsymbol{x}_t) 
\ \approx\  \prod_{i \in \mathcal{M}_t} p_\theta(\boldsymbol{x}_0^i \mid \boldsymbol{x}_t).
\end{equation}
where the right-hand side represents a factorization of the joint conditional distribution into independent per-token conditionals given the unmasked context.
The ``\(\approx\)'' captures the \textbf{conditional independence assumption}: masked tokens are predicted independently given the unmasked context.  
This assumption overlooks potential strong mutual dependencies between tokens—such as number and tense agreement in language, or structural constraints in code and tabular data—which cannot be fully captured when predicting them independently. Consequently, it induces a trade-off: decoding many tokens in a single step amplifies factorization error, while decoding fewer tokens necessitates more refinement steps.

The model is trained to minimize the cross-entropy loss over masked positions, where \(t \sim \mathcal{U}[0,1]\) controls the masking ratio:
\begin{equation} \label{eqn:mdm_loss}
\mathcal{L}(\theta) 
= \mathbb{E}_{t, \boldsymbol{x}_0, \boldsymbol{x}_t} \left[
\frac{\sigma'(t)}{1-\sigma(t)} 
\sum_{i \in \mathcal{M}_t} \log p_\theta(\boldsymbol{x}_0^i \mid \boldsymbol{x}_t)
\right].
\end{equation}
where $\sigma'(t)$ is the first-order derivative of $\sigma(t)$ w.r.t $t$.
This loss is known to upper bound the negative log-likelihood \(-\log p_\theta(\mathbf{x}_0)\)~\cite{shi2024simplified,ou2024your}, providing a principled training objective for generative modeling.  
Notably, \Cref{eqn:mdm_loss} closely resembles (up to the time-dependent scaling factor $-\frac{\sigma'(t)}{1-\sigma(t)}$) the training loss of \textit{any-order autoregressive models} (AO-ARMs)~\cite{uria2014deep,shih2022training,hoogeboom2021autoregressive,yang2019xlnet}.  
Indeed, masked generative models can be interpreted from two perspectives: as AO-ARMs~\cite{shi2024simplified,ou2024your,hoogeboom2021autoregressive} or as masked diffusion models~\cite{campbell2022continuous,ou2024your,shi2024simplified,zheng2024masked}. 
In this section, we focus on introducing the conditional independence assumption and defer the rigorous continuous-time formulation of MDMs to \Cref{appendix:mdm_math}.  

\subsubsection{Training MDMs} \label{sec:mdm_train}
High-quality generation in masked diffusion models (MDMs) fundamentally depends on effective training of the mask prediction model.
Recent works have significantly advanced the training of MDMs, which share strong similarities with AR training. 
In particular, MDMs also adopt a two-stage training pipeline: 
\begin{enumerate}
    \item \textbf{Pre-training}, where the model is trained on a large corpus to learn the data distribution \( p(\boldsymbol{x}_0) \).
    \item \textbf{Post-training}, where Supervised fine-tuning (SFT) or Reinforcement learning from human feedback (RLHF) is performed on prompt-response pairs \((p_0, r_0)\) to align the model with human preferences.
\end{enumerate}
In the following, we focus on illustrating the key techniques that have been proposed and tailored for MDMs and how they are different from traditional AR training.

\begin{table}[t]
  \centering
  \caption{Comparison of Masked Diffusion Language Models}
  \label{tab:diffusion_llm_comparison}
  \setlength{\tabcolsep}{3.5pt}
  \begin{tabular}{lccccc}
    \toprule
    \textbf{Model} & \textbf{Year} & \textbf{Params} & \textbf{Type} & \textbf{Open Source} & \textbf{Speed (tokens/s)} \\
    \midrule
    \multicolumn{6}{l}{\textbf{Large-Scale Proprietary Models}} \\
    Gemini Diffusion~\cite{google2025gemini} & 2025 & — & — & No  & 1489 (H20) \\
    Mercury~\cite{labs2025mercury} & 2025 & — & — & No  & 1109 (H20) \\
    Seed Diffusion~\cite{seed2025diffusion} & 2025 & — & — & No  & 2146 (H20) \\
    \midrule
    \multicolumn{6}{l}{\textbf{Large-Scale Open-Source Models ($\geq$ 7B)}} \\
    LLaDA~\cite{nie2025large} & 2025 & 1B-8B & Instruct & Yes & 30.5 (A800) \\
    LLaDA-2.0~\cite{bie2025llada2} & 2025 & 16B/100B & Instruct & Yes & 535 (-) \\
    LLaDA-MOE~\cite{zhu2025llada} & 2025 & 7B & Instruct & Yes & - \\
    Dream~\cite{dream2025} & 2025 & 7B & Instruct & Yes & 23.5 (A800) \\
    DreamCoder~\cite{xie2025dream} & 2025 & 7B & Instruct & Yes & - \\
    WeDLM~\cite{liu2025wedlm} & 2025 & 7B/8B & Instruct & Yes & 370 (-) \\
    DiffLLaMA~\cite{gong2024scaling} & 2024 & 127M-355M-7B & Instruct & Yes & — \\
    LLaDA-1.5~\cite{zhu2025llada} & 2025 & 8B & Reasoning & Yes & — \\
    d1-LLaDA~\cite{zhao2025d1} & 2025 & 8B & Reasoning & Yes & — \\
    DiffCoder~\cite{gong2025diffucoder} & 2025 & 7B & Reasoning & Yes & — \\
    LLaDA-V~\cite{you2025llada} & 2025 & 8B & Multimodal & Yes & — \\
    Dimple~\cite{Yu2025DimpleDD} & 2025 & 7B & Multimodal & Yes & — \\
    \midrule
    \multicolumn{6}{l}{\textbf{Small to Medium Scale Models ($\approx$ 1B)}} \\
    DiffusionLLM~\cite{ye2023diffusionllm} & 2023 & 86M-9.7B & Instruct & Yes & — \\
    RDM~\cite{zheng2023reparameterized} & 2023 & $\sim$1B & Instruct & Yes & — \\
    SEDD~\cite{lou2024discrete} & 2024 & 127M & Instruct & Yes & — \\
    RADD~\cite{ou2024your} & 2024 & 162M-405M & Instruct & Yes & — \\
    MDLM~\cite{sahoo2024simple} & 2024 & 110M & Instruct & Yes & — \\
    MD4~\cite{shi2024simplified} & 2024 & 198M & Instruct & Yes & — \\
    DDPD~\cite{liu2024think} & 2024 & 86M & Instruct & Yes & — \\
    SMDM~\cite{nie2024scaling} & 2024 & 110M & Instruct & Yes & — \\
    MGDM~\cite{ye2024beyond} & 2024 & 6M-85M-303M & Reasoning & Yes & — \\
    \bottomrule
  \end{tabular}
\end{table}

\subsubsubsection{Pre-training} \label{sec:pre_train}
In the following, we introduce MDM pre-training from three perspectives: training objective, initialization, and data.

\noindent\textbf{Training Objective.}
The mask-prediction cross-entropy loss (\Cref{eqn:mdm_loss}) is widely adopted in training MDMs~\cite{shi2024simplified,nie2024scaling,dream2025,ni2025difflm,lou2024discrete,ou2024your}, where a partially masked input is constructed by replacing clean tokens with mask tokens at a certain probability. 
Seed Diffusion~\cite{seed2025diffusion} augments the standard forward process with an edit-based corruption step to improve calibration and mitigate undesirable behaviors such as repetition during sampling. 
The forward process samples a corrupted sequence based on a predefined set of edit operations (e.g., deletions, insertions, substitutions). 
For the first \(80\%\) of training steps, a standard mask-based corruption process is used; for the remaining \(20\%\), the edit-based corruption is applied in addition to masking. 
This augmentation mitigates the detrimental inductive bias where the model learns a spurious correlation that unmasked tokens are always correct, leading to overconfidence and poor self-correction at inference. 
By introducing edits, the model is encouraged to re-evaluate all tokens, including those that were originally unmasked.

\noindent
\textbf{Initialization.}
MDMs typically employ a bi-directional Transformer as the mask predictor, where the bi-directional architecture allows the model to attend to the entire input during prediction. While LLaDA~\cite{nie2024scaling} demonstrated that MDM training can be performed from scratch, subsequent works~\cite{gong2024scaling,dream2025} found that initializing the diffusion language model with weights from an existing autoregressive (AR) model provides a strong starting point. This initialization strategy has been shown to be more effective than training from scratch, particularly in the early stages of training.

\noindent
\textbf{Data.}
MDMs can be trained with a dataset size and computational cost (in FLOPs) comparable to those of AR models~\cite{nie2024scaling}. More recent studies~\cite{ni2025difflm,prabhudesai2025diffusion} revealed that MDMs are in fact more data-efficient than AR models, benefiting substantially from repeated exposure to the same training data. Notably, increasing the number of repetitions continues to improve performance without clear signs of diminishing returns, highlighting a promising direction for future research into the data efficiency of MDMs.

\subsubsubsection{Post-training} \label{sec:post_train}
While post-training has been extensively studied for autoregressive (AR) models, its application to MDMs has only recently begun to attract attention. Existing approaches largely build upon the two dominant paradigms in AR LLMs: supervised fine-tuning (SFT) and reinforcement learning (RL) alignment. 

\noindent
\textbf{SFT.}
The implementation of SFT for MDMs is similar to pre-training; the only difference is that we leave the prompt unchanged and mask the
tokens in the response independently.

\noindent
\textbf{RLHF.}
In contrast, RL-based alignment for MDMs presents unique challenges due to their iterative, non-sequential generation process, which lacks the simple autoregressive log-probability factorization. To address this, several diffusion-native RL algorithms have been proposed. LLaDA~1.5~\cite{zhu2025llada} extends Direct Preference Optimization (DPO)~\cite{rafailov2023direct} to MDMs by estimating preference scores through ELBO approximations. To mitigate the bias and variance introduced by doubly Monte Carlo estimation, they propose Variance-Reduced Preference Optimization (VRPO), which combines increased sampling budgets, timestep-wise allocation, and antithetic sampling. This approach yields consistent improvements over SFT-only baselines across mathematics, coding, and general alignment benchmarks. The d1-LLaDA framework~\cite{zhao2025d1} adapts the GRPO algorithm~\cite{shao2024deepseekmath} to MDMs via a one-step log-probability estimator with random prompt masking, which acts as a regularizer, reduces the number of required online generations, and enables efficient scaling of gradient updates. Applied after SFT, this method leads to substantial gains in reasoning and planning tasks. 
Extending these efforts, a series of recent works~\cite{wang2025d2, tang2025wd1, wang2025revolutionizing,pan2025d, wang2025spg, espo,gdpo,justgrpo} aim to further stabilize training and enhance reasoning capabilities through refined policy optimization techniques. These approaches introduce mechanisms such as weighted policy updates~\cite{tang2025wd1}, step-aware value estimation~\cite{espo,wang2025revolutionizing}, and tree-based process reward~\cite{pan2025d} to better align the diffusion denoising process with logical reasoning chains. ~\cite{justgrpo} warns of a \texttt{flexibility trap}, arguing that the arbitrary generation order inherent to MDMs may fundamentally limit their potential for complex reasoning compared to autoregressive models.
For code generation, DiffuCoder~\cite{gong2025diffucoder} introduces coupled-GRPO, a diffusion-native RL method that avoids semi-autoregressive decoding by using a coupled-sampling scheme with complementary mask noise. This design reduces variance in policy gradient estimation and strengthens non-AR generation patterns, resulting in notable performance gains with limited training data.

\subsubsection{Decoding Strategy} \label{sec:decoding_strategy}
Once a mask prediction model is trained, the decoding strategy is crucial for accelerating generation without sacrificing quality. We first establish a non-adaptive baseline with $\tau$-leaping (\S\ref{sec:uninformed}). We then introduce the core design principles of adaptive strategies (\S\ref{sec:adaptive_overview}) and explore three distinct implementations: heuristic-based (\S\ref{sec:heuristic-informed}), planner-based (\S\ref{sec:planner-informed}), and externally-guided (\S\ref{sec:external-informed}).

\subsubsubsection{Non-Adaptive Parallel Decoding: $\tau$-Leaping} \label{sec:uninformed}
The foundational technique for enabling parallel decoding is $\tau$-leaping. Originally developed in chemical physics \cite{wilkinson2018stochastic,gillespie2001approximate}, $\tau-$leaping discretizes the continuous-time generation process. Instead of simulating each individual state change, it assumes constant transition rates over a small time interval (a "leap") and applies the cumulative updates at once to all tokens in parallel.

This process yields a simple, parallel sampling rule. For a decoding step from time $t$ to $s$ (where $t>s$), under a common linear noise schedule $\sigma(t) = 1-t$~\cite{shi2024simplified}, the update rule is:

\begin{equation} \label{eqn:tau-leaping}
p_{s \mid t}=\prod_{i=0}^{N-1} p_{s \mid t}\left(\boldsymbol{x}_s^i \mid \boldsymbol{x}_t\right), p_{s \mid t}\left(\boldsymbol{x}_s^i \mid \boldsymbol{x}_t\right)= \begin{cases}1, & \boldsymbol{x}_t^i \neq[\mathrm{mask}], \boldsymbol{x}_s^i=\boldsymbol{x}_t^i \\ \frac{s}{t}, & \boldsymbol{x}_t^i=[\mathrm{mask}], \boldsymbol{x}_s^i=[\mathrm{mask}] \\ \frac{t-s}{t} p_{\theta}\left(\boldsymbol{x}_s^i \mid \boldsymbol{x}_t\right), & \boldsymbol{x}_t^i=[\mathrm{mask}], \boldsymbol{x}_s^i \neq[\mathrm{mask}] .\end{cases}
\end{equation}
where $[\mathrm{mask}]$ denotes the mask state and $p_\theta(\cdot\mid \boldsymbol{x}_t)$ is the parametric mask prediction model. 
Based on \Cref{eqn:tau-leaping}, for $i$-th token, the update rule is:
1) if $\boldsymbol{x}_t^i \neq [\mathrm{mask}]$, it remains unchanged;  
2) if $\boldsymbol{x}_t^i = m$, it stays masked with probability $\frac{s}{t}$;  
3) if $\boldsymbol{x}_t^i = m$, it is unmasked with probability $\frac{t-s}{t}$, in which case its new value $\boldsymbol{x}_s^i$ is predicted by the denoiser’s predicted distribution $p_\theta(\cdot\mid \boldsymbol{x}_t)$. In practice, mean-parameterization is commonly adopted~\cite{shi2024simplified,zheng2024masked}, which predicts a scaler $\boldsymbol{x}_s^i$.

A closely related variant is the \textit{Tweedie $\tau$-leaping} sampler~\cite{sun2022score,lou2024discrete}, which analytically computes the posterior transition $p_{s|t}(x_s|x_t)$ via the Tweedie formula, analogous to posterior sampling in DDPM~\cite{ho2020denoising}. 
Under the commonly used linear noise schedule, however, Tweedie $\tau$-leaping simplifies exactly to the Euler rule above~\cite{zheng2024masked,ou2024your}.

\begin{figure}[htbp]
    \centering
    \includegraphics[width=1.0\linewidth]{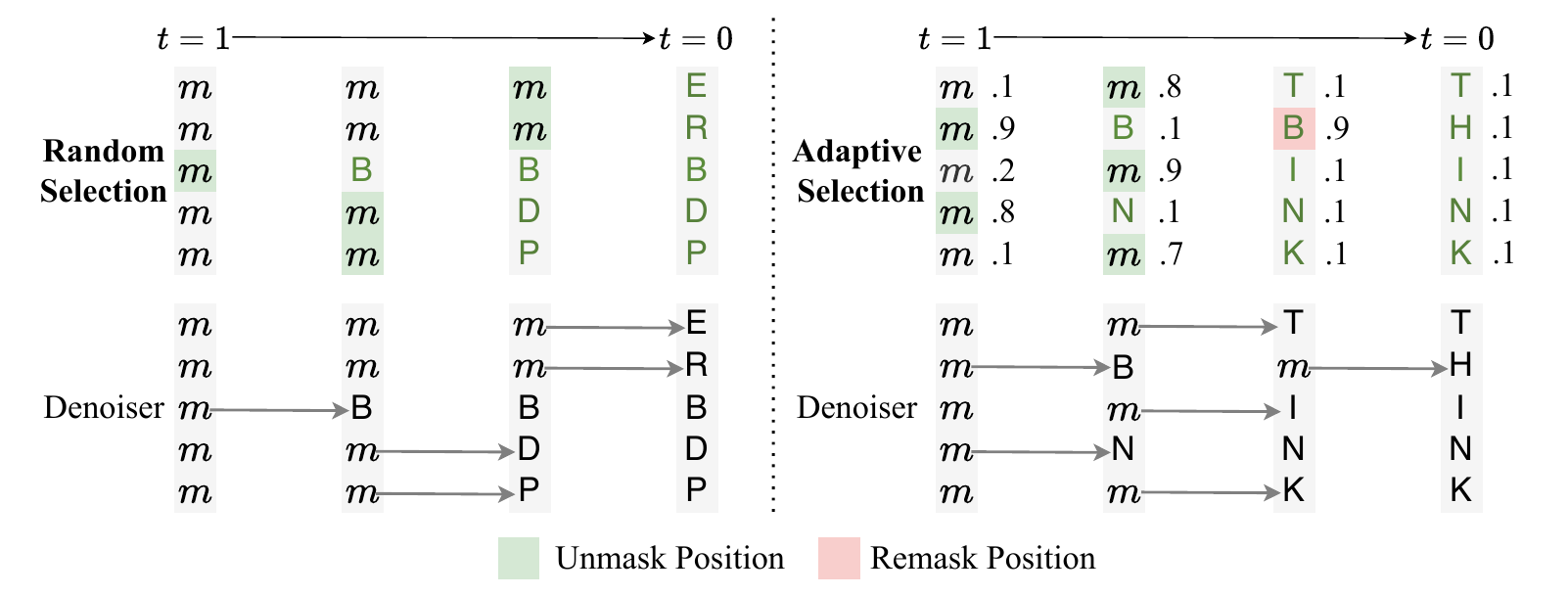}
    \caption{Comparison between \textbf{Non-Adaptive} and \textbf{Adaptive} parallel decoding in masked diffusion models. 
\textbf{Left}: Non-Adaptive decoding randomly selects unmasking positions according to a predefined time schedule (\S\ref{sec:uninformed}). 
\textbf{Right}: Adaptive decoding adaptively selects unmasking positions based on denoiser confidence (\S\ref{sec:heuristic-informed}), planner models (\S\ref{sec:planner-informed}), or external guidance (\S\ref{sec:external-informed}), and may further refine predictions by remasking already revealed tokens.}
    \label{fig:dllm}
\end{figure}

Despite its simplicity and high parallelism, $\tau$-leaping is an \textbf{non-adaptive} strategy (see the left side of \Cref{fig:dllm}): \textbf{it applies the same probabilistic rule to all masked tokens}, without considering token-specific uncertainty or potential prediction errors. 
As a result, decoding errors may accumulate when the model is uncertain, motivating the development of \textit{adaptive parallel decoding strategies} that selectively update tokens based on confidence or learned planning.

\subsubsubsection{Adaptive Parallel Decoding: Motivation} \label{sec:adaptive_overview}

The non-adaptive $\tau$-leaping strategy applies the same probabilistic rule to all masked tokens, disregarding token-specific uncertainty or dependencies. 
While simple, this uniform treatment can be problematic: it may be \textit{too conservative}, leaving many confidently predictable tokens masked and thus slowing down decoding, or \textit{too aggressive}, updating too many tokens simultaneously and introducing inconsistency due to violation of the conditional independence.

To address these issues, recent works aim to make parallel decoding \textbf{adaptive} (see the right side of \Cref{fig:dllm}), explicitly deciding 
\textit{which} tokens to decode and \textit{how many} to update at each step. 
Two key design dimensions have emerged:
\begin{itemize}
    \item \textbf{Parallel decoding size and conditional independence.}  
    Since the $\tau$-leaping decoding process (i.e., Eqn.~\eqref{eqn:tau-leaping}) is independently applied on each dimension, it implicitly assumes that token predictions are conditionally independent given the currently unmasked context. 
    This assumption breaks down when too many tokens are decoded simultaneously, particularly in regions of high uncertainty or strong semantic dependency, leading to correlated errors and error accumulation~\cite{wu2025fast}.
    Therefore, controlling the number of tokens updated per step is crucial to balance speed and accuracy.

   \item \textbf{Decoding order and model error.}  
The order in which tokens are revealed has a critical impact on generation quality because the denoising network is inherently imperfect. 
Ideally, a masked diffusion model learns to predict the true conditional distribution for any masking configuration~\cite{kim2025train,gong2024scaling}:
\begin{equation}
    \forall \mathcal{M}, \ p_\theta(\boldsymbol{x}_0^i \mid \boldsymbol{x}_{\mathcal{M}}) \approx p_{\text{data}}(\boldsymbol{x}_0^i \mid \boldsymbol{x}_{\mathcal{M}}),
\end{equation}
where \(\mathcal{M}\) denotes the set of masked indices, and \(\boldsymbol{x}_{\mathcal{M}}\) is obtained from \(\boldsymbol{x}_0\) by replacing the tokens at positions in \(\mathcal{M}\) with the mask token.
However, it has been empirically observed~\cite{ou2024your,shih2022training,li2021discovering} and theoretically demonstrated~\cite{kim2025train} that denoisers struggle to generalize to certain \textit{hard masking configurations}, where the context from unmasked tokens is insufficient to accurately predict the masked ones. 
The order of decoding thus plays a crucial role: predicting \textit{easy} tokens first—those that can be reliably inferred from the available context—provides stronger conditioning for predicting more difficult tokens later. 
An appropriate decoding order helps reduce error accumulation, whereas a poor one may lead to cascading mistakes, even if the number of tokens updated in parallel is carefully controlled.
\end{itemize}

The following subsections review representative methods that operationalize these principles, organized by how they control decoding order and parallel size:  
1) \textbf{Heuristic-based Decoding} (\S\ref{sec:heuristic-informed}): confidence- and uncertainty-based strategies that rank tokens by predicted reliability, including Top-K heuristics, confidence-guided decoding, and adaptive entropy-bounded sampling;  
2) \textbf{Planner-based Decoding} (\S\ref{sec:planner-informed}): planner-based methods that explicitly learn to decide which tokens to unmask or remask. 
3) \textbf{Externally-Guided Decoding} (\S\ref{sec:external-informed}): methods that incorporate external guidance or combine multiple models, including mixture distributions, speculative decoding, and guided diffusion;  

\subsubsubsection{Heuristic-based Decoding} \label{sec:heuristic-informed}

Heuristic-informed approaches improve upon the uninformed $\tau$-leaping by selectively updating tokens based on confidence or uncertainty estimates. 
These methods provide a lightweight way to control decoding order and parallel size without additional training.

\paragraph{Confidence Threshold.}  
Fast-dLLM~\cite{wu2025fast} introduces a \textit{confidence-aware parallel decoding} strategy that enables parallel token updates without compromising output quality by identifying approximately independent tokens. For each masked token $i$, a confidence score $c_i$ is computed, typically as the maximum softmax probability:
\begin{equation}
c_i = \max_j \, p_\theta(\boldsymbol{x}^i = j \mid \boldsymbol{x}_t).
\end{equation}
Tokens with $c_i$ exceeding a predefined threshold are decoded simultaneously, while the highest-confidence token is revealed to ensure progress if no token surpasses the threshold. When all selected tokens exhibit sufficiently high confidence, this strategy produces the same output as greedy sequential decoding. Theoretical analysis and empirical results further demonstrate that, under the condition of sufficiently high marginal confidence, the proposed confidence-aware parallel decoding—based on the assumption of independence and the use of the product of marginal distributions—closely approximates the true joint distribution well. Building on this theorem, \citet{wu2025fast} further proposes a practical \textit{factor-based parallel decoding} strategy as an extension of the threshold-based approach. This strategy adaptively determines the number of tokens to decode in parallel according to their confidence levels. Specifically, given the model's marginal confidence estimates for a block of tokens, we first sort the confidences in descending order and select the largest $n$ satisfying
\begin{equation} \label{eqn:factor}
(n + 1)\left(1 - c^{(n)}\right) < f,
\end{equation}
where $f$ is a fixed decoding-factor hyperparameter and $c^{(n)}$ is the $n$-th highest confidence score. The top-$n$ tokens are then decoded in parallel at each step.

\paragraph{Top-K Strategies.}  
Another popular family of parallel decoding strategies is \textit{Top-K}-based, which ranks masked tokens according to a confidence or uncertainty proxy and decodes the top $K$ most confident tokens at each iteration. Due to their simplicity and efficiency, these strategies are widely adopted in text generation~\cite{zheng2023reparameterized,wu2025fast,kim2025train,ben2025accelerated,ye2024beyond,nie2025large}. 
Common confidence proxies include:

\begin{itemize}
    \item \textbf{Top-K Probability}: Certainty for position $i$ is estimated by the maximum marginal probability:
    \begin{equation}
        \text{score}_i = \max_j p_\theta(\boldsymbol{x}^i = j \mid \boldsymbol{x}_t).
    \end{equation}
    This simple proxy has been shown to work well in practice~\cite{zheng2023reparameterized,ye2024beyond,nie2025large}; 
However, it performs poorly when multiple tokens have similar
probabilities. For example, when the model assigns two tokens nearly equal high probabilities, the position is still uncertain, yet Top-K Probability would incorrectly prioritize decoding it.

    \item \textbf{Top-K Margin}: To address this issue, \citet{kim2025train} propose a margin-based proxy that measures the probability gap between the two most likely tokens $j_1$ and $j_2$:
    \begin{equation}
        \text{score}_i = \left|p_\theta(\boldsymbol{x}^i = j_1 \mid \boldsymbol{x}_t) - p_\theta(\boldsymbol{x}^i = j_2 \mid \boldsymbol{x}_t)\right|.
    \end{equation}
    This approach provides a better estimate of uncertainty when multiple tokens have similar probabilities, effectively avoiding decoding uncertain positions. In cases where there is a clear single best choice, Top-K Margin behaves similarly to Top-K Probability.

    \item \textbf{Top-K Entropy}: Entropy provides an alternative way to address the drawback of Top-K probability proxies in situations where multiple tokens have similar probabilities. By considering the entire predictive distribution, entropy offers a more holistic measure of uncertainty~\cite{ben2025accelerated}:
    \begin{equation}
        \text{score}_i = -\sum_{j=1}^N p_\theta(\boldsymbol{x}^i = j \mid \boldsymbol{x}_t)\log p_\theta(\boldsymbol{x}^i = j \mid \boldsymbol{x}_t).
    \end{equation}
    Lower entropy indicates higher confidence, making it more reliable in distinguishing truly confident positions from ambiguous ones.
\end{itemize}

\paragraph{Adaptive Strategies.}  
While effective, using a fixed $K$ is often suboptimal. Selecting too many low-confidence tokens early can lead to correlated errors, whereas using a small $K$ underutilizes parallelism when confidence is uniformly high. 
To overcome the rigidity of a fixed $K$, recent work proposes \textit{adaptive} strategies that dynamically determine how many tokens to decode based on the confidence distribution. The Entropy-Bounded Sampler (EB-Sampler)~\cite{ben2025accelerated} selects the largest subset $\mathcal{S}$ of tokens satisfying:
\begin{equation}
    H(\mathcal{S}) - \max_{i \in \mathcal{S}} H(y_i) < \gamma,
\end{equation}
where $H(\cdot)$ denotes entropy and $\gamma$ is a predefined uncertainty budget. This criterion encourages the selection of high-confidence tokens that are approximately conditionally independent, thereby reducing error accumulation.
Similarly, the \textit{Factor}-Based Parallel Decoding strategy~\cite{wu2025fast}, previously discussed in the confidence-threshold section (see Equation~\eqref{eqn:factor}), also adaptively determines the decoding size. 
Both methods are theoretically grounded—EB-Sampler is derived from a KL-divergence-based analysis of decoding error~\cite{ben2025accelerated}, and Factor-Based Parallel Decoding stems from confidence-aware independence analysis~\cite{wu2025fast}. Empirically, these adaptive strategies achieve significant speedups, up to $2$--$3\times$ on reasoning and code-generation benchmarks, while maintaining or even improving generation fidelity compared to fixed-$K$ approaches. Complementing existing adaptive strategies, \citet{wei2025accelerating} proposes \textit{SlowFast} sampling, a two-phase decoding scheme for diffusion LLMs that alternates between cautious, exploratory updates of uncertain tokens (Slow phase) and aggressive, bulk parallel decoding of high-confidence tokens (Fast phase). Guided by three principles—Certainty, Convergence, and Positional—this method dynamically shifts modes to preserve coherence while maximizing throughput. Prophet~\cite{prophet} and Credit Decoding~\cite{wang2025creditdecoding} leverage confidence signals and early-layer determinism to achieve faster decoding.

\subsubsubsection{Planner-based Decoding} \label{sec:planner-informed}
Beyond heuristic-based strategies---where the decoding process is guided by simple uncertainty proxies such as confidence scores---\textit{Planner-informed} approaches aim to learn a planner model that selects which tokens should be unmasked at a given inference step, and optionally, which already unmasked tokens to be resampled. These methods typically decompose the parallel decoding process into two components:  
\begin{enumerate}
    \item \textbf{Planning step}: determining which tokens to unmask or remask at each step;
    \item \textbf{Denoising step}: predicting the target tokens conditioned on all currently available context.
\end{enumerate}

This learned approach addresses key limitations of masked diffusion models. For instance, the theoretical foundation of Masked Diffusion Models (MDM) assumes that every token is equally likely to be unmasked at any step, which ensures correct reconstruction of the data distribution only under a perfect denoiser. However, as discussed in the confidence-based strategies (\S\ref{sec:heuristic-informed}), practical denoisers are inherently imperfect, and empirical results show that a uniformly random unmasking order is often suboptimal.
Moreover, in masked diffusion sampling, once a token is unmasked, it is irrevocably fixed, preventing the model from revising erroneous predictions in future steps. This lack of course correction amplifies error propagation across steps, ultimately degrading generation quality. Planner-informed approaches aim to mitigate this issue by intelligently selecting the unmasking order to minimize error accumulation.

To mitigate the above issues, earlier work~\cite{campbell2022continuous}, inspired by Predictor-Corrector samplers in continuous state spaces~\cite{song2020score}, introduces a predictor-corrector framework for masked diffusion models. Specifically, an additional corrector step remasks already revealed tokens with a certain probability. However, this approach still treats all predicted tokens uniformly, without explicitly targeting potentially erroneous ones---a limitation in the masked diffusion setting. Nevertheless, \citet{campbell2022continuous} empirically demonstrates that even such an uninformed corrector can improve sample quality in practice. \citet{wang2025remasking} propose \textit{ReMDM} samplers, which extend the uninformed corrector in~\cite{campbell2022continuous} with confidence-informed correctors, motivated by the intuition that tokens for which the denoising model exhibits lower confidence should be assigned a higher probability of being remarked.
Similarly, \citet{shi2024simplified} replaces the uninformed corrector with a \textit{confidence-guided Gibbs sampler} that directly prioritizes resampling tokens most likely to be erroneous. At each step, $k$ token indices are selected without replacement from a categorical distribution:
\[
p(d) \propto \exp\!\left(-\frac{c_d}{\tau}\right),
\]
where $c_d$ is the probability or margin score for token $d$ (see definition of probability and margin score in \S\ref{sec:heuristic-informed}) and $\tau$ is a temperature parameter controlling selection sharpness. Among various uncertainty proxies, a \textit{margin-based} confidence score proves most effective, as it better distinguishes ambiguous cases where multiple candidates have similar probabilities, outperforming simple log-probability scores. To further accelerate sampling, this method supports \textit{parallel resampling} via a Hogwild-style Gibbs sampler~\cite{johnson2013analyzing}, where all $k$ selected tokens are updated simultaneously at each iteration.

Recent works extend the traditional predictor-corrector framework into a more principled \textit{planner-based framework}, which explicitly separates planning into two components: a \textit{mask planner}, assigning probabilities to whether a masked token should be unmasked, and an \textit{unmask planner}, assigning probabilities to whether an already unmasked token should be kept. In the traditional predictor-corrector framework, the corrector component effectively functions as an unmask planner, while the mask selection relies on an implicit, heuristic mask planner that only loosely determines which tokens to unmask.

Different instantiations of the mask and unmask planners have been explored. We first present the most general formulation introduced in P2~\cite{peng2025path}, and then discuss how this framework contrasts with DDPD~\cite{liu2024think} and RDM~\cite{zheng2023reparameterized} under specific choices of the mask and unmask planners.

\citet{peng2025path} propose P2 sampling, P2 departs from the vanilla MDM inference procedure, where the backward transition 
$q_{t, \theta}(x^i_{t-1} \mid x^i_t, D^i_\theta(\mathbf{x}_t))$ is denoised independently for each coordinate in the sequence by instead assigning the likelihood of denoising at $x^i_t$ 
as a \textit{function of the planner} $G_\phi$. This allows us to consider two cases, namely when 
the masked case $x^i_t = m$ and when $x^i_t \ne m$. 
Succinctly, P2 updates the partially noised sequence $\mathbf{x}_t$ by first sampling $\mathbf{z} \sim D_\theta(\mathbf{x}_t)$, after which we can leverage our planner to sample a position in the sequence to update, i.e., 
\[
i \sim \hat{G}_\phi(\boldsymbol{z}, \boldsymbol{x}_t) := \frac{G^i_\phi(\boldsymbol{z}, \boldsymbol{x}_t)}{\sum_{j=1}^{L} G^j_\phi(\boldsymbol{z}, \boldsymbol{x}_t)}.
\]

If $\boldsymbol{x}^i_t = m$, we sample using the following modified transition kernel 
$\boldsymbol{x}^i_{t-1} \sim q_{t,\theta}(\cdot \mid x^i_t, D^i_\theta(\boldsymbol{x}_t))$. Conversely, 
if $\boldsymbol{x}^i_t \ne m$, we construct $\bar{\boldsymbol{x}}_t$ from $\boldsymbol{x}_t$ via setting 
$\boldsymbol{x}^i_t$ to $\boldsymbol{m}$ (remasking), and then we resample 
$\boldsymbol{x}^i_{t-1} \sim q_{t,\theta}(\cdot \mid \boldsymbol{x}^i_t, D^i_\theta(\bar{\boldsymbol{x}}_t))$.

The \textsc{P2} framework formalizes the planner $G_\phi$ as two components: a mask planner $G_M^j$, which predicts whether a masked token should be unmasked, and an unmask planner $G_U^j$, which predicts whether an unmasked token should be kept. Formally,
\begin{equation}
G_\phi^j\left(\boldsymbol{z}, \boldsymbol{x}_t\right)=
\begin{cases}
G_M^j\left(\boldsymbol{z}, \boldsymbol{x}_t\right), & \boldsymbol{x}_t^j = m, \\
1 - G_U^j\left(\boldsymbol{z}, \boldsymbol{x}_t\right), & \boldsymbol{x}_t^j \neq m,
\end{cases}
\end{equation}
Three practical variants are then introduced to realize the mask planner $G_M$ and the unmask planners $G_U$. 1) \textit{Self-Planning} leverages the denoiser's own predictive distribution to guide both masking and unmasking decisions, i.e., $G^j_U(\boldsymbol{z},\boldsymbol{x}) = G^j_M(\boldsymbol{z},\boldsymbol{x}) = \text{Cat}(\boldsymbol{z}_j; D^j_\theta(\boldsymbol{x}))$. This approach is essentially equivalent to greedy decoding guided by a confidence proxy, with the only difference being that the planner model can also decide to modify an already decoded token. 2) \textit{BERT-Planning} employs a pretrained BERT model $B_\phi$ to assess token naturalness: the unmask planner is defined as $G^j_U(\boldsymbol{z},\boldsymbol{x}) = \text{Cat}(\boldsymbol{z}_j; B^j_\phi(\boldsymbol{z}))$, while the mask planner remains $G^j_M(\boldsymbol{z},\boldsymbol{x}) = \text{Cat}(\boldsymbol{z}_j; D^j_\theta(\boldsymbol{x}))$. BERT's versatility and widespread availability across domains (e.g., text, proteins, RNA) make it a flexible plug-in component. 3) \textit{Trained-Planner} involves training a separate planner network to imitate the optimal decoding path. With the denoiser frozen, the planner is optimized using cross-entropy loss to predict whether each token should be selected, based on whether the denoiser's output matches the ground truth. This idea was also previously proposed in the domain of masked models for image generation \cite{lezama2022improved}. 
Furthermore, P2 can be further generalized by introducing a stochasticity parameter $\eta$, which controls the frequency of remasking during sampling. Larger values of $\eta$ increase the likelihood of remasking. 

From this unified planner-based perspective, RDM~\cite{zheng2023reparameterized} and DDPD~\cite{liu2024think} can be viewed as special cases of P2. 
RDM~\cite{zheng2023reparameterized} is effectively equivalent to the \textit{Self-Planning} variant of P2, as it uses the denoiser for both mask and unmask planning. However, it lacks explicit stochasticity control, operating with a default stochasticity strength of $\eta = 1$. DDPD~\cite{liu2024think}, in contrast, introduces external planners and relies entirely on the planner for both mask and unmask decisions, also with a default $\eta = 1$. Crucially, the planner in DDPD only takes as input a partially unmasked sequence with randomly flipped tokens and is independent of the denoiser’s output. Consequently, the denoiser’s input is determined solely by the planner.  

One important advantage of planner-based decoding in masked diffusion models is its ability to refine previously erroneous tokens by explicitly deciding which positions to update at each step.
This refinement capability is not unique to masked diffusion: \textit{uniform-state diffusion models}~\cite{campbell2022continuous,austin2021structured,lou2024discrete} also inherently allow tokens to be revisited and corrected during sampling. However, earlier works~\cite{lou2024discrete,austin2021structured,campbell2022continuous} show that it is typically outperformed by masked diffusion models. However, recent works have narrowed the gap.
\citet{sahoo2025diffusion} establishes a theoretical connection between continuous and discrete diffusion and transfer techniques from Gaussian diffusion to improve both the training and sampling of uniform diffusion models. Specifically, Duo adapts two techniques from continuous Gaussian diffusion: a \textit{Gaussian-guided curriculum learning} strategy, which halves training time by stabilizing learning dynamics, and a \textit{Discrete Consistency Distillation} method, which adapts consistency distillation from the
continuous to the discrete setting. \citet{liu2024think} identify a key reason for the performance gap of uniform diffusion models by decomposing the transition probability into two components: the \textit{planning probability}, which determines whether a token should be corrupted, and the \textit{denoising probability}, which specifies the value to which a corrupted token should be changed. In the masked diffusion setting, the planning probability can be directly inferred from the explicit mask token. In contrast, uniform diffusion lacks such an explicit indicator, requiring the denosing model itself to compute or approximate this probability, thereby introducing potential errors. To address this issue, \citet{liu2024think} proposes parameterizing separate models for planning and denoising, effectively extending the model's capacity and improving performance.

\subsubsubsection{Externally-Guided Decoding} 
\label{sec:external-informed}

Externally-informed parallel decoding refers to strategies that incorporate guidance from models other than the masked diffusion model (MDM) itself to improve decoding accuracy or stability. Such external models—typically autoregressive or pretrained language models—provide complementary information, such as better modeling of token dependencies or additional confidence signals, which can be used to refine or validate the MDM’s predictions during parallel decoding.

\paragraph{Mixture Distribution}
Since a masked diffusion model gives a marginal distribution over a specific token, conditioning on the observed tokens, i.e., $p(\boldsymbol{x}_i \mid \boldsymbol{x}_{\text{obs}})$. Denote a subset of tokens as $\mathcal{S} \subseteq \mathcal{X}$, if we decode this subset in parallel, we effectively sample from the joint distribution $p_D$, which is defined as:
\begin{equation}
 p_D(x_{\mathcal{S}} \mid x_{\mathcal{O}}, \theta) = \prod_{i\in\mathcal{S}} p(x_i \mid x_{\mathcal{O}}, \theta).
\end{equation}  
where $\mathcal{O}$ denotes the observed tokens. If the denoiser is well-trained, the learned marginal distribution $p(x_i \mid x_{\mathcal{O}}, \theta)$ is close to the true marginal distribution $p(x_i \mid x_{\mathcal{O}})$. However, the ground truth joint distribution $p(x_{\mathcal{S}} \mid x_{\mathcal{O}})$ cannot be decomposed into the product of the marginal distributions $p(x_i \mid x_{\mathcal{O}})$ since the conditional independence assumption should not hold in general.

To address this, \citet{israel2025accelerating} proposes to leverage a smaller autoregressive model $p_{AR}$, which also defines a joint distribution through the chain rule:
\begin{equation}
    p_{AR}(\boldsymbol{x};\theta) = \prod_{i=1}^n p(\boldsymbol{x}_i \mid \boldsymbol{x}_{1:i-1}, \theta).
\end{equation}
where $\boldsymbol{x}_{1:i-1}$ denotes the observed tokens up to the $i$-th position.
Compared to $p_D$, $p_{AR}$ better models the dependency between tokens. However, the marginal distribution over a specific token $p(\boldsymbol{x}_i \mid \boldsymbol{x}_{\mathcal{O}}, \theta)$ is not accurate since we use a small autoregressive model. Therefore, the two distributions are both approximations and offer their own advantages. To combine the strengths of both, \citet{israel2025accelerating} leverages the intuition that if either of the models is highly confident—defined as having a high maximum probability in its logits over a token—then the prediction is likely to be accurate. 

A multiplicative mixture of distributions, also known as a product of experts, realizes the intuition above. \citet{israel2025accelerating} defines the multiplicative mixture of $p_D$ and $p_{AR}$ as follows:
\begin{equation}
    p_T(\boldsymbol{x}) = \frac{1}{Z} p_{\text{D}}(\boldsymbol{x})^R \, p_{\text{AR}}(\boldsymbol{x})^{1 - R},
\end{equation}
where $R$ is a hyperparameter that controls the mixture ratio. When $R = 1$, the target distribution is $p_D$, and the algorithm will accept every token from the diffusion model in one shot. When $R = 0$, the algorithm does not trust the diffusion model and instead only accepts tokens that $p_{AR}$ accepts. In addition, if either of the models is highly confident, the contribution of the distribution overweights the other, making the final distribution closer to the confident model.

\paragraph{Speculative Decoding}
\cite{guo2025reviving} introduces Any-Subset Speculative Decoding (ASSD), a decoding strategy tailored for discrete diffusion models. It leverages XLNet \cite{yang2019xlnet} to generate multiple token predictions independently and in parallel, using these conditionally independent guesses as the draft. In contrast to traditional speculative decoding—where a smaller draft model is verified by a larger oracle model—ASSD employs the same model for both drafting and verification. This is made possible by feeding this draft back through the same model (with computations for the already-visible tokens
cached) to calculate the oracle density estimates. The draft tokens are then
accepted or rejected based on their fidelity to the oracle density. 

\paragraph{Guidence}
\cite{hu2025accelerating} proposes \textit{Guided Diffusion}, a training-free approach that leverages a lightweight, pretrained autoregressive language model (ARM) to supervise the token unmasking process of the masked diffusion model. At each decoding iteration, the diffusion model proposes tokens for masked positions via a Top-1 sampling strategy. These token predictions are then evaluated by the ARM, which outputs a second set of predictions. Tokens are only unmasked if the predictions from the two models match. This agreement-based criterion provides a lightweight yet effective confidence signal, replacing conventional heuristics such as entropy or logit margin. When multiple tokens align, they are unmasked in parallel; if no match is found, only one token is revealed conservatively. This iterative process continues until all tokens are filled.

The proposed \textit{Guided Diffusion} framework is fast since for each guiding step, both the one-time diffusion process from the drafter and the one-time forward pass of the auto-regressive model are fast. 
Furthermore, the guidance from the auto-regressive model facilitates the coherence of the diffused output tokens with improved semantic logic.
Compared to speculative decoding, Guided Diffusion avoids repetitive speculative-correction loops and achieves lower latency without sacrificing generation quality. Moreover, the guiding ARM is model-agnostic and can be flexibly replaced with stronger or domain-specific models. The framework can also be extended into a correction stage, where ARM predictions overwrite diffusion outputs for additional refinement.


\subsubsection{Step Reduction} \label{sec:distillation}
Knowledge distillation is a technique that allows a student model to mimic the behavior of the teacher model by training the student model to match the teacher model's output distribution. Step reduction can be achieved through knowledge distillation, where a student model with fewer decoding steps is trained to replicate a teacher model's behavior by aligning its output distribution with the teacher's.  

\citet{deschenaux2024beyond} proposes Self-Distillation Through Time (SDTT), a method that distills a high-step diffusion teacher into a low-step student by minimizing the divergence between their output distributions. Compared to the distillation methods for continuous diffusion models, distillation for masked diffusion models lacks a deterministic mapping, e.g., probability flow ODE (PF-ODE); therefore, the distillation target must be generated from stochastic trajectories.
Specifically, the teacher model $p_T(x_0 \mid x_t)$ is trained using the standard denoising objective across a large number of timesteps $m$, while the student model $p_S(x_0 \mid x_t)$ is trained to mimic the teacher's output using a divergence loss such as Kullback-Leibler divergence (KLD), Total Variation Distance (TVD), and MeanSquared Error (MSE).  

SDTT is formulated as follows:
Let $p^{(m)}_\theta$ be the distribution of samples generated with $m$ steps, using a denoiser with parameters~$\theta$.
SDTT trains a denoiser with parameters~$\nu$ to minimize a divergence~$d$ between $p^{(m)}_\theta$ and $p^{(k)}_\nu$:
\begin{equation}
\min_{\nu} \; d\left(p^{(k)}_\nu \,\|\, p^{(m)}_\theta\right).
\end{equation}
where $d$ is a divergence measure such as KLD, TVD, or MSE.

This SDTT process allows the student to generate high-quality predictions using significantly fewer steps $k$ (e.g., half of the teacher's steps, $k=m/2$), enabling large-scale parallel decoding. To further reduce the number of decoding steps, the SDTT procedure can be applied iteratively, using the newly distilled student as the teacher for the next round—referred to as \textit{iterated SDTT}.
In practice, \citet{deschenaux2024beyond} choose $m = 2^{10}$ and $k_i = 2^{10 - i}$ with $0 \leq i \leq 7$ and sequentially minimize the objective
\begin{equation}
\min_{\nu} \; d\left(p^{(k_{j+1})}_{\nu_{j+1}} \,\|\, p^{(k_j)}_{\nu_j} \right),
\end{equation}
for $0 \leq j < 7$, where $\nu_j$ denotes the parameters of the $j$-th denoiser, with $\nu_0 = \theta$ (teacher).

Beyond SDTT, several concurrent works explore conceptually similar distillation-based step-reduction for other forms of parallel decoding. Although not strictly designed for masked diffusion, they share the same objective of accelerating iterative parallel decoding via teacher-student distillation, offering insights that may inspire future adaptations for masked diffusion.
\citet{kou2024cllms} introduce \textit{Consistency Learning for Large Language Models} (CLLM), designed for autoregressive LLMs equipped with \textit{Jacobi decoding}~\cite{santilli2023accelerating}, which applies a Jacobi fixed-point iteration method to iteratively map a randomly initialized $n$-token sequence to the ground truth.  
Crucially, although the underlying model is autoregressive rather than diffusion-based, the iterative trajectory of Jacobi decoding implicitly defines a parallel decoding path that is guaranteed to converge to the correct sequence under mild conditions. Inspired by consistency models in continuous diffusion~\cite{song2023consistency,song2023improved,lu2024simplifying}, CLLM introduces a consistency training objective that enforces either \textit{local consistency}, ensuring that adjacent points along the decoding trajectory produce consistent predictions, or \textit{global consistency}, requiring that the prediction at any intermediate state matches the final converged output. By aligning the student’s predictions with the fixed-point solution across multiple stochastic forward corruptions, CLLM enables substantial step reduction while maintaining competitive perplexity.
Similarly, \citet{sahoo2025diffusion} proposes \textit{Duo}, a framework specifically designed for uniform-state discrete diffusion~\cite{austin2021structured,campbell2022continuous,lou2024discrete}. Duo integrates \textit{Discrete Consistency Distillation} (DCD), which adapts consistency training from Gaussian diffusion to the uniform-state setting. Unlike SDTT, which matches teacher and student only at the final denoising step, DCD enforces consistency across intermediate steps, ensuring that the student remains robust to forward corruption throughout the generation process. Combined with a Gaussian-guided curriculum learning strategy, DCD achieves up to 100$\times$ faster sampling while outperforming autoregressive baselines on several language modeling benchmarks. 
Several recent works further explore aggressive step reduction for discrete diffusion language models via trajectory-level supervision. 
\citet{qian2026d3llm} propose \textit{d3LLM}, which introduces \emph{Pseudo-Trajectory Distillation} to compress a long diffusion process into a small number of coarse denoising steps by training a student model to imitate a deterministically constructed teacher trajectory. 
Similarly, \citet{d2f} demonstrates that \textit{Discrete Diffusion Forcing} can achieve faster-than-autoregressive inference by explicitly training the model to map partially denoised states to later diffusion states, implicitly distilling long diffusion trajectories into short-horizon transitions. 
\textit{dParallel}~\cite{chen2025dparallel} learns a parallel decoder that predicts multiple future diffusion states in a single forward pass via trajectory distillation, amortizing several denoising iterations and further reducing decoding latency.

\subsubsection{System-level Acceleration} \label{sec:kv-cache}
Although diffusion-based large language models (Diffusion LLMs) inherently support parallel decoding, their throughput still lags behind state-of-the-art autoregressive LLMs~\cite{nie2025large,dream2025}. A key reason is the lack of support for key-value (KV) caching. Unlike autoregressive models, which use causal attention and allow KV pairs to be incrementally reused, Diffusion LLMs adopt a mask-predict mechanism with bidirectional attention. Consequently, at each decoding step, all key-value pairs must be recomputed, preventing direct reuse for future updates.  
Recent works~\cite{wu2025fast,ma2025dkv,hu2025accelerating} address this limitation by leveraging the empirical observation that KV activations exhibit high similarity across adjacent inference steps. Motivated by this, \citet{wu2025fast} adopts a block-wise decoding strategy~\cite{arriola2025block} to enable KV caching. Specifically, the KV cache for the prompt is computed once and reused throughout the same block. Within each block, the cache is reused across multiple decoding steps. After completing the decoding of a block, the cache is refreshed by updating all tokens—rather than only the newly generated ones—thereby maintaining consistency for subsequent blocks. Other research introduces KV caching to Masked Diffusion Models by creating hybrid approaches with AR models~\cite{arriola2025block,sahoo2025esoteric,huang2025ctrldiff,liu2025wedlm,cheng2025sdar,d2f,gat2025set}. \citet{sahoo2025esoteric} proposes a two-stage sampling process: first, an MDM generates a partially masked sequence, and then an AR model completes it. The second, autoregressive stage naturally supports KV caching. To enable KV caching in the first stage as well, the model is trained to avoid using bidirectional attention over masked tokens. dInfer~\cite{ma2025dinfer} introduces a specialized inference system optimized for the unique computational patterns of diffusion models, while Lopa~\cite{xu2025lopa} improves throughput by adapting lookahead parallel decoding strategies to scale diffusion-based generation.

\subsection{Edit-Based Refinement} \label{sec:ebr}

Edit-based refinement approaches generate sequences by iteratively editing an initial draft, rather than producing the output in a single pass or by masking. These methods are inspired by the human process of writing, where a rough draft is incrementally refined through insertions, deletions, and substitutions. The iterative nature of these models allows for dynamic sequence length adjustment and the correction of errors in a non-monotonic fashion—capabilities that are challenging for strictly autoregressive or fully non-autoregressive methods. Existing edit-based refinement models can be broadly classified into three distinct approaches:
\begin{itemize}
    \item \textbf{Discrete Edition}: These models directly models the editing process by learning to apply a discrete set of human-like operations such as insertion, deletion, and replacement to the token sequence. 
    \item \textbf{Continuous Optimization}: These models shifts the refinement process from the discrete token space to a continuous latent space, where an entire sequence representation is iteratively optimized, typically via gradient-based methods, before being decoded into the final output.
    \item \textbf{Hybrid Refinement}: These models encompasses a diverse range of methods that either combine different generation paradigms, such as mixing autoregressive and non-autoregressive steps, or decompose the task into specialized stages, like first locating errors and then separately revising them.
\end{itemize}

\subsubsection{Discrete Edition} \label{sec:ebr_eeo}
This category of models defines a discrete set of discrete edit operations—such as insertion, deletion, and replacement—which are learned and applied iteratively. These models directly mimic the intuitive human process of text editing, offering high interpretability and control over the generation process.

A foundational work in this area is the Insertion Transformer \cite{stern2019insertion}, which generates sequences by iteratively inserting tokens into a partial hypothesis. This model accommodates arbitrary generation orders and can be trained to follow specific orderings, such as left-to-right or a binary tree traversal, demonstrating flexibility in both fully and partially autoregressive decoding. Building on this, the Levenshtein Transformer \cite{gu2019levenshtein} introduced a model capable of both insertion and deletion operations, allowing for dynamic changes in sequence length. This dual-operation framework has proven effective not only for generation tasks like machine translation and summarization but also for refinement tasks such as automatic post-editing.

Subsequent research has expanded on these core ideas. EDITOR \cite{xu2021editor} enhanced the Levenshtein Transformer by introducing a novel "reposition" operation, which disentangles lexical choice from word positioning. This allows for more effective use of soft lexical constraints and achieves faster decoding speeds. FELIX \cite{mallinson2020felix} decomposed the editing task into two non-autoregressive sub-tasks: a tagging model with a pointer mechanism to select and reorder input tokens, and an insertion model based on a Masked Language Model to add new tokens. This design proved efficient, especially in low-resource settings.

The principles of edit-based refinement have been successfully applied to other domains. For instance, Levenshtein OCR (LevOCR) \cite{da2022levenshteinocr} adapted the iterative refinement process for scene text recognition, casting it as a sequence of deletion and insertion operations on an initial prediction from a vision model. Similarly, FastCorrect \cite{leng2022fastcorrectfasterrorcorrection} proposed a non-autoregressive error correction model for Automatic Speech Recognition (ASR) based on edit alignment. It learns to predict the number of tokens to insert, delete, or substitute for each source token, enabling parallel generation and achieving significant latency reduction with minimal impact on accuracy. More recently, \cite{zhang2023non} proposed a method for text editing using latent CTC alignments, which was extended to include a copy operation in the edit space. This approach efficiently handles textual overlap and demonstrates strong performance and generalizability on tasks like Grammatical Error Correction (GEC) and sentence fusion.

Researchers have also explored advanced training strategies for these models. \cite{wang-etal-2024-reinforcement} applied reinforcement learning to the Levenshtein Transformer, demonstrating that training with self-generated data can mitigate exposure bias and improve performance by optimizing for either stepwise or episodic rewards. For text summarization, EditKSum \cite{liang2024summarizing} proposed using prominent keywords from the source document as an initial draft, which is then refined through repositioning, inserting, and deleting operations, implicitly addressing the challenging problem of length prediction in summarization.

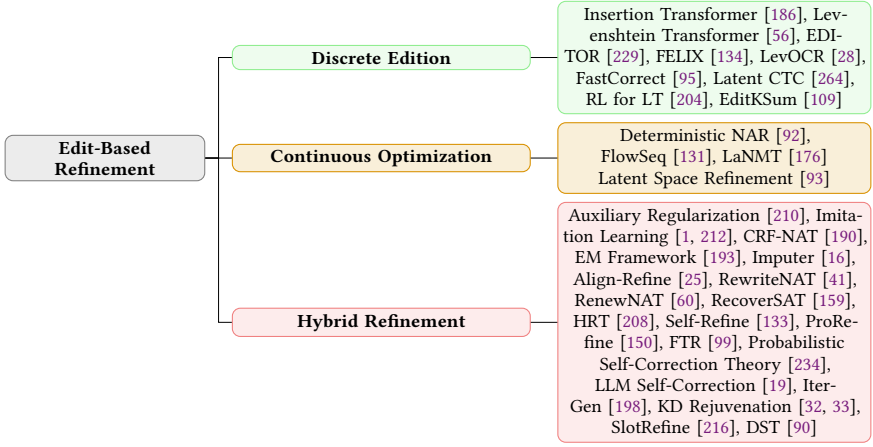
\begin{figure}[htbp]
	\centering
	\tikzset{
		my node/.style={
			draw,
			align=center,
			thin,
			text width=1.2cm, 
			rounded corners=3,
		},
		my leaf/.style={
			draw,
			align=center,
			thin,
			text width=8.5cm, 
			rounded corners=3,
		}
	}
	\forestset{
		every leaf node/.style={
			if n children=0{#1}{}
		},
		every tree node/.style={
			if n children=0{minimum width=1em}{#1}
		},
	}
	\begin{forest}
		nonleaf/.style={font=\bfseries\scriptsize},
		for tree={%
			every leaf node={my leaf, font=\scriptsize},
			every tree node={my node, font=\scriptsize, l sep-=4.5pt, l-=1.pt},
			anchor=west,
			inner sep=2pt,
			l sep=10pt, 
			s sep=3pt, 
			fit=tight,
			grow'=east,
			edge={ultra thin},
			parent anchor=east,
			child anchor=west,
			if n children=0{}{nonleaf}, 
			edge path={
				\noexpand\path [draw, \forestoption{edge}] (!u.parent anchor) -- +(5pt,0) |- (.child anchor)\forestoption{edge label};
			},
			if={isodd(n_children())}{
				for children={
					if={equal(n,(n_children("!u")+1)/2)}{calign with current}{}
				}
			}{}
		}
		[Edit-Based Refinement, draw=gray, fill=gray!15, text width=2.5cm, text=black
		[Discrete Edition, color=lightgreen, fill=lightgreen!15, text width=3.8cm, text=black
		[{Insertion Transformer~\cite{stern2019insertion},
			Levenshtein Transformer~\cite{gu2019levenshtein}, EDITOR~\cite{xu2021editor}, FELIX~\cite{mallinson2020felix},
			LevOCR~\cite{da2022levenshteinocr}, FastCorrect~\cite{leng2022fastcorrectfasterrorcorrection},
			Latent CTC~\cite{zhang2023non}, RL for LT~\cite{wang-etal-2024-reinforcement}, EditKSum~\cite{liang2024summarizing}},
		color=lightgreen, fill=lightgreen!15, text=black, text width=4.0cm
		]
		],
		[Continuous Optimization, color=harvestgold, fill=harvestgold!15, text width=3.8cm, text=black
		[{Deterministic NAR~\cite{lee-etal-2018-deterministic}, FlowSeq~\cite{ma2019flowseq}, LaNMT~\cite{shu2020latent}
			Latent Space Refinement~\cite{lee2020iterative}}, 
		color=harvestgold, fill=harvestgold!15, text=black, text width=4.0cm]
		],
		[Hybrid Refinement, color=lightcoral, fill=lightcoral!15, text width=3.8cm, text=black
		[{Auxiliary Regularization~\cite{wang2019non},
			Imitation Learning~\cite{wei2019imitation, agrawal2022imitation}, 
			CRF-NAT~\cite{sun2019fast}, EM Framework~\cite{sun2020approach},
			Imputer~\cite{chan2020imputer}, Align-Refine~\cite{chi2020alignrefinenonautoregressivespeechrecognition}, RewriteNAT~\cite{geng2021learning}, RenewNAT~\cite{guo2023renewnat}, RecoverSAT~\cite{ran2020learning}, HRT~\cite{wang2022hybrid}, 
            Self-Refine~\cite{madaan2023selfrefine}, ProRefine~\cite{pandita2025prorefine}, FTR~\cite{li2025ftr}, Probabilistic Self-Correction Theory~\cite{yang2025selfcorrection}, LLM Self-Correction~\cite{chen2024iterativetranslationrefinementlarge}, IterGen~\cite{ugare2025itergeniterativesemanticawarestructured},
			KD Rejuvenation~\cite{ding2021rejuvenating, ding2021understandingimprovinglexicalchoice},
			SlotRefine~\cite{wu2020slotrefine}, DST~\cite{le2020non}},
		color=lightcoral, fill=lightcoral!15, text=black, text width=4.0cm]
		]
		]
	\end{forest}
	\caption{Taxonomy of edit-based refinement methods}
	\label{fig:taxonomy_ebr}
\end{figure}

\subsubsection{Continuous Optimization} \label{sec:ebr_cso}
Instead of operating in the discrete token space, this class of models performs iterative refinement within a continuous, low-dimensional latent space. An initial sequence is encoded into a latent representation, which is then progressively optimized using techniques like gradient-based methods. The final sequence is decoded from the refined latent variable. This approach abstracts the editing process, often leading to more efficient and effective optimization.

An early model in this vein, proposed by \cite{lee-etal-2018-deterministic}, introduced a deterministic non-autoregressive model based on iterative refinement, designed on the principles of latent variable models and denoising autoencoders. The model iteratively refines the entire output sequence by feeding it back as input, progressively improving generation quality. FlowSeq \cite{ma2019flowseq} utilized generative flows to model the conditional density of sequential latent variables, enabling efficient non-autoregressive generation with almost constant decoding time.

More recent works have focused on explicitly optimizing latent variables. LaNMT \cite{shu2020latent} proposed a latent-variable NAR model with a deterministic inference procedure that finds the target sequence by maximizing the evidence lower bound (ELBO). During inference, the model iteratively updates the latent variables, which automatically adapts the translation length. Following this, \cite{lee2020iterative} proposed an even more efficient inference procedure that refines the translation purely in the continuous latent space. The method trains an inference network to approximate the gradient of the marginal log probability, enabling gradient-based optimization of the latent variable. It proves faster and more effective than hybrid-space optimization, significantly narrowing the performance gap with autoregressive models while achieving substantial speedups.

\subsubsection{Hybrid Refinement}\label{sec:ebr_hr}
Hybrid models combine elements from different paradigms, such as autoregressive and non-autoregressive generation, or decompose the refinement process into distinct stages. These approaches aim to leverage the strengths of multiple strategies to achieve a better balance between speed, quality, and modeling flexibility.

Several works focus on improving the training or decoding process of NAR models by incorporating auxiliary objectives or multi-step procedures. \cite{wang2019non} addressed error patterns in NAT by introducing two auxiliary regularization terms during training: one to ensure distinguishability between adjacent hidden states and another to enforce informational completeness via a backward reconstruction loss. \cite{sun2019fast} improved decoding consistency by incorporating an efficient approximation of a Conditional Random Field (CRF) into a non-autoregressive model, using a dynamic transition technique to model positional contexts. \cite{wei2019imitation} framed NAT as an imitation learning problem, which allows the model to consider context from previous decoding steps while maintaining parallel generation speed. Similarly, \cite{sun2020approach} proposed a unified Expectation-Maximization (EM) framework that jointly optimizes an autoregressive "teacher" model and a non-autoregressive "student" model, where the AR model helps remove multi-modality for the NAR model.

Other hybrid approaches redesign the generation process itself. Imputer \cite{chan2020imputer} introduced a model that generates sequences via iterative imputations, requiring only a constant number of generation steps. It can be trained to marginalize over all possible alignments and generation orders using a dynamic programming algorithm. Align-Refine \cite{chi2020alignrefinenonautoregressivespeechrecognition} proposed iterative realignment for speech recognition, where refinements are applied to latent CTC alignments rather than the output sequence, enabling length-changing edits. RewriteNAT \cite{geng2021learning} explicitly learns to rewrite translations by using a "locator" module to identify erroneous parts and a "revisor" module to correct them, both trained with an iterative strategy to mimic multi-step decoding. RenewNAT \cite{guo2023renewnat} introduced a two-stage framework that first generates a potential translation using a fully NAT model and then renews it in a single pass—improving performance without increasing latency.

The integration of autoregressive and non-autoregressive steps is another prominent hybrid strategy. RecoverSAT \cite{ran2020learning} proposed a semi-autoregressive model that generates a sequence of segments in parallel, where each segment is predicted token-by-token. By dynamically determining segment lengths, it can recover from common NAT errors like repetition. Hybrid-Regressive Translation (HRT) \cite{wang2022hybrid} formalized a two-stage prototype: it first generates a sparse, discontinuous sequence autoregressively (e.g., every k-th token) and then fills in the missing tokens non-autoregressively in a single step. This approach inherits the robustness of AR models while achieving significant speedups.

Recent work has also explored iterative refinement with Large Language Models (LLMs). ProRefine~\cite{pandita2025prorefine} performs inference-time prompt refinement via an agentic loop of LLMs that generate and apply textual feedback, significantly improving mathematical reasoning over zero-shot chain-of-thought. Feedback-Triggered Regeneration (FTR)~\cite{li2025ftr} triggers regeneration only on negative user feedback to avoid error propagation and introduces Long-Term Multipath decoding to explore multiple reasoning trajectories, improving mathematical and code-generation benchmarks. A probabilistic inference scaling theory~\cite{yang2025selfcorrection} models how accuracy evolves over self-correction rounds ($\mathrm{Acc}_t = U_{\mathrm{pp}} - \alpha^t(U_{\mathrm{pp}} - \mathrm{Acc}_0)$), providing a theoretical basis for iterative refinement. Other work includes iteratively prompting for self-correction of translation~\cite{chen2024iterativetranslationrefinementlarge} and IterGen~\cite{ugare2025itergeniterativesemanticawarestructured} for grammar-guided generation with backtracking.

Finally, a significant body of work has focused on empirical analysis and curriculum learning to improve edit-based models. \cite{ding2021rejuvenating} and \cite{ding2021understandingimprovinglexicalchoice} analyzed the issue of knowledge distillation (KD) causing errors on low-frequency words in NAT. They proposed exposing models to raw data and using reverse KD to rejuvenate these words, significantly boosting performance. \cite{agrawal2022imitation} identified a train-inference mismatch in imitation learning for text editing and proposed a curriculum that starts with easy edit operations and gradually increases difficulty, improving generalization. An empirical study~\cite{xiao2025empirical} analyzed existing iterative NAR models and proposed an efficient refinement strategy that achieves state-of-the-art performance with fewer decoding steps. This body of work highlights the ongoing effort to understand, diagnose, and systematically improve the complex dynamics of iterative refinement. This paradigm has also been applied to tasks like joint intent detection and slot filling \cite{wu2020slotrefine} and dialogue state tracking \cite{le2020non}, demonstrating its versatility.

\section{Comparison of Acceleration Paradigms}\label{sec:comparison}

Building on the preceding analysis of parallel text generation methods, this section provides a detailed theoretical comparison from three perspectives: (1) the standalone strengths and trade-offs of each paradigm, (2) the potential for combining methods to achieve greater acceleration, and (3) comparisons with non-parallel acceleration techniques. In this section, we primarily analyze empirical data summarized from related work, while a more detailed theoretical analysis is provided in Appendix~\ref{sec:appendix-comparison}.

\begin{table}[h]
\centering
\caption{Qualitative comparison of six parallel generation methods in terms of speedup, resource usage, and output quality.}
\setlength{\tabcolsep}{3.5pt}
\begin{tabular}{l|c|cc|c}
\toprule
\multirow{2}{*}{\textbf{Parallelism Strategy}} & \multirow{2}{*}{\textbf{Speedup}} & \multicolumn{2}{c|}{\textbf{Resource Usage}} & \multirow{2}{*}{\textbf{Quality}} \\
\cline{3-4}
 & & \textbf{Memory} & \textbf{Computation} &  \\
\midrule
Draft-and-Verify        & Medium (1.8x-2.4x)~\cite{li2024eagle, xia2024unlocking} & Medium & High   & High \\
Decomposition-and-Fill       & Medium (2.4x-5.7x)~\cite{kolawole2025parallelprompt} & Medium & Medium & Medium \\
Multiple Token Prediction & Medium (1.8x-3.6x)~\cite{monea2023pass, cai2024medusa}   & Low    & Low    & High \\
One-Shot Generation      & High (10x-16.5x)~\cite{gu2020fullynonautoregressiveneuralmachine}   & Medium & Low    & Low \\
Masked Generation        & High (up to 32x)~\cite{dream2025,lou2024discrete} & High   & High   & Medium \\
Edit-Based Refinement    & Low (1.2x-3x)~\cite{xu2021editor, gu2019levenshtein}    & Low    & High   & High \\
\bottomrule
\end{tabular}
\label{tab:qualitative-analysis}
\end{table}

\subsection{Standalone Analysis: Speed, Quality, and Resource}

In parallel text generation, faster generation often comes at the cost of increased resource consumption and potential quality degradation. To enable a fair comparison, we first formally define the theoretical speed-up potential, output quality, and resource consumption of each parallel generation paradigm when applied independently.

Based on the analysis of recent works, the qualitative comparison results are summarized in Table~\ref{tab:qualitative-analysis}. Below is a more detailed discussion of each paradigm, with emphasis on how quality and resource usage are influenced by their design.

Draft-and-Verify: This method achieves medium speedup by leveraging a lightweight draft model to propose tokens and a heavier verifier to selectively accept them. Quality is generally high because the verifier enforces correctness, but errors may propagate if the acceptance threshold is miscalibrated. Resource usage is unbalanced: memory usage is moderate (due to maintaining caches for two models), but computational cost is high, as both models must run—sometimes sequentially—at every generation step. Recent systems ~\cite{li2024eagle, xia2024unlocking} report speedups of 1.8x-2.4x.

Decomposition-and-Fill: By splitting input into $n$ independent segments, this approach reduces dependency chains and enables parallel processing, leading to medium speedup. Quality is medium because local segment accuracy is good, but global coherence can degrade when cross-segment dependencies are strong. Resource consumption is also moderate: each worker handles only a small segment, keeping the memory footprint manageable,, but overall computation scales approximately linearly with $n$. Speedup of 2.4x-5.7x has been observed~\cite{kolawole2025parallelprompt}.

Multiple Token Prediction (MTP): MTP predicts $k$ tokens at a time, reducing sequential steps and achieving high speedup. Quality degrades as $k$ increases because the model struggles to capture interactions among jointly predicted tokens, often resulting in incoherence or factual errors. Resource usage is favorable: both memory and computation per step are low, as only one forward pass is required for every $k$ tokens, making it attractive for throughput-oriented scenarios. Reported speedups are 1.8–3.6x~\cite{monea2023pass, cai2024medusa}.

One-shot Generation: This paradigm outputs the entire sequence in a single pass, achieving very high speedup with minimal computation cost per output. However, quality is low: without autoregressive conditioning, outputs often exhibit semantic drift, hallucinations, and poor factuality, especially on complex tasks. Resource usage is modest, as only one forward pass is needed, with no iterative refinement or caching. Typical speedups are 10–16.5x~\cite{gu2020fullynonautoregressiveneuralmachine}.

Masked Generation: 
In this approach, output sequences are generated iteratively by filling in masked positions, with the option to refine previously generated erroneous tokens in subsequent rounds. Recent advances in MDMs have achieved medium output quality (comparable to LLaMA-2 with the same size~\cite{nie2024scaling,dream2025,zhu2025llada}) by decoding multiple tokens in parallel while only minimally violating conditional dependencies, provided that a sufficient number of decoding rounds ($R$) are performed.
However, resource consumption is high: memory usage increases since the model predicts all the masked tokens at each decoding step, and computational cost scales approximately linearly with $R$. The achievable speedup depends on the number of tokens generated per step, which is governed by the sampling time schedule. This parameter can be dynamically adjusted, offering a tunable trade-off between speed and quality: fewer steps enable faster but coarser outputs, while more steps improve fidelity at the cost of higher computation. Recent studies \citep{dream2025,lou2024discrete} report that masked diffusion models can achieve a 10-32x speedup over autoregressive models without noticeable quality degradation under well-optimized configurations.

Edit-Based Refinement: This approach focuses on iterative edits to improve a draft’s accuracy and fluency. Quality is high because each pass targets specific errors, but it depends heavily on the effectiveness of the editing model. Resources are heavily tilted toward computation: multiple passes incur high computational cost, while memory usage remains low. Speedup is minimal (1.2–3x), as the editing overhead can offset the parallel gains from the initial draft~\cite{xu2021editor, gu2019levenshtein}. Empirical efficiency numbers remain to be verified.

These refined analysis make clear that the quality–resource–speed trade-off is intrinsic to each paradigm. Approaches that maximize speed (MTP, One-shot) tend to compromise quality, while quality-focused methods (Masked, Edit-Based) require significant compute. Balanced strategies (Draft-and-Verify, Decomposition-and-Fill) show promise but remain sensitive to task characteristics and hardware configurations.

\subsection{Promising Combinations}

While each parallel generation paradigm provides independent acceleration, further gains may be achievable by combining multiple techniques in a single decoding pipeline. In this section, we analyze which combinations among the six paradigms—\textbf{One-shot Generation}, \textbf{Decomposition-and-Fill}, \textbf{Masked Generation}, \textbf{Edit-Based Refinement}, \textbf{Draft-and-Verify}, and \textbf{Multiple Token Prediction (MTP)}—can be theoretically or already composed for greater speedup.

\begingroup
\setlength{\tabcolsep}{9pt}
\begin{table}[h]
\centering
\begin{tabular}{l|cccccc}
\toprule
 & One-shot & Decomp.-Fill & Masked & Edit & Draft-Verify & MTP \\
\midrule
One-shot       & \cellcolor{gray!10} & \cellcolor{gray!10} & \cellcolor{gray!10} & \cellcolor{gray!10} & \cellcolor{gray!10} & \cellcolor{gray!10} \\
Decomp.-Fill   & \checkmark & \cellcolor{gray!10} & \cellcolor{gray!10} & \cellcolor{gray!10} & \cellcolor{gray!10} & \cellcolor{gray!10} \\
Masked         & $\times$ & \S \ref{sec:daf-all} & \cellcolor{gray!10} & \cellcolor{gray!10} & \cellcolor{gray!10} & \cellcolor{gray!10} \\
Edit           & \checkmark & \S \ref{sec:daf-all} & \S \ref{sec:mg-ebf} & \cellcolor{gray!10} & \cellcolor{gray!10} & \cellcolor{gray!10} \\
Draft-Verify   & $\times$ & \S \ref{sec:daf-all} & \S \ref{sec:dav-all} & \ref{sec:dav-all} & \cellcolor{gray!10} & \cellcolor{gray!10} \\
MTP            & $\times$ & \S \ref{sec:daf-all} & $\times$ & \checkmark & \S \ref{sec:dav-all} & \cellcolor{gray!10} \\
\bottomrule
\end{tabular}
\caption{Composability matrix of parallel generation paradigms. Upper-triangular cells are shaded to indicate symmetry. 
Here, \checkmark\ denotes a feasible combination, $\times$ an incompatible one, and entries with a section reference (\S) highlight combinations analyzed in detail as particularly promising.}
\label{tab:composability-matrix}
\end{table}
\endgroup

The pairwise composability of different methods is shown in Table~\ref{tab:composability-matrix}. In this table, \checkmark\ indicates combinations that are feasible, and $\times$ marks combinations that are incompatible. Cells annotated with a section number (\S) correspond to combinations that are both compatible and identified as either highly promising or particularly popular, which we analyze in more detail in the corresponding sections.

\subsubsection{Combinations Involving Draft-and-Verify} \label{sec:dav-all}

Draft-and-Verify serves as a versatile backbone that synergizes with multiple parallel generation paradigms to balance speed and quality. Its core principle—using a lightweight draft model to propose tokens followed by a main model to verify and accept them—naturally complements other strategies that suffer from quality degradation as parallelism is increased.

One of the most commonly adopted pairings is with Multiple Token Prediction (MTP). As discussed in Section~\ref{sec:mtp}, nearly all MTP techniques incorporate Draft-and-Verify to mitigate the accuracy loss associated with predicting multiple tokens simultaneously~\cite{liu2025mtp, mehra2025multi, cai2024medusa, gerontopoulos2025multi, stern2018blockwise, monea2023pass, li2024eagle, samragh2025your, qi2020prophetnet, gloeckle2024better, liu2024deepseek, xiaomi2025mimo}. In this setup, a draft model generates $k$ tokens per step, while the main model selectively accepts or rejects them. This hybrid achieves significant acceleration by reducing decoding steps by a factor of $k$ while preserving fidelity through verification. For example, MTAD~\cite{qin2024mtad} reports about 1.4× speedup and energy savings over pure speculative decoding, with reduced perplexity and improved model performance. Resource usage remains moderate: although two model passes are involved, fewer decoding iterations keep the overall cost manageable.

A second promising combination is with Masked Generation, particularly in the context of diffusion-based LLMs where autoregressive dependencies are absent and quality control becomes critical. Integrating Draft-and-Verify enables speculative token drafting, followed by masked generation refinement to correct rejected or low-confidence spans~\cite{christopher2025speculative, deaccelerated, hu2025accelerating}. SpecDiff~\cite{christopher2025speculative} demonstrates the potential of this approach, achieving up to 7.2× speedup over standard autoregressive decoding and 1.75× over pure speculative decoding, while maintaining high quality via diffusion-style masked refinement. The approach combines span-level parallelism with powerful corrective iterations but comes at the cost of high GPU memory usage and increased system complexity.

Lastly, Draft-and-Verify can also integrate with Edit-Based Refinement to further enhance generation quality. In this pipeline, speculative verification quickly filters out low-confidence tokens, after which an edit model iteratively corrects residual errors. Although less explored in existing literature, this combination is conceptually appealing: speculative decoding accelerates generation, while edit-based refinement ensures fine-grained semantic coherence. The trade-off lies in resource overhead, as iterative edits add extra passes, and coordinating verification with edits increases pipeline complexity. Nonetheless, this design holds promise for tasks requiring both efficiency and high accuracy.

Overall, combinations involving Draft-and-Verify effectively leverage its verification mechanism to compensate the weaknesses of other acceleration paradigms. Whether paired with MTP for span-level parallelism, masked generation for robust refinement, or edit-based refinement for high-fidelity outputs, Draft-and-Verify consistently provides a balanced trade-off between speed, quality, and resource usage, making it a central strategy in the landscape of parallel text generation.

\subsubsection{Combinations Involving Decomposition-and-Fill} \label{sec:daf-all}

Since Decomposition-and-Fill inherently partitions generation into independent units, it can be flexibly combined with almost all other parallel text generation paradigms. This is because its segmentation-based strategy serves as a natural wrapper: once the input is divided into $n$ semantically coherent segments—guided by outlines, key phrases, or structural cues—any downstream parallel decoding method can be applied to each segment independently. For example, segments can be filled using Masked Generation, verified through Draft-and-Verify, accelerated with Multiple Token Prediction, or refined using Edit-Based strategies, all without altering the initial decomposition logic.

The advantages of this universal combinability are evident. Speed-up arises from segment-level parallelism, where segments can be decoded simultaneously across GPUs or threads, and from the inherent acceleration provided by the chosen secondary paradigm (e.g., MTP’s multi-token steps or masked generation’s rapid refinement). Quality also benefits when decomposition preserves global coherence, while the applied decoding method ensures strong local accuracy within each segment. However, resource usage tends to grow because each segment invokes its own decoding pipeline—potentially involving multiple forward passes or verification stages—and executing all segments in parallel may lead to high aggregate memory and computation costs. Careful scheduling, resource balancing, and batched execution are therefore essential to make this approach practical.

Although few works have explicitly explored an end-to-end pipeline that integrates Decomposition-and-Fill with multiple parallel paradigms simultaneously, research in outline-conditioned generation, speculative verification, and text infilling demonstrates the feasibility of such hybrid designs~\cite{li2024advancing, christopher2025speculative, qin2024mtad}. These findings suggest that Decomposition-and-Fill can serve as a powerful backbone for unifying diverse parallel generation strategies, opening a promising direction for future work.

\subsubsection{Masked Generation + Edit-Based Refinement} \label{sec:mg-ebf}

This combination leverages masked generation to quickly produce an initial draft through parallel span infilling, followed by edit-based refinement that incrementally corrects errors and improves fluency. Masked generation, particularly in diffusion-inspired approaches, can rapidly generate the coarse structure of text but often leaves inconsistencies or local inaccuracies due to its non-autoregressive nature. Edit-based refinement complements this by applying targeted insertions, deletions, and substitutions, progressively transforming the draft into a polished final output without re-decoding the entire sequence.

Although this pairing has been explored only sparingly, a few recent works demonstrate its feasibility. For instance, Seed-Diffusion~\cite{song2025seed} integrates diffusion-based masked generation with iterative edit refinement, showing that edits can effectively correct structural and semantic errors left by the initial draft. Similar approaches, though rare, suggest that combining the high-speed, non-autoregressive parallelism of masked generation with the fine-grained corrective capabilities of edit-based refinement can deliver both efficiency and quality. In such pipelines, quality remains high because local errors are systematically eliminated and semantic coherence is reinforced through iterative edits.

However, the approach is resource-intensive: masked generation rounds require full-sequence processing, and multiple edit passes add further computational and memory overhead. Moreover, coordinating two distinct decoding mechanisms increases implementation complexity, which may hinder real-world deployment despite the clear potential benefits.

\subsection{Beyond Parallel Generation: Compatibility with Other Acceleration Techniques}

This section examines how parallel generation paradigms interact with a broader class of acceleration techniques operating at the model, system, and architectural levels. We discuss their compatibility, qualitative speedup potential, and limitations arising from decoding assumptions.

\subsubsection{Model Compression}
Model compression techniques—including distillation, quantization, and pruning—are fully compatible with all six parallel generation paradigms, as they do not alter the decoding logic. When applied alongside speculative decoding or other parallel strategies, they can significantly amplify overall speedups, as the compressed (and hence faster) model propagates improvement across multiple decoding rounds.

A prime example of this synergy is DistillSpec~\cite{zhou2023distillspec}, which improves speculative decoding by applying knowledge distillation to better align the draft model with the target model, resulting in a 10–45\% speedup over standard speculative decoding—even before applying compression on the draft model itself . In practical implementations, distillation-enabled draft models can reduce decoding latency by 6–10x with minimal quality loss~\cite{khoshnoodi2024comprehensive}.

Furthermore, self‑speculative decoding methods (e.g., “Draft \& Verify” with selective layer skipping) eliminate auxiliary draft models altogether and yet benefit from smaller or pruned models in the verification step, yielding nearly 2x speedups with no additional model size overhead~\cite{zhang2024draft}.

While aggressive quantization or pruning (e.g., AWQ, GPTQ) enable up to 2–8x inference acceleration~\cite{lin2024awq, frantar2022gptq}, their expressivity loss can impair paradigms like One-shot Generation or MTP that lack corrective refinement. Consequently, compression is most effective when combined with correction-based methods (e.g., Edit-Based Refinement or Draft-and-Verify), where potential quality degradation can be recovered via downstream verification or editing steps.

\subsubsection{Caching (KV Reuse)}
KV caching accelerates decoding by reusing previously computed key–value (KV) pairs in attention layers, thus avoiding redundant computations during sequential token generation. By design, caching is inherently effective for all autoregressive (AR) paradigms—such as Edit-Based Refinement and Draft-and-Verify—because these methods generate tokens incrementally, allowing cached states to be directly reused across decoding steps.

In contrast, caching is theoretically incompatible with non-autoregressive (non-AR) paradigms, including One-shot Generation, Decomposition-and-Fill, and classical forms of Masked Generation or MTP, as these approaches either produce outputs in a single forward pass or rely on iterative updates that do not preserve stable token positions across steps.

However, recent works have proposed approximate caching mechanisms to extend its benefits into non-AR settings. For example, LazyMAR~\cite{yan2025lazymar} introduces cache-aware attention with selective KV refresh strategies to enable partial reuse in masked refinement. Similarly, Fast-dLLM~\cite{wu2025fast}, dKV-Cache~\cite{ma2025dkv}, and dLLM-Cache~\cite{liu2025dllm} demonstrate that block-wise KV reuse can be applied in masked or diffusion-style decoding by anchoring caches to stable span positions. For MTP, cache compatibility depends on whether the decoding preserves token order within multi-token predictions, which remains an open challenge. These advancements suggest that while caching fundamentally favors AR pipelines, ongoing innovations are pushing its applicability into broader non-AR paradigms.

\subsubsection{Infra-Level Optimization}
Infrastructure-level optimizations—such as FlashAttention~\cite{daoflashattention}, TensorRT-LLM, and vLLM scheduling—improve kernel efficiency, memory usage, and scheduling without altering generation logic. They are fully compatible with all parallel paradigms. These optimizations can offer substantial throughput gains, especially on long sequences or large batch sizes. However, their benefits diminish for short prompts or low-latency tasks, where overheads like memory bandwidth or kernel launch latency dominate. Nevertheless, these techniques provide “free” performance improvements that stack multiplicatively with other accelerators.

\begin{table}[h]
\centering
\setlength{\tabcolsep}{3pt}
\begin{tabular}{l|cccccc}
\toprule
\textbf{Accelerator} & \textbf{Draft-Verify} & \textbf{Decomp.-Fill} & \textbf{MTP} & \textbf{One-shot} & \textbf{Masked} & \textbf{Edit} \\
\midrule
Model Compression           & \checkmark & \checkmark & \checkmark & \checkmark & \checkmark & \checkmark \\
Caching (KV reuse)          & \checkmark     & \checkmark  & \checkmark   & $\times$  & $\triangle$ & $\triangle$  \\
Infra-Level Optimization    & \checkmark & \checkmark & \checkmark & \checkmark & \checkmark & \checkmark \\
\bottomrule
\end{tabular}
\caption{Compatibility of non-parallel accelerators with parallel generation paradigms. Here, 
\checkmark\ denotes full compatibility, $\times$ indicates incompatibility, and $\triangle$ marks partial compatibility subject to specific decoder designs.}
\label{tab:composability-other-expanded}
\end{table}

As summarized in Table~\ref{tab:composability-other-expanded}, most non-parallel accelerators can be seamlessly integrated with parallel generation paradigms, but their effectiveness varies. Model compression offers universal compatibility and multiplicative speed gains but may reduce quality in the absence of corrective mechanisms. KV caching requires precise alignment to remain effective in non-autoregressive scenarios. Infrastructure-level optimizations provide universal benefits but depend on workload characteristics to reach their full potential. Future research should explore how to unify these accelerators into a cohesive framework that automatically adapts configurations to task constraints, maximizing both speed and quality.

\section{Challenges and Future Directions}
\label{sec:challenges}

\begin{figure}[ht]
\centering
\tikzset{
        my node/.style={
            draw,
            align=center,
            thin,
            text width=1.2cm, 
            rounded corners=3,
        },
        my leaf/.style={
            draw,
            align=left,
            thin,
            text width=8.5cm, 
            rounded corners=3,
        }
}
\forestset{
  every leaf node/.style={
    if n children=0{#1}{}
  },
  every tree node/.style={
    if n children=0{minimum width=1em}{#1}
  },
}
\begin{forest}
    nonleaf/.style={font=\scriptsize},
     for tree={%
        every leaf node={my leaf, font=\scriptsize},
        every tree node={my node, font=\scriptsize, l sep-=4.5pt, l-=1.pt},
        anchor=west,
        inner sep=2pt,
        l sep=10pt, 
        s sep=3pt, 
        fit=tight,
        grow'=east,
        edge={ultra thin},
        parent anchor=east,
        child anchor=west,
        if n children=0{}{nonleaf}, 
        edge path={
            \noexpand\path [draw, \forestoption{edge}] (!u.parent anchor) -- +(5pt,0) |- (.child anchor)\forestoption{edge label};
        },
        if={isodd(n_children())}{
            for children={
                if={equal(n,(n_children("!u")+1)/2)}{calign with current}{}
            }
        }{}
    }
    [\textbf{Challenges} (\S \ref{sec:challenges}), draw=gray, fill=gray!15, text width=1.5cm, text=black
    [\textbf{General Challenges} (\S \ref{sec:general-challenges}), color=brightlavender, fill=brightlavender!15, text width=3cm, text=black
            [Increased Overhead in Implementation and Optimization (\S  \ref{sec:overhead}), color=brightlavender, fill=brightlavender!15, text width=8.0cm, text=black
            ]
            [The Fundamental Trade-off between Quality and Speed (\S  \ref{sec:tradeoff-quality-speed}), color=brightlavender, fill=brightlavender!15, text width=8.0cm, text=black
            ]
        ]
        [\textbf{Technique-Specific Challenges} (\S  \ref{sec:technique-specific}), color=lightgreen, fill=lightgreen!15, text width=3cm, text=black
            [The Challenge of Ignored Dependencies in High-Entropy Scenarios (\S  \ref{sec:ignored-dependencies}), color=lightgreen, fill=lightgreen!15, text width=8.0cm, text=black
            ]
            [Conflicts with the Existing Optimization Ecosystem (\S  \ref{sec:conflicts-ecosystem}), color=lightgreen, fill=lightgreen!15, text width=8.0cm, text=black
            ]
        ]
    ]
\end{forest}
\caption{Taxonomy of Challenges and Future Directions of Parallel Decoding Techniques.}
\label{fig_taxonomy_of_challenges}
\vspace{-3mm}
\end{figure}
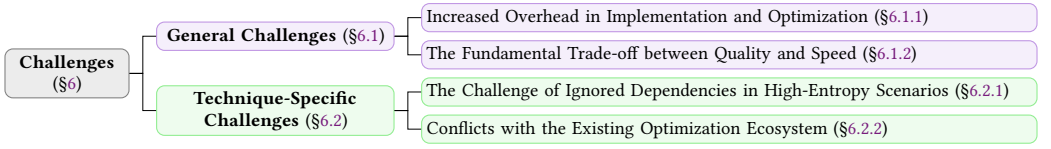

This paper has introduced the classification of parallel decoding techniques, along with comparisons between these techniques and their combinability. This section discusses the challenges of these techniques and future directions. 

Because this paper covers many categories of parallel decoding techniques, some challenges may not apply to all of them. Therefore, we divide the challenges into general challenges (\S \ref{sec:general-challenges}) and technique-specific challenges (\S \ref{sec:technique-specific}). In the latter, we focus on challenges that arise in more than one technique and are closely related to parallel decoding. We do not cover challenges that apply only to a single technique and have limited connection to parallel decoding.

\subsection{General Challenges} \label{sec:general-challenges}

\subsubsection{Increased Overhead in Implementation and Optimization}\label{sec:overhead}

Parallel processing leads to higher system complexity. This is similar to the difference between parallel and serial processing in classic CPUs, where parallel approaches require handling coordination issues, making system implementation more complex \cite{hennessy2011computer}. This can also increase resource usage.

Specific techniques face this issue in different ways. Speculative decoding requires careful selection or training of a draft model that aligns closely with the target model's distribution, and it also needs precise adjustment of the draft length (that is, how many tokens to predict at once). This forms a complex optimization problem \cite{yin2024theoretical}. Decomposition-and-Fill (\S \ref{sec:sfg}) requires analysis of the specific task to design the decoding strategy. Multi-token prediction (\S \ref{sec:mtp}) needs changes to the model architecture, such as adding multiple prediction heads, and new training objective functions to balance relations among predictions. This can cause training instability or raise the risk of error propagation. For all Non-AR-based methods (\S \ref{sec:Non-AR}), the entire pre-training, post-training, and inference processes require complex changes, which are certainly more complicated than those for AR-based methods (\S \ref{sec:AR}).

Parallel methods generally trade off increased system complexity for speed, but this added complexity leads to higher costs and risks. Future work should aim to improve speed while maintaining simplicity in design.

\subsubsection{The Fundamental Trade-off between Quality and Speed} \label{sec:tradeoff-quality-speed}

Almost all parallel decoding techniques trade generation quality for speed to some degree, which can appear in different aspects. For example, in speculative decoding, relaxing the threshold for rejecting drafts can lead to higher acceptance rates and faster generation, but it may also result in lower text quality. In Decomposition-and-Fill (\S \ref{sec:sfg}), using a more conservative decomposition strategy to reduce decomposition granularity can improve quality but decrease speed. For Non-AR-based methods (\S \ref{sec:Non-AR}), the number of steps used in generation is closely tied to quality. For instance, in masked generation (\S \ref{sec:mg}), recovering fewer tokens from masks per step can improve quality but reduce speed. In edit-based refinement (\S \ref{sec:ebr}), reducing the number of refinement steps can increase speed but lower quality.

This trade-off stems from the balance between parallelism granularity and generation quality, which also creates a balance between speed and quality. Finding the optimal point in this balance is very difficult for different methods.

\subsection{Technique-Specific Challenges}\label{sec:technique-specific}

\subsubsection{Ignored Dependencies in High-Entropy Scenarios} \label{sec:ignored-dependencies}

This challenge applies differently across categories. Draft-and-verifying (\S \ref{sec:dav}) and Decomposition-and-Fill (\S \ref{sec:sfg}) are less affected, as draft-and-verifying uses a small model's autoregressive prediction, and Decomposition-and-Fill minimizes dependencies between positions through decomposition. Other categories are more affected, such as multi-token prediction (\S \ref{sec:mtp}) and all Non-AR-based methods (\S \ref{sec:Non-AR}).

Parallel generation often decodes multiple positions simultaneously, typically relying on the basic assumption that these positions are independent. In high-certainty, low-entropy scenarios, this assumption holds reasonably well due to strong determinism. However, in high-entropy scenarios, it poses significant challenges.

Current methods are limited by this issue. Multi-token prediction methods (\S \ref{sec:mtp}) see a large drop in acceptance rates. Non-AR-based methods, including one-shot generation (\S \ref{sec:osg}), masked generation (\S \ref{sec:mg}), and edit-based refinement (\S \ref{sec:ebr}), face strict constraints because they also decode different positions simultaneously.

This is a fundamental problem that may not be fully solvable. It can only be eased by reducing parallelism or adding light dependencies to avoid full independence.

\subsubsection{Conflicts with the Existing Optimization Ecosystem} \label{sec:conflicts-ecosystem}

This challenge does not affect AR-based methods (\S \ref{sec:AR}), but it does impact Non-AR-based methods (\S \ref{sec:Non-AR}).

Modern LLM inference relies heavily on the KV cache mechanism to avoid repeated computation on historical sequences. This mechanism is a key way in accelerating autoregressive models, as it reduces the computational complexity of generating $N$ tokens from $O(N^3)$ to $O(N^2)$.

However, Non-AR-based methods can hardly use the KV cache mechanism. This is mainly because bidirectional attention architectures make the generation influence of each token global.

Although some methods show that caching is possible for Non-AR methods through certain means \cite{wu2025fast, liu2025dllm}, these caches are approximate and may sacrifice some accuracy. Moreover, such caching essentially introduces block-level autoregression \cite{wu2025fast}.

While this problem may be solvable, addressing it would require substantial changes to the model's architecture and inference algorithms, which is highly challenging. For example, potential improvements might involve attention mechanisms that lie between causal and fully bidirectional designs.
\section{Conclusion}\label{sec:conclusion}

To meet the growing demands for faster and more efficient language model inference, parallel text generation has emerged as a promising paradigm—ranging from parallel decoding techniques to more recent developments like diffusion-based language models. In this survey, we provide a comprehensive overview of parallel text generation methods. We categorize existing approaches into autoregressive-compatible methods (e.g., draft-and-verify, decomposition-and-fill, multiple token prediction) and non-autoregressive paradigms (e.g., one-shot generation, masked generation, edit-based refinement), and present an in-depth analysis of the core techniques in each category.

Building on this taxonomy, we assess the theoretical trade-offs these methods make across three key dimensions: decoding speed, output quality, and resource efficiency. We also explore how different strategies can be composed together and how they interact with traditional acceleration techniques (e.g., model compression, caching, compiler-level optimizations).

Despite recent progress, several important challenges remain unresolved. These include the persistent quality–speed trade-off, the lack of task-specific benchmarks that jointly evaluate accuracy and efficiency, uncertainties about practical utility in real-world deployments, and the absence of unified, modular tooling for combining and deploying these techniques at scale.

In conclusion, parallel text generation is a rapidly evolving field with significant potential to reshape the landscape of language model inference. Continued progress is essential not only for reducing latency and improving throughput in large-scale applications (e.g., chatbots, real-time assistants, content generation), but also for enhancing accessibility and efficiency in resource-constrained settings (e.g., mobile devices, on-device inference). By addressing the challenges outlined in this survey, future research can pave the way for more reliable, efficient, and broadly usable parallel generation systems.

\bibliographystyle{ACM-Reference-Format}
\bibliography{reference}

\clearpage
\appendix
\section{Theoretical Comparison Analysis}\label{sec:appendix-comparison}

Building on the preceding analysis of parallel text generation methods, this section provides a detailed theoretical comparison from two perspectives: (1) the standalone strengths and trade-offs of each paradigm, (2) the potential for combining methods to achieve greater acceleration.

\subsection{Standalone Trade-offs: Speed, Quality, and Resource}

In parallel text generation, faster generation often comes at the expense of increased resource consumption and potential quality degradation. To enable a fair comparison, we first formally define the theoretical speed-up potential, output quality, and resource consumption of each parallel generation paradigm when applied independently.

As a reference point, we assume a baseline autoregressive generation system using the target model $\mathcal{M}_q$ with the following characteristics:
\begin{itemize}
    \item Resource usage: $M$ GPUs.
    \item Decoding speed: $S$ tokens per second.
    \item Generation quality: $P$ (e.g., measured by human preference or automated metrics such as BLEU or METEOR).
\end{itemize}

\subsubsection{Draft-and-Verifying}

\paragraph{\textbf{Speedup}}

Draft-and-Verifying introduces a lightweight draft model $\mathcal{M}_p$ and modifies the decoding process to perform speculative generation. Let $A$ denote the expected number of accepted tokens per verification step, and $L(\mathcal{M}_p)$ and $L(\mathcal{M}_q)$ be the per-step latency of the draft and target models, respectively. Under speculative decoding, the expected throughput is illustrated Equation~\ref{eq:dav-throughput}.

\begin{equation}
S' = \mathbb{E}\left[\frac{A}{L(\mathcal{M}_p) + L(\mathcal{M}_q)}\right]
\label{eq:dav-throughput}
\end{equation}

Given that $S = 1 / L(\mathcal{M}_q)$ in the baseline, the relative speedup ratio can be calculated as Equation~\ref{eq:dav-speedup}.

\begin{equation}
\text{Speedup} = \frac{S'}{S} = A \cdot \frac{L(\mathcal{M}_q)}{L(\mathcal{M}_p) + L(\mathcal{M}_q)}
\label{eq:dav-speedup}
\end{equation}

If $L(\mathcal{M}_p) \ll L(\mathcal{M}_q)$ and $A \approx K$ (i.e., most speculative tokens are accepted), the speedup approaches $A$—in practice, $2\times \sim 4\times$ is achievable.

\paragraph{\textbf{Quality}}

The output quality $P'$ under this paradigm remains close to baseline $P$, because:
(1) All accepted tokens are verified by $\mathcal{M}_q$. (2) Incorrect drafts are corrected immediately using $\mathcal{M}_q$.

However, aggressive drafting (larger $K$) or overly weak $\mathcal{M}_p$ can increase rejection and correction frequency, slightly degrading fluency or diversity. In controlled settings, $P' \approx P$ or has negligible degradation.

\paragraph{\textbf{Resource}}

The method requires one additional draft model $\mathcal{M}_p$ to be run per decoding step. Let $r = L(\mathcal{M}_p) / L(\mathcal{M}_q)$ denote the relative cost ratio. The total compute per step increases by a factor of $1 + r$, assuming both models run on the same hardware.

\begin{itemize}
    \item \textbf{If $\mathcal{M}_p$ runs on the same $M$ GPUs as $\mathcal{M}_q$}: There may be minor memory overhead and a small throughput penalty unless pipelining or parallelism is used.
    \item \textbf{If $\mathcal{M}_p$ runs on separate lightweight compute (e.g., CPU, edge GPU)}: Additional resource cost is negligible with proper scheduling.
\end{itemize}

Therefore, the resource usage becomes approximately $(1 + \delta) \cdot M$, where $\delta$ is a small constant ($\delta \ll 1$) if $\mathcal{M}_p$ is small or offloaded efficiently.

\subsubsection{Decomposition-and-Fill}

\paragraph{\textbf{Speedup}}
In Decomposition-and-Fill, the input prompt or intended output is first decomposed into semantically disjoint or loosely dependent components, enabling \textbf{parallel generation} of multiple segments. Suppose the decomposition yields $n$ independent sub-prompts, each of which is filled independently and concurrently.

Assuming each fill operation takes roughly equal time $T$ and the decomposition time is $T_{\text{decomp}}$, the total latency can be calculated as Equation~\ref{eq:daf-latency}.

\begin{equation}
T_{\text{decomp+fill}} \approx T_{\text{decomp}} + \max_{i=1}^{n} T_{\text{fill}_i} \approx T_{\text{decomp}} + T
\label{eq:daf-latency}
\end{equation}

In contrast, standard autoregressive decoding would require roughly $n \cdot T$ sequentially. Thus, the theoretical speedup is illustrated as Equation~\ref{eq:daf-speedup}.

\begin{equation}
\text{Speedup} \approx \frac{n \cdot T}{T_{\text{decomp}} + T}
\label{eq:daf-speedup}
\end{equation}

When $T_{\text{decomp}} \ll T$, this approaches an ideal $n\times$ speedup.

\paragraph{\textbf{Quality}} 
The output quality of Decomposition-and-Fill methods depends critically on the semantic adequacy of the decomposition and the coherence of the filling stage. If the decomposition phase successfully captures the global structure of the desired text (e.g., via outlines or keyphrases), the filling stage can generate fluent and contextually appropriate content under strong conditioning, often achieving comparable or even superior quality to autoregressive baselines.

However, suboptimal decompositions—e.g., overly coarse or semantically inconsistent structures—can lead to incoherent or repetitive text in the final output. Moreover, the local generation within each fill segment may lack global coordination, introducing inconsistencies across segments. Overall, while these methods can maintain high quality when decomposition is meaningful and fill models are well-trained, quality degradation may occur if either stage underperforms.

\paragraph{\textbf{Resource}} 
The parallel filling of $n$ segments typically requires $n$ concurrent decoding processes. If each uses a model with the same resource profile as the target model $\mathcal{M}_q$, the total GPU usage is illustrated in Equation~\ref{eq:daf-resource}.

\begin{equation}
M_{\text{decomp+fill}} \approx M_{\text{decomp}} + n \cdot M
\label{eq:daf-resource}
\end{equation}

where $M$ is the GPU count in the baseline, and $M_{\text{decomp}}$ is the overhead for decomposition (usually small). Using lighter models for fill stages can reduce this cost.

\subsubsection{Multiple Token Prediction}

\paragraph{\textbf{Speedup}} 

Multiple Token Prediction (MTP) modifies the autoregressive decoding process by generating $k$ tokens at each decoding step instead of one. Assuming the model $\mathcal{M}_q$ can be adapted to predict a span of $k$ tokens in parallel, the number of forward passes required to generate a sequence of length $L$ can be shown in Equation~\ref{eq:mtp-latency}, where $L(\mathcal{M}_{q}^{(k)})$ denotes the latency of a single $k$-token prediction. 

\begin{equation}
T_{\text{MTP}} = \frac{L}{k} \cdot L(\mathcal{M}_{q}^{(k)})
\label{eq:mtp-latency}
\end{equation}

Compared to standard AR decoding, which requires $L$ forward passes, as shown in Equation~\ref{eq:mtp-speedup}.

\begin{equation}
\text{Speedup} = \frac{L \cdot L(\mathcal{M}_q)}{\frac{L}{k} \cdot L(\mathcal{M}_q^{(k)})} = \frac{k \cdot L(\mathcal{M}_q)}{L(\mathcal{M}_q^{(k)})}
\label{eq:mtp-speedup}
\end{equation}

When the multi-token model introduces little overhead ($L(\mathcal{M}_q^{(k)}) \approx L(\mathcal{M}_q)$), near $k\times$ speedup is achievable.

\paragraph{\textbf{Quality}} 
Multiple Token Prediction (MTP) methods aim to generate several tokens per step, improving efficiency at the cost of modeling more complex output distributions. As the number of predicted tokens per step increases, it becomes more difficult to accurately model the joint probability of the full output span, potentially degrading generation quality.

This degradation arises from compounded prediction errors and misalignment with the true distribution, particularly in longer spans or open-ended generation tasks. Nevertheless, techniques such as span-level training, knowledge distillation, and output denoising can mitigate some of these issues. Overall, while MTP may approach baseline quality for small $k$, quality degradation tends to grow with larger spans.

\paragraph{\textbf{Resource}} 
MTP often requires architectural changes or retraining to support multi-token prediction. Let $L(\mathcal{M}_q^{(k)})$ represent the per-step latency of the modified model. If the modified model is larger or uses more attention computation, the per-step resource cost may increase. Denoting the baseline GPU usage as $M$, the resource cost is calculated as Equation~\ref{eq:mtp-resource}, where $\gamma_k \geq 1$ captures any overhead due to the expanded model or span prediction head.

\begin{equation}
M_{\text{MTP}} = M \cdot \gamma_k
\label{eq:mtp-resource}
\end{equation}

\subsubsection{One-shot Generation}

\paragraph{\textbf{Speedup}} 
One-shot Generation aims to produce the entire output sequence in a single forward pass. Let $L$ be the desired output length, and $\mathcal{M}_q^{(1s)}$ the one-shot generation model. The theoretical decoding latency is reduced as Equation~\ref{eq:one-shot-latency}.

\begin{equation}
T_{\text{one\_shot}} = L(\mathcal{M}_q^{(1s)})
\label{eq:one-shot-latency}
\end{equation}

compared to standard autoregressive decoding which takes $L \cdot L(\mathcal{M}_q)$. Therefore, the maximum theoretical speedup is shown as Equation~\ref{eq:one-shot-speedup}

\begin{equation}
\text{Speedup} = \frac{L \cdot L(\mathcal{M}_q)}{L(\mathcal{M}_q^{(1s)})}
\label{eq:one-shot-speedup}
\end{equation}

This can approach $L\times$ in ideal cases, assuming $L(\mathcal{M}_q^{(1s)}) \approx L(\mathcal{M}_q)$, though in practice it often incurs significantly larger latency per forward pass.

\paragraph{\textbf{Quality}} 
One-shot generation produces the entire output sequence in a single forward pass without intermediate feedback or correction, which can lead to notable quality degradation. Common issues include semantic inconsistencies, hallucinations, and syntactic errors, especially in longer or more structured texts.

The lack of step-wise conditioning removes the opportunity to refine predictions based on past outputs, which is a key advantage in autoregressive models. As a result, while one-shot generation can perform reasonably on short, well-constrained tasks, its generation quality often lags behind AR-based approaches in open-ended or complex scenarios.

\paragraph{\textbf{Resource}} 
One-shot Generation often requires longer context windows, increased memory usage, or larger model capacity to maintain output quality. Let $\gamma$ be the resource overhead relative to standard AR inference, as shown in Equation~\ref{eq:one-shot-resource}, where $\gamma > 1$ due to higher memory/computation from e.g., full-sequence cross-attention or custom decoding heads.

\begin{equation}
M_{\text{one\_shot}} = M \cdot \gamma
\label{eq:one-shot-resource}
\end{equation}

\subsubsection{Masked Generation}

\paragraph{\textbf{Speedup}} 
Masked Generation (e.g., Mask-Predict, Iterative Masking, BERT-style decoding) decodes the output in multiple rounds. In each round, a subset of tokens is masked and predicted in parallel. Let $L$ be the sequence length, $R$ the total number of decoding rounds, and $B_r$ the number of masked tokens in round $r$. The total number of forward passes is $R$, typically $R \ll L$, and each pass predicts $B_r$ tokens in parallel.

Assuming each round takes a latency of $L(\mathcal{M}_q)$ (same model used each round), the theoretical decoding latency is shown in Equation~\ref{eq:masked-latency}.

\begin{equation}
T_{\text{masked}} = R \cdot L(\mathcal{M}_q)
\label{eq:masked-latency}
\end{equation}

This yields a theoretical speedup, as shown in Equation~\ref{eq:masked-speedup}.

\begin{equation}
\text{Speedup} = \frac{L \cdot L(\mathcal{M}_q)}{R \cdot L(\mathcal{M}_q)} = \frac{L}{R}
\label{eq:masked-speedup}
\end{equation}

With effective scheduling or confidence-based masking, $R$ can be reduced significantly (e.g., 4–10), enabling $5\times$–$20\times$ speedups in ideal cases.

\paragraph{\textbf{Quality}} 
The quality of masked generation depends heavily on the number of refinement rounds and the accuracy of its masking strategy. If the model performs too few decoding steps or misidentifies low-confidence tokens, it may result in incoherent, repetitive, or degenerate outputs.

While increasing the number of refinement rounds can improve quality by progressively correcting errors, it also leads to higher computational cost. Therefore, masked generation faces a trade-off between quality and efficiency, and its performance often hinges on well-designed heuristics or learned masking schedules.

\paragraph{\textbf{Resource}} 
Masked Generation requires full-sequence prediction per round (like BERT), which results in quadratic attention overhead. The total GPU usage is illustrated in Equation~\ref{eq:masked-resource}, where $\beta > 1$ accounts for larger memory footprint due to global self-attention over the entire output. Additionally, $R$ forward passes amplify total compute cost despite the lower number of iterations than AR.

\begin{equation}
M_{\text{masked}} = M \cdot \beta
\label{eq:masked-resource}
\end{equation}

\subsubsection{Edit-Based Refinement}

\paragraph{\textbf{Speedup}}

Edit-based methods accelerate generation by iteratively editing an initial rough draft instead of generating from scratch. Let $T$ be the total number of tokens to generate. Rather than generating $T$ tokens sequentially using $\mathcal{M}_q$ (which would take $T / S$ seconds), edit-based methods first produce a coarse sequence using a lightweight model $\mathcal{M}_p$, and then apply $E$ refinement steps, each modifying a subset of tokens in parallel.

Assuming the editing model $\mathcal{M}_e$ has latency $L(\mathcal{M}_e) \ll L(\mathcal{M}_q)$ and each edit pass can modify $r \cdot T$ tokens (where $0 < r < 1$), the total time can be calculated as Equation~\ref{eq:ebr-latency}.

\begin{equation}
T_{\text{edit}} = L(\mathcal{M}_p) + E \cdot L(\mathcal{M}_e)
\label{eq:ebr-latency}
\end{equation}

If $T_{\text{edit}} < T / S$, the method achieves theoretical speedup, as illustrated in Equation~ref{eq:ebr-latency}.

\begin{equation}
\text{Speedup Ratio} = \frac{T / S}{L(\mathcal{M}_p) + E \cdot L(\mathcal{M}_e)} > 1
\label{eq:ebr-latency}
\end{equation}

In practice, the speedup depends on how quickly convergence is achieved and how effective each edit pass is.

\paragraph{\textbf{Quality}} 
The generation quality in edit-based refinement depends on both the initial draft and the effectiveness of the iterative editing process. Early edits typically improve structural consistency, while later ones enhance fluency and factual accuracy. 

When sufficient refinement steps are allowed, the final quality can approach or even exceed that of standard autoregressive generation. However, if constrained to a small number of editing rounds—e.g., due to latency constraints—the quality may suffer, especially for long-form or complex outputs.

\paragraph{\textbf{Resource}}

Edit-based refinement decouples generation into lighter submodules. The initial draft is typically generated by a small model $\mathcal{M}_p$ or even heuristics, and refinement uses an efficient model $\mathcal{M}_e$ (e.g., T5, BART). If all steps are executed on the same $M$ GPUs, resource usage per step is low, and memory usage is reduced due to shorter attention spans per edit step. Compared to baseline autoregressive decoding, as shown in Equation~\ref{eq:ebr-resource}.

\begin{equation}
\text{Total GPU load} \ll M \cdot T / S
\label{eq:ebr-resource}
\end{equation}

Especially when $\mathcal{M}_e$ is smaller than $\mathcal{M}_q$. Thus, these methods are attractive for resource-constrained or batched inference settings.

\subsection{Composability: Combining Acceleration Paradigms}

While each parallel generation paradigm provides independent acceleration, further gains may be achievable by combining multiple techniques in a single decoding pipeline. In this section, we analyze which combinations among the six paradigms—\textbf{One-shot Generation}, \textbf{Decomposition-and-Fill}, \textbf{Masked Generation}, \textbf{Edit-Based Refinement}, \textbf{Draft-and-Verify}, and \textbf{Multiple Token Prediction (MTP)}—can be theoretically composed for greater speedup, and which exhibit fundamental conflicts.

\paragraph{Composability Overview}

We consider all pairwise and multi-way compositions among the six paradigms. Each method modifies a distinct stage of the generation process, such as task granularity, decoding behavior, or refinement strategy. Composability is thus governed by whether their operational assumptions are compatible.

\paragraph{Valid Multi-way Compositions}

We analyze several valid multi-way compositions, building on the theoretical framework established in earlier sections. Each composition is evaluated with respect to its speedup potential and resource efficiency.

We begin by establishing a common baseline for fair comparison. Let the standard autoregressive generation setup use $M$ GPUs, decode at a rate of $S$ tokens per second, and achieve a quality score of $P$. Each acceleration method modifies the decoding process in different ways. The following parameters are used to characterize these effects:

\begin{itemize}
\item $n$: Number of decomposed sub-tasks.
\item $k$: Number of tokens predicted per step in Multi-Token Prediction (MTP).
\item $E$: Number of iterative steps in Edit-Based Refinement.
\item $R$: Number of masked generation iterations.
\item $\mathcal{M}_p$: Draft model used in Draft-and-Verify, with latency $L(\mathcal{M}_p)$.
\item $\mathcal{M}_q$: Target model, with latency $L(\mathcal{M}_q)$.
\item $\mathbb{E}[A]$: Expected number of tokens accepted per speculation in Draft-and-Verify.
\end{itemize}

Based on these definitions, we now provide a detailed analysis of each viable composition strategy.

\paragraph{[A] Decomposition + One-shot + Edit-Based Refinement}

This pipeline first applies \textbf{Decomposition-and-Fill} to break a complex input into $n$ semantically disjoint sub-tasks. Each segment is then decoded independently using \textbf{One-shot Generation}, allowing for highly parallel execution. The resulting outputs are subsequently refined using $E$ iterations of \textbf{Edit-Based Refinement} to improve fluency, coherence, and factual accuracy.

\textbf{Speedup:} The total decoding speedup is achieved by parallelizing both across sub-tasks and leveraging the faster decoding speed of one-shot generation. Assuming one-shot decoding runs at $S_{\text{one}}$ tokens/sec per segment, the effective speedup over the baseline autoregressive rate $S$ can be calculated as Equation~\ref{eq:comp-a-speedup}.

\begin{equation}
\text{Speedup}_{\text{A}} = \frac{n \cdot S_{\text{one}}}{S}
\label{eq:comp-a-speedup}
\end{equation}

\textbf{Resource Usage:} Each one-shot decoding can be distributed over $n$ GPUs. The $E$ steps of edit refinement incur additional resource usage, modeled as a fraction $c_E$ of baseline GPU usage per step. The total GPU consumption is estimated as Equation~\ref{eq:comp-a-resource}.

\begin{equation}
\text{GPU}_{\text{A}} = M + c_E \cdot E \cdot M
\label{eq:comp-a-resource}
\end{equation}

\textbf{Quality:} One-shot generation, while fast, often suffers from degraded fluency and consistency due to the absence of step-wise autoregressive conditioning. This typically results in lower initial quality compared to traditional generation. However, the subsequent edit-based refinement serves to recover factuality, coherence, and surface fluency. The final quality depends on both the severity of degradation introduced during one-shot decoding and the number of refinement steps applied. With sufficient editing, this stack can match—or in some cases exceed—the quality of fully autoregressive baselines, particularly for structured or decomposable tasks.

\paragraph{[B] Decomposition-Fill + Multiple Token Prediction + Edit-Based Refinement}

This pipeline applies \textbf{Decomposition-and-Fill} to divide the input into $n$ parallelizable sub-tasks. Each segment is decoded using \textbf{Multiple Token Prediction (MTP)}, which generates $k$ tokens per decoding step, significantly accelerating generation compared to autoregressive decoding. To mitigate quality degradation caused by MTP, $E$ steps of \textbf{Edit-Based Refinement} are applied for post-hoc correction.

\textbf{Speedup:} MTP enables decoding $k$ tokens per step instead of one. When combined with $n$-way decomposition, the total theoretical speedup over a baseline decoding $T$ tokens sequentially is calculated as Equation~\ref{eq:comp-b-speedup}.

\begin{equation}
\text{Speedup}_{\text{B}} = \frac{n \cdot k}{T}
\label{eq:comp-b-speedup}
\end{equation}

This assumes full parallelism across segments and that the entire sequence length is $T$ tokens.

\textbf{Resource Usage:} Similar to the [A] pipeline, the edit phase contributes to resource consumption. The total GPU usage is calculated as Equation~\ref{eq:comp-b-resource}.

\begin{equation}
\text{GPU}_{\text{B}} = M + c_E \cdot E \cdot M
\label{eq:comp-b-resource}
\end{equation}

Where $c_E$ is the fractional cost per edit step relative to the base system.

\textbf{Quality:} MTP may degrade quality due to imperfect modeling of joint token probabilities across $k$-length spans. Let $\Delta P_{\text{mtp}}(k)$ denote this degradation, which increases with larger $k$. The refinement phase improves quality by $\Delta P_{\text{edit}}(E)$, giving as Equation~\ref{eq:comp-b-quality}.

\begin{equation}
P_{\text{B}} = P - \Delta P_{\text{mtp}}(k) + \Delta P_{\text{edit}}(E)
\label{eq:comp-b-quality}
\end{equation}

This composition balances high decoding throughput with post-hoc quality recovery. It is particularly effective when MTP quality drop is moderate and sufficient refinement steps can be afforded within latency constraints.

\paragraph{[C] Decomposition-Fill + Masked Generation}

This composition applies \textbf{Decomposition-and-Fill} to partition the input into $n$ semantically disjoint segments. Each segment is then completed using $R$ rounds of \textbf{Masked Generation}, where tokens are iteratively predicted and filled in masked positions without strict autoregressive dependency.

\textbf{Speedup:} Assuming that each segment requires $R$ rounds to converge and that segments are processed in parallel, the speedup relative to sequential decoding is shown in Equation~\ref{eq:comp-c-speedup}.

\begin{equation}
\text{Speedup}_{\text{C}} = \frac{n}{R}
\label{eq:comp-c-speedup}
\end{equation}

This assumes roughly one masked fill per segment per round, and that masking converges uniformly.

\textbf{Resource Usage:} Masked generation requires iterative forward passes, and thus introduces overhead. The effective GPU usage is calculated as Equation~\ref{eq:comp-c-resource}.

\begin{equation}
\text{GPU}_{\text{C}} = M \cdot R
\label{eq:comp-c-resource}
\end{equation}

This reflects either actual hardware usage (if rounds are pipelined) or accumulated compute cost (if rounds are sequential).

\textbf{Quality:} The overall output quality is influenced by both stages in this composition. First, Decomposition-and-Fill can introduce quality degradation if the segmentation fails to preserve global coherence—e.g., semantic drift, entity inconsistency, or incorrect task decomposition. This risk increases when the segments are loosely coupled or when dependencies span across segments.

Second, the quality of Masked Generation depends on its convergence behavior. If the number of refinement rounds $R$ is too small, the output may remain incoherent or incomplete. Increasing $R$ generally improves local fluency and factual consistency, but at the cost of additional compute.

Therefore, the quality trade-off arises from both decomposition granularity and the effectiveness of iterative masked refinement. The composition is best suited when segments are weakly interdependent and the masked decoder exhibits stable convergence.

\paragraph{[D] Decomposition-Fill + MTP + Edit + Draft-and-Verify}

This pipeline aggressively exploits parallelism and speculation. First, Decomposition-and-Fill splits the input into $n$ segments. Each is generated coarsely using Multiple Token Prediction (MTP) with step size $k$. The coarse outputs are then refined using $E$ steps of Edit-Based Refinement. Finally, Draft-and-Verify applies speculative decoding to further accelerate AR-based refinement.

\textbf{Speedup:} The overall speedup comes from three factors: segment-level parallelism ($n$), span-level MTP ($k$), and token-level speculation (via expected acceptance $\mathbb{E}[A]$). Dividing by the combined latency of the draft and target models and accounting for $E$ edit rounds gives as Equation~\ref{eq:comp-d-speedup}.

\begin{equation}
\text{Speedup}_{\text{D}} = n \cdot k \cdot \frac{\mathbb{E}[A]}{L(\mathcal{M}_p) + L(\mathcal{M}_q)} \cdot \frac{1}{E}
\label{eq:comp-d-speedup}
\end{equation}

\textbf{Resource Usage:} This stack requires running both the draft and target models during verification, plus additional compute for $E$ edit rounds, as calculated in Equation~\ref{eq:comp-d-resource}.

\begin{equation}
\text{GPU}_{\text{D}} = M + M_{\text{draft}} + E \cdot M
\label{eq:comp-d-resource}
\end{equation}

Here, $M_{\text{draft}}$ is the number of GPUs used by the lightweight speculative model $\mathcal{M}_p$. The total resource footprint scales with the number of refinement iterations.

\textbf{Quality:} The final output quality in this stack is shaped by multiple interacting stages. First, the use of MTP introduces potential degradation due to inaccuracies in predicting token spans without full autoregressive context. This risk grows with larger step sizes $k$. The subsequent Edit-Based Refinement helps to correct these coarse errors, especially for fluency and local coherence.

Finally, the Draft-and-Verify stage acts as a safeguard against semantic or factual errors by verifying speculative drafts before acceptance. If well-calibrated, it can effectively reject low-quality predictions from the earlier stages and ensure high fidelity.

Overall, this composition offers strong error recovery mechanisms. When all components are tuned appropriately, the quality can approach or even exceed that of standard autoregressive decoding—especially in long-form or structured generation tasks where coarse-to-fine refinement is beneficial.

\paragraph{[E] Decomposition-Fill + One-shot}

This is the lightest-weight composition, aimed at maximizing speed while minimizing compute cost. The input is first split into $n$ independent sub-tasks via \textbf{Decomposition-and-Fill}, and each segment is generated in parallel using \textbf{One-shot Generation}, without any further refinement or correction.

\textbf{Speedup:} With $n$ sub-tasks processed in parallel, and assuming each one-shot decoder achieves speed $S_{\text{one}}$, the total speedup relative to baseline autoregressive speed $S$ is calculated as Equation~\ref{eq:comp-e-speedup}.

\begin{equation}
\text{Speedup}_{\text{E}} = \frac{n \cdot S_{\text{one}}}{S}
\label{eq:comp-e-speedup}
\end{equation}

\textbf{Resource Usage:} Since no refinement or speculative decoding is used, the stack requires only the baseline GPU pool, as shown in Equation~\ref{eq:comp-e-resource}.

\begin{equation}
\text{GPU}_{\text{E}} = M
\label{eq:comp-e-resource}
\end{equation}

\textbf{Quality:} Due to the lack of any refinement or verification stages, this composition is more susceptible to quality degradation. Common issues include hallucinations, inconsistency, and lack of coherence—particularly for complex or tightly constrained tasks. Since each segment is generated independently without autoregressive context or feedback, the quality may fall short of more robust pipelines. However, for tasks where minor inaccuracies are acceptable—such as early-stage drafts, informal outputs, or high-volume data augmentation—the trade-off can be justified by the significant speed and resource advantages.

\subsubsection{Incompatible Paradigms}

Despite the flexibility of most acceleration paradigms, certain combinations are fundamentally incompatible due to irreconcilable input/output assumptions or conflicting control flow. Table~\ref{tab:composability-matrix} identifies such combinations as $\times$. Below, we elaborate on these truly disjoint pairs.

\paragraph{One-shot Generation $\times$ Masked Generation:}
One-shot generation produces complete outputs without intermediate tokens or masks. Masked generation, in contrast, expects input with unresolved placeholders (e.g., [MASK]) and iteratively fills them. Since the output of one-shot decoding has no remaining uncertainty or structure to complete, it cannot serve as valid input to a masked decoder.

\paragraph{One-shot Generation $\times$ Draft-and-Verify:}
Draft-and-Verify operates on incremental token-level predictions and requires token-by-token verification. One-shot generation, by definition, generates the full sequence in a single forward pass without exposing intermediate states or token-level traceability. This makes it fundamentally incompatible with the verification phase.

\paragraph{One-shot Generation $\times$ MTP:}
MTP assumes the model continues generating from an existing prefix, predicting multiple future tokens at once. One-shot decoding provides no such context—it emits a full sequence without further continuation points. Thus, MTP has no prefix to work with, and one-shot has no room for extension, rendering this pairing structurally incoherent.

\paragraph{Masked Generation $\times$ MTP:}
MTP is designed to predict consecutive future tokens (typically at the tail of a sequence), while masked generation fills arbitrary non-contiguous positions across the sequence. There is no direct alignment between MTP’s output format and the input structure expected by masked generation. Without a coordination mechanism, the two operate on incompatible assumptions.

\section{Continuous Time Formulation of Masked diffusion models} \label{appendix:mdm_math}
Recent works~\cite{shi2024simplified,sahoo2024simple,zheng2024masked,ou2024your} have shown that masked generative models~\cite{chang2022maskgit,ghazvininejad2019mask} can be formally reinterpreted as a discrete masked diffusion process. This perspective offers a principled foundation for understanding and improving masked models. Accordingly, we adopt the continuous time formulation of the masked diffusion model (MDM) to unify these approaches, which we begin to introduce below. For a more comprehensive mathematical treatment of masked diffusion models, readers are referred to~\cite{campbell2022continuous,shi2024simplified,zheng2024masked,ou2024your}.

In masked diffusion, we aim to model a probability distribution over a discrete state space $\mathcal{X} = \{1, \dots, m-1\}$, representing tokens from a finite vocabulary. To incorporate masking, we augment this space with an additional mask state indexed by $m$. Thus, the state distribution is represented as a probability mass vector $p \in \mathbb{R}^m$, where $p \ge 0$ and $\sum_{i=1}^m p_i = 1$. Given a training dataset, let $p_{\text{data}}$ denote the underlying data distribution.
Masked diffusion models generate samples by learning to reverse a corruption process that progressively transforms a clean token $x_0 \sim p_{\text{data}}$ into a fully masked token $x_T$, where $x_T$ follows a reference distribution $p_{\text{ref}}$. For the masked diffusion model, it is a delta probability mass function concentrated on the mask state $m$: $p_{ref}^{mask}(x):=\delta_{m, x}$.
Akin to the diffusion models for continuous data~\cite{sohl2015deep,ho2020denoising,song2021scorebased}, this corruption procedure is referred to as the \textit{forward process}, while its reversal is the \textit{reverse process}. 
The forward process can be formulated either as a \textit{Discrete-time Markov chain} (DTMC)~\cite{austin2021structured}, where transitions occur at discrete time steps, or as a \textit{Continuous-time Markov chain} (CTMC)~\cite{anderson2012continuous}, where the perturbation happens on a continuous time process from $t=0$ to $t=T$. Most recent studies adopt the CTMC formulation~\cite{campbell2022continuous,lou2024discrete,zheng2024masked,ou2024your}, we follow this convention and elaborate on the CTMC formulation below.

\paragraph{Forward process}
Masked diffusion models utilize a CTMC to define the forward corruption process for each individual token. Specifically, for a single token (i.e., a one-dimensional state), the infinitesimal transition probability is given by:
\begin{equation} \label{eqn:forward_process}
p_{t+\Delta t \mid t}(y \mid x)
= \delta_{x, y} + Q_t(x, y)\,\Delta t + o\left(\Delta t\right),
\end{equation}
where $p_{t+\Delta t \mid t}(y\mid x)$ denotes the probability of being in state $y$ at time $t+\Delta t$ given that the process is in state $x$ at time $t$.  
$Q_t \in \mathbb{R}^{|\mathcal{X}| \times|\mathcal{X}|}$ is the transition rate matrix of the CTMC at time $t$. The off-diagonal elements satisfy $Q_t(x, \hat{x}) \ge 0$ for $x \neq \hat{x}$, and the diagonal elements are defined as  
$Q_t(x, x) = - \sum_{\hat{x} \neq x} Q_t(x, \hat{x}) \le 0,$
ensuring that each row sums to zero and thus preserving total probability mass.

Given the CTMC defined in Equation~\eqref{eqn:forward_process}, the evolution of the marginal distribution $p_t = (p_t(x))_{x\in\mathcal{X}}$ follows the Kolmogorov forward equation:
\begin{equation} \label{eqn:ode}
\frac{d p_t}{d t} = p_t Q_t, \quad p_0 = p_{\text{data}},
\end{equation}
where $p_t$ represents the marginal distribution of the token state at time $t$.

Efficient computation of the forward process relies on obtaining a closed-form expression for $p_{t|0}$. A common strategy is to parameterize the rate matrix as $Q_t = \sigma(t)\,Q_{\text{absorb}}$, where $\sigma(t)$ is a time-dependent scaling function (i.e., noise schedule) and $Q_{\text{absorb}}$ is a fixed transition rate matrix defined as:
\begin{equation}
\boldsymbol{Q}_{\text {absorb}}
:= \mathbf{1} e_m^T - I
=
\begin{bmatrix}
-1 & 0 & \cdots & 0 & 1 \\
0 & -1 & \cdots & 0 & 1 \\
\vdots & \vdots & \ddots & \vdots & \vdots \\
0 & 0 & \cdots & -1 & 1 \\
0 & 0 & \cdots & 0 & 0
\end{bmatrix},
\end{equation}
where $\mathbf{1}$ is an all-ones vector of length $m$, $e_m$ is a one-hot vector with a $1$ at the $m$-th entry, and $I$ is the $m\times m$ identity matrix.  
To extend the forward process to a multi-dimensional case, the single token forward process is applied independently to each token, progressively replacing tokens with the mask state and thus converting clean data into a fully masked sequence.

\paragraph{Reverse process}
The forward process Equation~\eqref{eqn:ode} has a reversal given by another rate matrix $\overline{Q}_t$~\cite{kelly2011reversibility,lou2024discrete}:
\begin{equation} \label{ode_reverse}
    \frac{dp_{T-t}}{dt} = p_{T-t}\overline{Q}_{T-t}
\end{equation}
where $\overline{Q}_t$ is a transition rate matrix for the reversed CTMC that satisfies:
\begin{equation}
\bar{Q}_t(x, \hat{x})= \begin{cases}\frac{p_t(\hat{x})}{p_t(x)} Q_t(\hat{x}, x), & \hat{x} \neq x \\ -\sum_{y \neq x} \bar{Q}_t(x, y), & \hat{x}=x\end{cases}
\end{equation}
The only unknown quantity in the reversal process is the ratio $\frac{p_t(y)}{p_t(x)}$, which is known as the \textit{concrete score}~\cite{meng2022concrete,lou2024discrete}. We use a parametric neural network $p_\theta(x)\in\mathbb{R}^m$ to approximate the concrete score s.t. $p_\theta(x)_y \approx \frac{p_t(y)}{p_t(x)}$.
We can generate samples by simulating the reversed CTMC:
\begin{equation} \label{eq:reverse_kernel}
    p_{t-\Delta t\mid t}(y \mid x) = \delta_{xy} +  \overline{Q}_t(y,x)\Delta t + o(\Delta t)
\end{equation}

Exact simulation of the reverse process using algorithms like Gillespie's~\cite{wilkinson2018stochastic,gillespie1976general,gillespie1977exact} is inherently sequential, rendering it ill-suited for efficient parallel generation. Because this is a continuous-time Markov chain with the corruption process applied independently to each dimension~\cite{campbell2022continuous,lou2024discrete}, the probability of two or more dimensions transitioning at the same instant is zero. As a result, any transition in the full-dimensional process, whether forward or reverse, alters at most one dimension. This fundamental constraint creates a bottleneck on decoding speed, necessitating approximate simulation methods to enable the parallel updating of multiple tokens simultaneously.

\end{document}